\documentclass{article}

\PassOptionsToPackage{numbers, compress}{natbib}

\usepackage[nonatbib,preprint]{neurips_2023}
\usepackage{authblk}

\usepackage[normalem]{ulem}

\usepackage{setspace}
\usepackage{comment}
\usepackage{amsmath,amsthm,amssymb}
\usepackage{ulem}
\usepackage{makecell}

\newcommand{\bs}{\boldsymbol}
\def\RR{ \mathbb R}
\newcommand{\refeqp}[1]{Equation (\ref{#1})}

\providecommand{\keywords}[1]
{\textbf{\text{Keywords: }} #1}
\newcommand{\ee}{\end{equation}}
\newcommand{\be}{\begin{equation}}
\newcommand{\ec}{\end{center}}
\newcommand{\bc}{\begin{center}}
\newcommand{\eea}{\end{eqnarray}}
\newcommand{\bea}{\begin{eqnarray}}
\newcommand{\bd}{\begin{description}}
\newcommand{\ed}{\end{description}}
\newcommand{\bi}{\begin{itemize}}
\newcommand{\ei}{\end{itemize}}
\newcommand{\pa}{\partial}
\newcommand\bphi{\boldsymbol{\phi}}

\newcommand{\bx}{\bs{x}}

\newcommand{\bt}{\bs{\theta}}

\newcommand{\Ncal}{\mathcal{N}}
\newcommand{\Bcal}{\mathcal{B}}
\newcommand{\Acal}{\mathcal{A}}

\newcommand{\Ucal}{\mathcal{U}}

\newcommand{\Gcal}{\mathcal{G}}

\newcommand{\bmx}{\bm{x}}
\newcommand{\bmbeta}{\bm{\beta}}

\usepackage{commath}
\usepackage{mathtools}
\usepackage{algorithm,algpseudocode}
\usepackage{hyperref}
\usepackage{color}
\usepackage{amsfonts}

\usepackage{diagbox}
\usepackage{bm}
\usepackage{booktabs}       
\usepackage{multirow}
\usepackage{subfigure}
\usepackage[percent]{overpic}
\usepackage{caption}

\newcommand{\revise}[1]{\textcolor{red}{#1}}

\newcommand{\bmtheta}{{\bm{\theta}}}

\algnewcommand{\Inputs}[1]{%
  \State \textbf{Inputs:}
  \Statex \hspace*{\algorithmicindent}\parbox[t]{.8\linewidth}{\raggedright #1}
}
\algnewcommand{\Initialize}[1]{%
  \State \textbf{Initialize:}
  \Statex \hspace*{\algorithmicindent}\parbox[t]{.8\linewidth}{\raggedright #1}
}
\algnewcommand{\Outputs}[1]{%
  \State \textbf{Outputs:}
  \Statex \hspace*{\algorithmicindent}\parbox[t]{.8\linewidth}{\raggedright #1}
}

\title{DGenNO: A Novel Physics-aware Neural Operator for Solving Forward and Inverse PDE Problems based on Deep, Generative Probabilistic Modeling}

\author[a]{Yaohua Zang}
\author[a,b]{Phaedon-Stelios Koutsourelakis}
\affil[a]{Technical University of Munich, Professorship of Data-driven Materials Modeling, School of Engineering and Design, Boltzmannstr. 15, 85748 Garching, Germany}
\affil[b]{Munich Data Science Institute (MDSI - Core member), Garching, Germany}
\affil[ ]{\text{\{yaohua.zang, p.s.koutsourelakis\}@tum.de}}
\begin{document}
\maketitle
\begin{abstract}
Solving parametric partial differential equations (PDEs) and associated PDE-based, inverse problems is a central task in engineering and physics, yet existing neural operator methods struggle with high-dimensional, discontinuous inputs and require large amounts of {\em labeled} training data. We propose the Deep Generative Neural Operator (DGenNO), a physics-aware framework that addresses these challenges by leveraging a deep, generative, probabilistic model in combination with a set of lower-dimensional, latent variables that simultaneously encode PDE-inputs and PDE-outputs. This formulation can make use of unlabeled data and significantly improves inverse problem-solving, particularly for discontinuous or discrete-valued input functions. 
DGenNO enforces physics constraints without labeled data by incorporating as virtual observables, weak-form residuals based on  compactly supported radial basis functions (CSRBFs). These  relax regularity constraints and eliminate higher-order derivatives from the objective function. We  also introduce MultiONet, a novel neural operator architecture, which is a more expressive generalization of the popular DeepONet that significantly enhances the approximating power of the proposed model. These innovations make DGenNO particularly effective for challenging forward and inverse, PDE-based problems, such as  those involving  multi-phase media.  
Numerical experiments demonstrate that DGenNO achieves higher accuracy across multiple benchmarks while exhibiting robustness to noise and strong generalization to out-of-distribution cases. Its adaptability, and the ability to handle sparse, noisy data while providing probabilistic estimates, make DGenNO a powerful tool for scientific and engineering applications.
The implementation of DGenNO is available through the Github repository \href{https://github.com/pkmtum/DGenNO}{https://github.com/pkmtum/DGenNO}.
\end{abstract}
\section{Introduction}
\label{sec:introduction}
Partial Differential Equations (PDEs) serve as fundamental mathematical models for describing various physical phenomena, including fluid dynamics \cite{batchelor2000introduction}, heat conduction \cite{hahn2012heat}, electromagnetism \cite{jones2013theory}, and material deformation \cite{meakin1991models}. Solving PDEs, both in forward and inverse settings, is crucial for scientific discovery and engineering applications such as medical imaging \cite{adler2021electrical,li2009vivo,scholz2025weak}, climate modeling \cite{randall2007climate}, non-destructive testing \cite{gholizadeh2016review}, and material design \cite{zang2024psp}. The forward problem involves computing the PDE solution given initial/boundary conditions, and parameters, whereas the inverse problem seeks to infer unknown parameters, initial/boundary conditions, or source terms from observations/measurements. 
Despite the maturity of numerical methods for the solution of PDEs, their repeated solution under different parametric values in the context of many-query applications such as sensitivity analysis, uncertainty quantification, and inverse problems, represents a major computational roadblock in achieving analysis and design objectives. 

In recent years, machine learning, and neural network-based methods, in particular, have emerged as a powerful tool for approximating PDE solutions. Among these, physics-aware deep learning approaches have gained significant popularity due to their ability to embed governing physics directly into the solution process. These methods utilize deep neural networks to approximate PDE solutions, by employing loss functions based on residuals of governing equations and mismatches in initial/boundary conditions. Notable examples include physics-informed neural networks (PINNs) \cite{raissi2019physics} and their variants \cite{pang2019fpinns,jin2021nsfnets,gao2021phygeonet}, which make use of strong-form residuals in the learning objectives. Other approaches, such as the DeepRitz method (DRM) \cite{yu2018deep}, weak adversarial networks (WAN) \cite{zang2020weak,bao2020numerical}, and variational PINNs (VPINNs) \cite{kharazmi2019variational}, formulate the loss function by using weak-form residuals \cite{zhu_physics-constrained_2019}.
These methods offer several advantages over traditional numerical techniques for solving forward and inverse PDE-based problems. Key benefits include their mesh-free nature, the ability to mitigate the curse of dimensionality \cite{han2018solving,sirignano2018dgm,zang2020weak}, robustness to noisy data \cite{bao2020numerical,both2021deepmod}, and inherent regularization properties \cite{yan2019robustness,mowlavi2023optimal}.
However, despite these strengths, existing approaches are typically restricted to solving a single PDE instance with fixed coefficients, source terms, and initial/boundary conditions. Any change in these parameters requires retraining the model, making them inefficient for applications that demand repeated PDE solutions, such as parametric PDE problems and inverse problems.

A promising approach to overcoming the aforementioned limitation is the use of deep neural operators (DNOs), which have gained significant attention as a general framework for learning mappings between function spaces. By parameterizing the solution operator of a PDE with a neural network, DNOs efficiently generalize across varying input conditions, coefficients, and domains. 
A notable example is the DeepONet \cite{lu2021learning}, which leverages the universal operator approximation theorem \cite{chen1995universal} to learn nonlinear operator mappings between infinite-dimensional function spaces. Its architecture consists of two neural networks: a branch network that encodes input functions and a trunk network that encodes output function coordinates. This structure significantly outperforms classical fully connected networks in learning parametric PDEs and provides advantages for solving inverse problems as well \cite{kaltenbach_semi-supervised_2023}.
Another prominent neural operator is the Graph Neural Operator (GNO) \cite{li2020neural}, which learns the kernel of integral transformations through a message-passing framework on graph networks. However, the GNO architecture often exhibits instability as the number of hidden layers increases. To address this, the Fourier Neural Operator (FNO) \cite{li2020fourier} was introduced, leveraging Fourier transforms to perform integral operations at each layer, thereby enhancing efficiency and scalability. Subsequent advancements have refined this approach, including the Adaptive Fourier Neural Operator (AFNO) \cite{guibas2021adaptive}, the Implicit Fourier Neural Operator (IFNO) \cite{you2022learning}, the Wavelet Neural Operator (WNO) \cite{tripura2023wavelet}, and the Geo-Fourier Neural Operator (Geo-FNO) \cite{li2023fourier}. Additionally, the Spectral Neural Operator (SNO) \cite{fanaskov2023spectral} was proposed to mitigate opaque outputs and aliasing errors inherent in FNO-based models.
While these neural operator frameworks effectively learn mappings from input functions-such as coefficients, source terms, and initial/boundary conditions-to the PDE solutions, they are largely data-driven. Their predictive accuracy heavily depends on an impractically large amount of high-precision training data, typically generated by computationally expensive, traditional, numerical methods such as FDM and FEM.

To address the limitations of data-driven DNOs, physics-aware DNOs have been developed to solve parametric PDEs by incorporating governing physics into the training process. The Physics-Informed Neural Operator (PINO) \cite{li2024physics} extends the FNO by integrating physics-based constraints into the data-driven learning of function space mappings. When fine-tuned, PINO demonstrates promising results for several PDEs. However, it relies on point-wise differentiation and fine mesh grids to approximate derivatives in its loss function, limiting its efficiency.
As an alternative, the Physics-Aware Neural Implicit Solver (PANIS) \cite{chatzopoulos2024physics} employs a probabilistic learning objective, using weighted residuals to probe the PDE and generate virtual data. This enables probabilistic predictions with improved generalization. While PANIS achieves comparable accuracy to PINO and superior performance on out-of-distribution cases, its application has so far been restricted to time-independent forward PDE problems.
Another notable approach is the Physics-Informed DeepONet (PI-DeepONet) \cite{wang2021learning,goswami2023physics}, which adapts the DeepONet architecture and can be trained purely on physics constraints. This allows it to learn the solution operator of arbitrary PDEs without requiring {\em labeled} training  data, i.e. input-output pairs. However, its reliance on strong-form residuals makes it challenging to handle PDEs with singular or discontinuous inputs and outputs. The Physics-Informed Variational DeepONet (PI-VDeepONet) \cite{goswami2022physics} extends the DeepONet framework by incorporating the variational form of PDEs, making it particularly useful for predicting crack paths in quasi-brittle materials. Despite its advantages, PI-VDeepONet is limited to PDEs with energy-like function formulations.
The Physics-Informed Wavelet Neural Operator (PI-WNO) \cite{navaneeth2024physics} builds upon wavelet-based neural operators (WNO) \cite{gupta2021multiwavelet,tripura2023wavelet} to solve parametric PDEs using purely physics-based losses. A key innovation is its stochastic projection method for approximating the solution's derivatives. However, this approach requires a high density of sample points near derivative locations and may struggle with accuracy for complex output solutions.
The Physics-Informed Deep Compositional Operator Network (PI-DCON) \cite{zhong2024physics} introduces a compositional adaptation of DeepONet, designed for generalized applications to various discrete representations of PDE parameters and irregular domain geometries. This framework was later extended to handle variable domain geometries \cite{zhong2025physics}. While PI-DCON possesses enhanced expressivity, its reliance on strong-form residuals, like PI-DeepONet, limits its effectiveness for PDEs with singular inputs or outputs.
Despite their advancements, most physics-aware DNOs remain focused on forward problems and do not address inverse problems, such as recovering input coefficients from noisy output observations.

To address both forward and inverse PDE problems, the Physics-Informed PointNet (PI-PointNet) \cite{kashefi2022physics} was introduced for solving steady-state incompressible flow. This method integrates the geometric feature extraction capabilities of PointNet \cite{qi2017pointnet} with physics constraints, allowing it to handle problems across multiple irregular geometries. Although PI-PointNet considers inverse problems, the specific case studied is relatively simple—recovering the output function from its noise-free observation over the spatiotemporal domain. Additionally, its reliance on differentiating max-pooling layers to compute output derivatives degrades performance \cite{zhong2025physics}.
In \cite{vadeboncoeur_random_2023}, Random Grid Neural Processes (RGNPs), a probabilistic deep learning approach for solving forward and inverse problems PDE-based problems were proposed. By marginalizing over random collocation grids, RGNPs improve flexibility, computational efficiency, and predictive performance compared to traditional physics-informed machine learning methods. The framework also integrates noisy data from arbitrary grids while maintaining uncertainty quantification through Gaussian processes but it is based on collocation-type residuals.
Although it does rely on DNOs, the Physics-Driven Deep Latent Variable Model (PDDLVM) \cite{vadeboncoeur_fully_2023} offers a probabilistic framework for learning forward and inverse maps of parametric PDEs using Gaussian distributions parameterized by deep neural networks. However, it depends on conventional PDE discretization and spectral Chebyshev representation, limiting its applicability to high-dimensional problems and those with singularities. Furthermore, PDDLVM tackles inverse problems by first reconstructing the solution from fixed sensor observations before predicting the input function. 
The most relevant work to ours is the recent study in \cite{jiao2024solving}, where PI-DeepONet was applied to PDEs and inverse problems on unknown manifolds. In this method, the forward mapping from input functions to output solutions was learned using PI-DeepONet, while the inverse problem was tackled using Bayesian Markov Chain Monte Carlo (MCMC) with the trained DNO serving as the forward simulator. However, this approach requires a well-chosen prior for the target coefficient and it is unsuitable for high-dimensional problems. Moreover, its reliance on strong-form residuals presents difficulties in handling problems with discontinuities.

\revise{This paper introduces a fundamentally new framework, called Deep Generative Neural Operator (DGenNO), by integrating deep generative modeling with neural operator learning in a purely physics-driven regime. The framework is specifically designed to address two under-explored but important challenges: (1) solving both forward and inverse problems within a unified framework and
particularly in cases involving discrete-valued input fields such as those encountered in multi-phase media, and (2) enabling  purely  physics-based training, without requiring labeled, i.e., PDE input–output pairs. Existing DNO approaches treat these problems separately and usually follow a data-driven paradigm—learning direct mappings from high-dimensional, potentially discontinuous input (or observation) functions to PDE solutions (or inverse target) using large amounts of labeled training data. While physics-aware DNO methods incorporate physical constraints to enhance performance, they  generally still rely on labeled data and do not fully operate in a physics-driven regime. Moreover, these methods are often sensitive to input noise and inflexible when  observation locations vary, as they are typically trained on fixed spatial grids—limitations that become particularly pronounced in inverse problem settings.} \revise{In contrast, the proposed DGenNO method takes a fundamentally different approach by using governing equations (expressed as weighted residuals) as the sole source of supervision, and by unifying the treatment of forward and inverse problems within a deep generative modeling framework. This direction is not only methodologically novel but also motivated by the practical realities of scientific computing, where labeled input–output pairs are often costly or infeasible to obtain. To address these challenges,} the DGenNO leverages a deep, generative model and latent variables that provide a lower-dimensional, continuous representation of PDE-input functions as well as of PDE-solutions through neural operators \cite{rixner_probabilistic_2021}.
The use of latent representations offers two key advantages. First, it simplifies learning by transforming the complex function-to-function mapping into a more tractable latent-variable-to-function mapping. Second, it provides an efficient mechanism for solving both forward and inverse problems. By simultaneously learning a generative, reconstruction map from the latent space to the input function space, the inverse problem reduces to identifying the latent representation that best aligns with noisy observations. This approach is especially beneficial for high-dimensional and discontinuous input functions, as optimizing in a lower-dimensional, continuous latent space is significantly more efficient. This becomes crucial in problems where the input field is discrete-valued, as is the case in multi-phase media. In such cases and regardless of the accuracy of the forward model surrogate derivatives of the forward map with respect to the PDE input would be unavailable. As a result, solving inverse problems would require cumbersome, non-derivative-based evolutionary strategies (such as random-walk MCMC), which are only effective in very low dimensions.

We express both generative maps to the PDE-input and PDE-output with a novel, neural operator architecture, called MultiONet, which constitutes  a more expressive generalization of the popular DeepONet. We incorporate physics-based information   into the learning objectives, in a fully Bayesian fashion by  introducing {\em virtual observables} \cite{kaltenbach_incorporating_2020}. These are expressed in the form of  weighted residuals computed with compactly supported radial basis functions (CSRBFs) as weighting functions \cite{finlayson_method_1972} in PDEs. \revise{To the best of our knowledge, this is the first \textit{general-purpose} DNO framework that defines the PDE loss using weak-form residuals, which applies to a broader class of PDEs beyond those with energy-like formulations.} The use of weighted residuals, proposed in \cite{zang2023particlewnn}, significantly reduces the number of integration points required to approximate integrals in the loss function. This approach also provides several benefits, including relaxing the regularity requirements of the solution and eliminating higher-order derivatives from the loss expressions. As a result, our model excels at handling irregular problems, such as Darcy flow with multiphase (piecewise-constant) permeability fields.
Combined with the proposed MultiONet architecture, the DGenNO framework demonstrates clear advantages over the state-of-the-art method in both forward and inverse PDE problems.
The main contributions of this work can be summarized as follows:
\begin{itemize}
    \item A novel physics-aware neural operator method is developed based on a deep generative probabilistic framework \cite{rixner_probabilistic_2021} for solving both forward and inverse PDE problems. This formulation  is  capable of making use of unlabeled data (i.e., only inputs)  and  leverages  weak-form residuals of the governing  PDE, enabling the solution of these problems without labeled training data. The probabilistic nature of the framework also allows for the quantification of uncertainty in the predictions. \revise{A comparative summary of the proposed DGenNO against several representative DNO methods is provided in Table \ref{tab:compact_comparison}, focusing on aspects such as problem settings, supervision requirements, learning paradigms, and flexibility in handling inverse problems.}
    \item The original high-dimensional, discontinuous input function space is transformed into a low-dimensional, well-structured latent space. This shift redefines the learning task from a function-to-function mapping to a latent-to-function mapping, offering significant advantages for solving inverse problems involving discontinuous or discrete-valued inputs.
    \item A novel neural operator architecture is proposed to learn mappings from the latent space to both the PDE-input and PDE-output function spaces, demonstrating greater expressivity than the standard DeepONet architecture.
    \item Comprehensive experiments on challenging benchmarks involving  forward and inverse PDE problems are conducted, demonstrating the efficiency and superiority of the proposed framework compared to state-of-the-art methods.
\end{itemize}
\begin{table}[htbp]
\centering
\scriptsize
\caption{Comparison of DGenNO with existing DNO methods. Abbreviations: Fwd=Forward, Inv=Inverse, Fun2Fun = function-to-function, Lat2Fun = latent-to-function, Req.= requirement, Prob. = problems, w/o =without, Obs.=Observation, Loc.=Location.}
\begin{tabular}{|l|c|c|c|c|c|}
\hline
\diagbox{\textbf{Aspect}}{\textbf{Methods}} & \textbf{DeepONet} & \textbf{FNO} & \textbf{PI-DeepONet} & \textbf{PINO} & \textbf{DGenNO (Ours)} \\
\hline
\textbf{Target} & Fwd & Fwd & Fwd/Inv & Fwd & Fwd \& Inv \\
\hline
\textbf{Learning} & Fun2Fun & Fun2Fun & Fun2Fun & Fun2Fun & Lat2Fun \\
\hline
\makecell[l]{\textbf{Labeled}\\\textbf{Data Req.}} & Yes (large) & Yes (large) & Yes (small) & Yes (small) & No \\
\hline
\makecell[l]{\textbf{Training}\\\textbf{Strategy}} & Data-driven & Data-driven & Physics-guided & Physics-guided &  Physics-driven \\
\hline
\textbf{Mesh Invariant} & No & Yes & No & Yes & Yes \\
\hline
\makecell[l]{\textbf{Regular}\\\textbf{Mesh Req.}} & No & Yes & No & Yes & No \\
\hline
\makecell[l]{\textbf{Inverse Prob.}\\\textbf{(w/o pairs)}} & No & No & Limited & No & Yes \\
\hline
\makecell[l]{\textbf{Uncertainty}\\\textbf{Quantification}} & No & No & No & No & Yes \\
\hline
\makecell[l]{\textbf{Obs. Loc.}\\\textbf{Flexibility}} & Fixed & Fixed & Flexible & Fixed & Flexible \\
\hline
\end{tabular}
\label{tab:compact_comparison}
\end{table}

The remainder of the paper is organized as follows. Section \ref{sec:problem} defines the operator learning task within the context of parametric PDE problems, using an example that maps coefficients to PDE solutions. The inverse problem addressed in this work is also formally defined. 
Furthermore, it introduces MultiONet, a novel neural operator architecture for approximating all operators in this study. In Section \ref{sec:method}, we provide a detailed presentation of the proposed DGenNO framework, including the actual and virtual data employed, the latent variables, the constituent densities, and model parameters. We also present the training procedure which is based on the Variational Bayesian Expectation-Maximization scheme \cite{beal_variational_2003} and discuss how one can obtain efficient, probabilistic predictions for both the forward and inverse problems using the trained DGenNO model. Section \ref{sec:experiments} provides comparative results on several parametric, forward, and inverse PDE problems, showcasing the effectiveness of the proposed method and its advantages over state-of-the-art approaches. Finally, Section \ref{sec:conclusion} summarizes the findings and discusses potential directions for future enhancements.
\section{Problem Statement and the  MultiONet Architecture}
\label{sec:problem}
\subsection{Parametric PDEs and related inverse problems}
The primary objective of this work is to develop a novel DNO framework for solving  general families of parametric PDEs, as well as the associated inverse problems. To provide a clear context for our discussion, we denote a parametric PDE as:
\begin{equation}\label{eq:pde_general}
\begin{cases}
    \Ncal_a[u](\bmx) = f(\bmx), \quad \bmx\in\Omega, \\
    \Bcal[u](\bmx) = g(\bmx), \quad \bmx\in\partial\Omega,
\end{cases}
\end{equation}
where $a \in \Acal$, $f \in \Ucal^*$, and $g \in \Ucal$ represent the coefficient function, source term, and boundary conditions, respectively. Here, $\Omega \subset \mathbb{R}^d$ is a bounded domain. The solution $u: \Omega \rightarrow \mathbb{R}$ is assumed to reside in a Banach space $\Ucal$, while the operator $\Ncal_a: \Acal \times \Ucal \rightarrow \Ucal^*$ depends on the coefficient $a$, mapping the Banach space $\Ucal$ of the solution to its dual space $\Ucal^*$, where the source term resides. A prototypical example of $\Ncal_a$ is the second-order elliptic equation operator, $\Ncal_a = -\text{div}(a \nabla \cdot)$. The operator $\Bcal$ represents the boundary operator, which maps the solution to the boundary conditions, such as Dirichlet or Neumann conditions. 
\paragraph{The forward problem} The forward problem involves approximating the solution $u \in \Ucal$ of the PDE for any given coefficient function $a \in \Acal$. Such an approximation can be expressed with a neural operator $\mathcal{G}$ that maps the coefficient function $a$ to the solution $u$, as follows:
\begin{equation}
    \Gcal: a\in\Acal \rightarrow u\in\Ucal.
    \label{eq:op}
\end{equation}
\paragraph{The inverse problem} The inverse problem involves recovering the coefficient function $a$ from observations $u_{obs}$ related to the solution/output $u$ of the PDE. These observations are typically noisy and are often available only at sparsely distributed locations within the problem domain or its boundaries.
\paragraph{\revise{Challenges of parametric PDEs and inverse problems without labeled training pairs}}\revise{Solving forward and inverse problems governed by PDEs in the absence of labeled training pairs presents fundamentally greater challenges than standard data-driven approaches. In this physics-driven regime, several interrelated difficulties arise:
\begin{itemize}
    \item\textbf{Absence of labeled training data:} Without access to pairs of PDE-input–output samples, existing DNO methods that rely on supervised learning to approximate mappings become inapplicable. This affects both the forward problem, where the solution operator $\mathcal{G}(a)$ must be approximated, and particularly the inverse problem, where recovering $a$ from sparse and noisy  observations of $u$ becomes severely under-determined. In such cases, models must rely entirely on the governing PDEs (i.e., physical laws) for training, requiring fundamentally different strategies that do not assume access to any precomputed solutions.
    \item\textbf{Singular, discontinuous or discrete-valued PDE inputs}: Many real-world problems involve parameter fields with sharp discontinuities or non-smooth variations, such as those encountered in multi-phase, heterogeneous media. 
    These present two major challenges: 
    \begin{itemize}
        \item \textbf{Forward problem:} existing Physics-aware DNOs typically rely on strong-form PDE residuals, such as $-\nabla \cdot (a \nabla u) - f$. However, when $a$ is discontinuous, its gradient does not exist at the discontinuity interfaces. Numerical approximations of these gradients lead to large errors near the discontinuities, resulting in significant degradation of predictive accuracy.
        \item \textbf{Inverse problem:} In the absence of labeled pairs, data-driven-based DNOs that learn a direct map from observations of $u$ to target $a$ become infeasible. Physics-aware DNO methods could be an alternative to provide a physics-informed surrogate model for the forward map $\Gcal(a)$. However, such a model would struggle to solve this problem due to the following reasons: (1) Gradient-based optimization/inference methods require differentiating with respect to $a$, but the fact that $a$ takes discrete values implies that such derivatives are not even defined;  (2) Non-gradient methods such as evolutionary techniques or MCMC would struggle when the representation of $a$ is high- or infinite-dimensional and are only practical in low-dimensional settings. 
    \end{itemize}
    \item\textbf{Sparse and noisy solution observations}: Inverse problems typically rely on limited and potentially noisy measurements of the solution $u$ at (potentially random) scattered spatial locations. This severely underdetermines the inverse mapping from $u$ to $a$. Even in supervised settings, sparse observations pose challenges for model identifiability. Noise further exacerbates instability and amplifies the ill-posedness. To address this, models must incorporate structural priors, low-dimensional structure, or probabilistic representations to promote stable and physically consistent inference.
    \item\textbf{Generalization and uncertainty quantification}: In neural operator learning, models must generalize to varying problem instances, including different input distributions, spatial discretizations, and observation placements.  Many existing DNO methods are trained on fixed grids and struggle to extrapolate to new conditions. Moreover, uncertainty quantification becomes essential in problems with noise, but most DNO methods do not inherently support uncertainty quantification or treat it only in the context of supervised learning. Finally, integrating all these elements—generalization, stability, and uncertainty—into a single coherent framework that leverages physical laws alone remains a largely open challenge.
\end{itemize}
We emphasize that these challenges are fundamental and, in some cases, have not been previously addressed due to their inherent difficulty. Our work aims to provide a unified solution framework that explicitly tackles these issues. 
}
\subsection{The MultiONet architecture}
We present a novel neural operator architecture, termed the MultiONet architecture\footnote{\revise{To avoid confusion, we now consistently refer to the MultiONet architecture (also the DeepONet architecture) as a network “structure” to distinguish it from a complete DNO method, which includes not only the network structure
but also the training paradigm, loss formulation, and problem scope.}}, designed to parametrize the subsequent operator models used in this paper. We note that, formally, the method presented involves a finite-to-infinite-dimensional neural map (as it is (implicitly) the case with several alternatives). The finite-dimensional input vector is denoted as $\bm{\beta}$. In Section \ref{sec:method}, we discuss how this is actually a learned, latent representation of the actual, PDE-input coefficient $a$ in the context of a generative architecture that enables the extension to infinite-to-infinite-dimensional maps. For illustration and comparison, we first present the MultiONet architecture as a model for approximating the forward PDE solution map, as described in \refeqp{eq:op}.

\begin{figure}[tb]
\centering
\includegraphics[width=1.\textwidth]{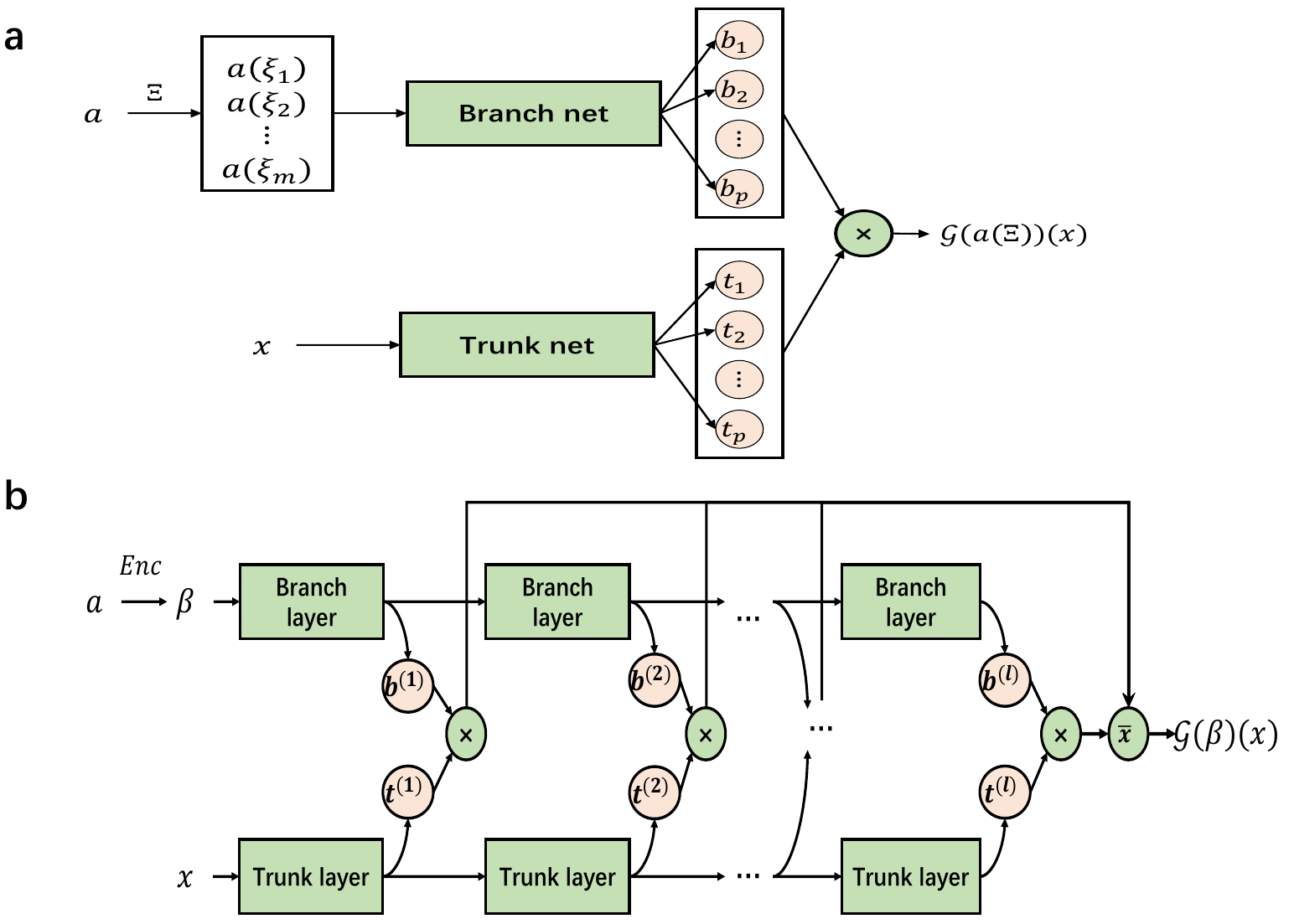}
\caption{a) DeepONet architecture   \cite{lu2021learning} vs. b) MultiONet architecture (proposed).}
\label{fig:WNIO_vs_DeepONet}
\end{figure}

The structure of the proposed architecture is illustrated in Figure \ref{fig:WNIO_vs_DeepONet}b. Similar to the DeepONet architecture, the MultiONet architecture employs a separable representation consisting of two neural networks: one, called the trunk network, encodes the spatial coordinates $\bm{x} \in \Omega$ of the output solutions, while the other, called the branch network, extracts features from the input  vector $\bm{\beta}$, which is automatically learned from the input coefficient $a$ using an encoder model. However, unlike the DeepONet architecture, the MultiONet architecture computes the final output as the average of the output products from multiple trunk and branch layers, instead of relying solely on the product of the output layers of the trunk and branch networks. This design improves performance compared to the DeepONet architecture, without increasing the number of network parameters.

As shown in Figure \ref{fig:WNIO_vs_DeepONet}b, the input to the branch network in MultiONet is the (latent) representation $\bm{\beta}$ of the input function. This representation can be obtained either through learning an encoder network or by extracting features using methods such as Fourier, Chebyshev, or wavelet transforms. In contrast, the branch network in the DeepONet architecture directly takes as input the discretized and finite-dimensional  representation of the coefficient function, $a(\Xi) = \{a(\xi_1), \cdots, a(\xi_m)\}$, which correspond to the values of  $a$ at a predefined set of so-called sensors $\Xi = \{\xi_1, \cdots, \xi_m\}$. This key difference allows the MultiONet architecture to offer greater flexibility in choosing the input to the branch network and reduces computational time, particularly when the sensor set $\Xi$ is large. Furthermore, while $a(\Xi)$ resides in a high-dimensional and irregular space—such as in multi-phase media, where $a$ is a discontinuous and discrete-valued field, the latent space of $\bm{\beta}$ is lower-dimensional, continuous-valued, and regular. This transformation provides significant advantages in solving the inverse problem, as optimization in the latent space is more efficient and robust.
Moreover, as demonstrated in the sequel, the MultiONet architecture shows superior approximation capabilities compared to the DeepONet architecture, even when both models have the same number of trainable parameters. \revise{This improvement primarily stems from the averaging mechanism inherent in the MultiONet architecture as illustrated in Figure \ref{fig:WNIO_vs_DeepONet}, which resembles the effect of ensemble learning by aggregating predictions from multiple functional bases.} In the DeepONet architecture, the output is calculated as the inner product of the outputs of the branch and trunk networks, expressed as:
\be\label{eq:deeponet}
\Gcal(a(\Xi))(\bm{x}) = \sum^{p}_{k=1}b_k(a(\Xi))t_k(\bm{x}) + b_0,
\ee
where $b_k(a(\Xi))$ and $t_k(\bm{x})$ are the $k$-th components of the outputs from the branch and trunk networks, respectively, and $b_0$ is a bias term. In contrast, the proposed MultiONet architecture calculates the output by averaging the weighted inner products across the outputs of multiple layers of the branch and trunk networks, expressed as:
\be\label{eq:multionet}
\Gcal(\bm{\beta})(\bm{x}) =\frac{1}{l} \sum^{l}_{k=1}w^{(k)}\left(b^{(k)}(\bm{\beta})\odot t^{(k)}(\bm{x}) \right)+b_0,
\ee 
where $b^{(k)}(\bm{\beta})$ and $t^{(k)}(\bm{x})$ denote the outputs from the $k$-th layers of the branch and trunk networks, $l$ represents the total number of layers, $w^{(k)}$ indicates trainable weight, $b_0$ is the bias term, and $\odot$ represents the inner product operation. It is easy to see that the DeepONet architecture can be considered a special case of the MultiONet architecture when only the outputs from the final layers of the branch and trunk networks are utilized in \eqref{eq:multionet}.
To handle cases where the branch and trunk networks have different numbers of layers, the architecture modifies the output computation as follows. Let $l_t$ and $l_b$ denote the number of layers in the trunk and branch networks, respectively, with the assumption that $l_t > l_b$. The final output is then computed by averaging the inner products of the following form:
\be\label{eq:multionet_general}
\Gcal(\bm{\beta})(\bm{x}) =\frac{1}{l_b} \sum^{l_b}_{k=1}w^{(k)}\left(b^{(k)}(\bm{\beta})\odot t^{(k+l_t-l_b)}(\bm{x}) \right)+b_0.
\ee 
The MultiONet architecture contributes to DGenNO by providing stronger expressive power compared to the DeepONet architecture. This increased representational capacity leads to improved performance when the MultiONet architecture is used as the operator approximator. This is evident from the comparative results between PI-DeepONet and PI-MultiONet, where the latter replaces the DeepONet architecture with the MultiONet architecture, across various benchmark problems in Section~\ref{sec:experiments}.

\section{Methodology}
\label{sec:method}
\revise{
We would like to clarify several key concepts that are essential for understanding the proposed DGenNO method:
\begin{itemize}
    \item \textit{Weighting functions:} Such functions are used in the weak formulation of PDEs. They are selected from a suitable function space and are used to project the residual of the governing equation, enabling the derivation of integral (weak-form) expressions that are less sensitive to discontinuities or singularities in the solution or coefficients.
    \item \textit{Weighted residuals:} In physics-informed learning, residuals quantify the discrepancy between a candidate solution and the governing physical laws (e.g., PDEs). Weighted residuals are obtained by integrating these residuals against carefully chosen weighting functions (test functions) over the domain, offering a flexible weak-form formulation that enhances numerical stability and allows for irregular inputs.
    \item \textit{Actual observables:} These refer to the available measurements or data samples of the physical field (e.g., the solution $u$) at limited locations. In our setting, actual observables are sparse and noisy values of the PDE solution, typically used as constraints in inverse problems and provided for a specific instance.
    \item \textit{Virtual observables:} Virtual observables are fictitious observations 
    used in order to incorporate  physical laws in the probabilistic learning objectives. They are not real measurements but are instrumental in guiding the model to learn solutions that are consistent with the governing equations. Importantly, they allow us to train the model without ground truth solutions $u$ or labeled pairs $(a,u)$—a significant advantage in many real-world problems where such data is costly or unavailable \cite{kaltenbach_incorporating_2020,rixner_probabilistic_2021}.
    \item \textit{Amortized variational inference:} This is a technique from probabilistic machine learning in which inference over latent variables is carried out via a shared neural network (encoder) rather separately for each data instance \cite{kingma2013auto}. In our formulation, this allows efficient inference of the latent representation $\bm{\beta}$ from sparse observations, enabling generalization and fast solution of pertinent forward and inverse problems.
\end{itemize}
}

\begin{figure}[tb]
\centering
\includegraphics[width=0.65\textwidth]{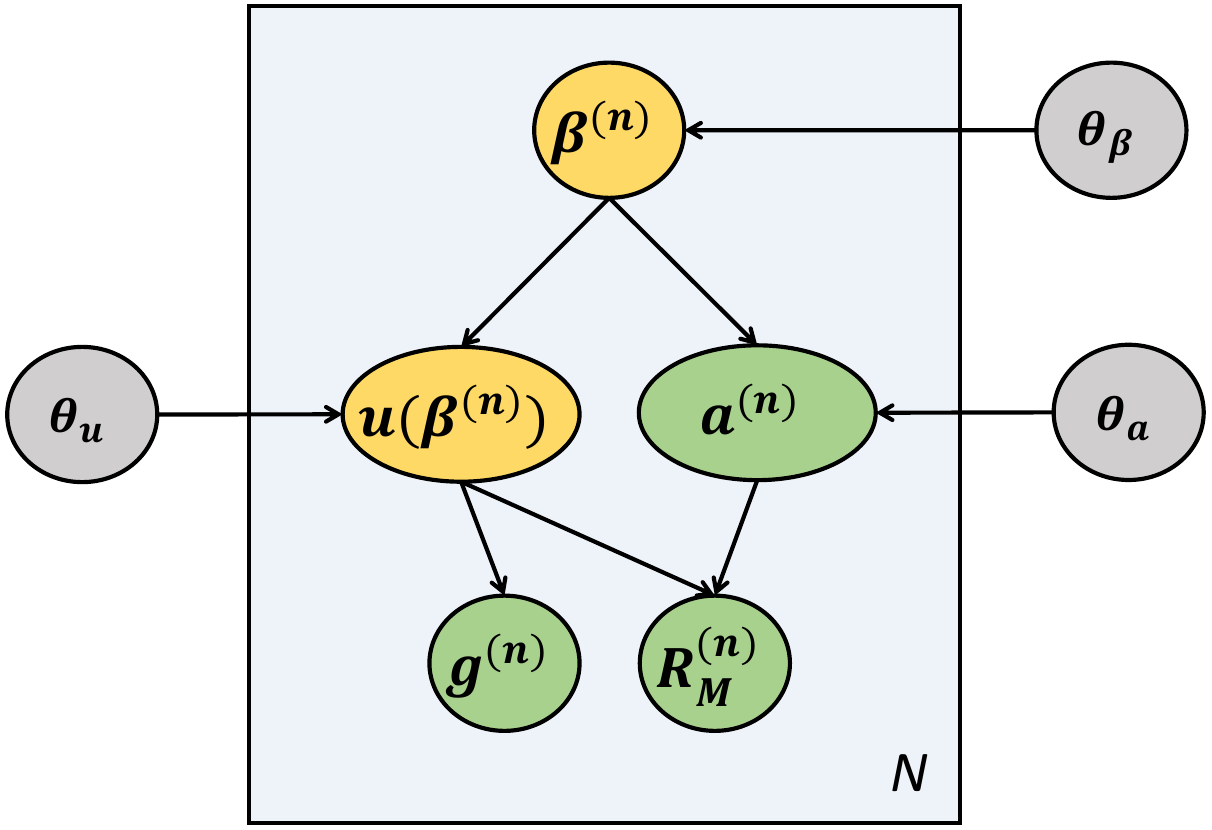}
\caption{Probabilistic graphical model illustration for the DGenNO framework proposed. The nodes in yellow represent unobserved (latent) variables, the nodes in green correspond to observed variables, and the nodes in gray represent (learnable) parameters. The arrows indicate dependencies, with the parent node conditioning the child node. The conditional densities defined within the model give rise to the overall model density, as captured in \refeqp{eq:Prob_Model}. This density encodes the forward mapping and is trained as described in Section \ref{sec:train}. Further details regarding the structure and parametrization of the constituent densities are provided in the main text.}
\label{fig:DGenNO}
\end{figure}

We propose  Deep Generative Neural Operators (DGenNO), a generative, probabilistic, physics-aware, deep-neural-operator-based framework that accurately approximates the forward map for parametric PDEs and introduces latent representations to enable efficient and robust inverse problem solving, as illustrated in the graphical model of Figure \ref{fig:DGenNO}. We believe that the probabilistic structure of our framework provides a key advantage over alternatives as it enables the quantification of both epistemic and aleatoric uncertainties—particularly important in Small-Data and extrapolative (out-of-distribution) settings where uncertainty estimates are critical for robust decision making. A key innovation of this framework is the introduction of a finite-dimensional, latent variable vector, denoted by  $\bm{\beta} \in \RR^{d_{\beta}}$, which serves as a generator of both the PDE-input function $a$ as well as of the solution $u$. \revise{The rationale is as follows: suppose that $\bm{\beta}$ serves as a latent representation of the input function $a$—that is, $a$ can be accurately reconstructed from $\bm{\beta}$. Then, since the output $u$ is uniquely determined by $a$ through the forward map $\mathcal{G}(a)$ (assuming the PDE is well-posed), it follows that $\bm{\beta}$ also implicitly encodes sufficient information to represent $u$.} This latent variable $\bm{\beta}$ resides in a lower-dimensional and well-structured space, replacing the original coefficient function $a$ as the input to the neural operator models. \revise{This approach offers several key benefits: a) it simplifies the challenging function-to-function mapping into a more tractable latent-to-function mapping for the forward problem and, b) the latent representation is invariant to the discretization and resolution of the input, rendering the DGenNO framework inherently mesh- and resolution-independent, c) it provides an effective pathway for solving forward and inverse problems by leveraging a reconstruction map from the latent space to the space of input coefficients $a$, thereby shifting the optimization to a smooth, low-dimensional latent space. This shift makes gradient-based optimization practical and robust, even for challenging inverse problems involving sparse, noisy observations and discontinuous or singular inverse targets.}
In brief, while most efforts focus on approximating the forward map $a \longrightarrow u$, we attempt instead to approximate the bidirectional map $a \longleftarrow \bm{\beta} \longrightarrow u$. Instead of acting as a lower-dimensional filter or projection of $a$ (e.g., $a \longrightarrow \bm{\beta} \longrightarrow u$), $\bm{\beta}$ functions as a set of {\em generators} for both $a$ and $u$.

In the following subsections, we present the training data and likelihoods (Section \ref{sec:actual_virtual_data}), the role of latent variables such as $\bm{\beta}$ and their associated priors (Section \ref{sec:latent_prior}), and the procedure for inference and learning (Section \ref{sec:train}). Finally, we discuss how the trained model can yield efficient, probabilistic solutions for both forward and inverse problems (Section \ref{sec:predictions}).

\subsection{Actual and virtual data and corresponding likelihoods.}
\label{sec:actual_virtual_data}
We first describe the data modalities used for training the proposed generative model. These consist of the following:
\bi
\item \textbf{Input functions}: We consider  the dataset $\{ \hat{\bs{a}}^{(i)} \}_{i=1}^{N}$, which consists of values of the PDE-input function $a$ at a  number $d_a$ of spatial locations $\bs{X}_a=\{x_j\}_{j=1}^{d_a}$ in the problem domain $\Omega$. These might be associated with a regular (fine) grid or a randomly selected set of points and are generally inexpensive to generate (i.e. no PDE needs to be solved). We note that each of the $N$ data instances does not need to have the same number of observations, nor do the observations need to be at the same locations. We incorporate these observables with a likelihood of $
p_{\bt_a}(\hat{\bs{a}}^{(i)} | \bs{\beta}^{(i)})$ where $\bt_a$ denotes the associated trainable parameters. The vector $\bs{\beta}^{(i)}$ denotes the unobserved (latent), generators mentioned earlier. For continuous-valued, PDE-input fields, it takes the form:
\be
\label{eq:cts}
p_{\bt_a}(\hat{\bs{a}}^{(i)} | \bs{\beta}^{(i)}) = \Ncal(\mu_{\bt_a}({\bs{\beta}^{(i)}}, \bs{X}_a), \lambda^{-1}_{rec}\bm{I}),
\ee
where $\mu_{\bt_a}({\bs{\beta}^{(i)}}, \bs{X}_a)$ is parametrized by the MultiONet architecture with parameters $\bt_a$ and evaluated at the observation points $\bs{X}_a$.
For piecewise-constant fields that correspond to e.g. two-phase media, we use the following model\footnote{For multiphase media, the associated model can be generalized using the softmax-function.}:
\be
\label{eq:pwc}
p_{\bt_a}(\hat{\bs{a}}^{(i)} | \bs{\beta}^{(i)}) =
\prod_{j=1}^{d_a} sig(\mu_{\bt_a}(\bs{\beta}^{(i)},x_j))^{z_j^{(i)}}~
(1-sig(\mu_{\bt_a}(\bs{\beta}^{(i)},x_j) ) )^{(1-z_j^{(i)})},
\ee
where $z_j^{(i)}=1$ if $\hat{\bm{a}}^{(i)}_j$ is in  the first phase and $z_j^{(i)}=0$ otherwise. Although   $\mu_{\bt_a}(\bs{\beta}^{(i)},x)$ is expressed with the MultiONet architecture and is evaluated as each observation point $x_j$ as before, its meaning is different compared to \refeq{eq:cts}. Finally, we denote the sigmoid function with $sig(.)$.
\item \textbf{Boundary/initial conditions}: We define the dataset $\{ \hat{\bs{g}}^{(i)} \}_{i=1}^{N}$, which represents prescribed boundary (or initial) condition values of the unknown PDE-solution $u^{(i)}$. Therefore, it is naturally conditioned on the PDE solution. We assume that these are associated with a prescribed set of points on the boundary $\pa \Omega$ (which could differ with $i$), which are incorporated in our model with a Gaussian likelihood $
p(\hat{\bs{g}}^{(i)} | \bs{u}_g^{(i)})$ having the form:
\be
p(\hat{\bs{g}}^{(i)} | \bs{u}_g^{(i)}) = \mathcal{N}(\hat{\bs{g}}^{(i)} | \bs{u}_g^{(i)}, \lambda_{bc}^{-1} \bs{I}),
\label{eq:likeg}
\ee
where $\bs{u}_g^{(i)}$ denotes the values of the unknown solution at the same boundary points and the precision $\lambda_{bc}$ controls the strength with which these are enforced (i.e. as $\lambda_{bc} \to \infty$, then $\bs{u}^{(i)} \to \hat{\bs{g}}^{(i)}$). We note that we consider problems with the same BCs/ICs which means that $\hat{\bs{g}}^{(i)}=\hat{\bs{g}}$ are identical for all $i$'s. One could enforce these conditions a priori by e.g. in the case of Dirichlet BCs expressing the solution $u$ as $u(\bx)=g(\bx)+\psi(\bx)\tilde{u}(\bx)$ where $g(\bx)$ satisfies these BCs, $\psi(\bx)=0$ on the boundary and subsequently approximating $\tilde{u}$ instead of $u$.
\item \textbf{Weighted residuals}:  Instead of solving the governing equations multiple times for different values of $a$ to generate a labeled training dataset (i.e., pairs of $(a,u)$), we leverage weighted residuals (see Section \ref{sec:weak_residual}) as {\em virtual observables} \cite{kaltenbach_incorporating_2020}. In particular, given $M$ distinct weighting functions $w_j$, for each input $\hat{\bs{a}}^{(i)}$ above,  the corresponding weighted residuals $r_{w_j}(\hat{\bs{a}}^{(i)},u^{(i)})$ are assumed to have been {\em virtually} observed and are equal to $0$ and their values are $\hat{r}_j^{(i)}=0$. This implies that solution pairs $(\hat{\bs{a}}^{(i)},u^{(i)})$ of the PDE perfectly match these virtual observations. Therefore, the {\em virtual} observables $\hat{\bs{R}}_M^{(i)}=\{ \hat{r}_j^{(i)}=0 \}_{j=1}^M, \forall i$, which consist of $M$ weighted residuals, depend on both $(\hat{\bs{a}}^{(i)},u^{(i)}$) and induce a likelihood which is assumed to be of the form:
\be
\begin{array}{ll}
 p(\hat{\bs{R}}_M^{(i)} | \hat{\bs{a}}^{(i)},u^{(i)}) & =\prod_{j=1}^M p(\hat{r}_j^{(i)}=0 | \hat{\bs{a}}^{(i)},u^{(i)}) \\
 & = \prod_{j=1}^M \sqrt{\lambda_{pde}}  e^{ - \lambda_{pde} r_{w_j}^2(\hat{\bs{a}}^{(i)},u^{(i)})}.
  \end{array}
\label{eq:virtuallike}
\ee
The hyperparameter $\lambda_{pde}$ plays a crucial role in controlling the decay rate of the likelihood for a given pair $(\hat{\bs{a}}^{(i)},u^{(i)})$ when the corresponding residuals $r_{w_j}(\hat{\bs{a}}^{(i)},u^{(i)})$ deviate from zero. Alternative formulations of the likelihood, such as a Laplace distribution, are also possible \cite{chatzopoulos2024physics}. The primary function of virtual observables and the associated likelihood is to incorporate information from the governing equations without explicitly solving them. In Section \ref{sec:train}, we discuss how this likelihood is utilized. Although the proposed DGenNO framework is compatible with both weak-form and strong-form formulations, we adopt weak-form residuals in this work due to their key advantage: they eliminate the need to compute derivatives of potentially discontinuous inputs, such as piecewise-constant coefficients $a$, which can lead to significant numerical errors when using strong-form residuals. The specific formulations of the weighted residuals, the choice of weighting functions, and—most importantly—how the residuals are computed without requiring reference solutions $u$ are detailed in Section~\ref{sec:weak_residual}.
\ei

\noindent \textbf{Remarks:} \\
\bi
\item The previous discussion was based on the unavailability of actual PDE-solutions $u^{(i)}$ for any or all of the $N$ instances where the PDE-input $\hat{\bs{a}}^{(i)}$ is given. As a result, the unobserved solutions $u^{(i)}$ are latent variables in \refeqp{eq:virtuallike}. Nevertheless, we note that if such partial or full observations are available, they can also be incorporated with a likelihood similar to that of \refeqp{eq:likeg}.  
\item It is not necessary to employ virtual data and therefore residuals for all $N$ instances where the PDE-input $\hat{\bs{a}}^{(i)}$ is given. The proposed model can incorporate fully {\em unlabeled} data (i.e., just PDE-inputs) which are inexpensive but nevertheless can provide valuable information as our previous investigations have shown \cite{rixner_probabilistic_2021}.
\item In the weighted residuals and the corresponding likelihood of \refeqp{eq:virtuallike}, we assumed that the residuals depend solely on the known values $\hat{\bs{a}}^{(i)}$ of $a^{(i)}$. This does not impose any practical restrictions since, as we explain in \ref{sec:weak_residual}, the numerical computation of the weighted residuals involves a finite number of integration points.
\item The choice of Gaussian conditional models is primarily for their practicality and for consistency with a large body of prior work \cite{cotter2010approximation,zhang2021bayesian,vadeboncoeur2023fully}. Gaussian modeling enables efficient training and inference, while the flexibility of neural network parameterizations allows the learned conditional distributions to approximate complex, nonlinear relationships.
We note that, in principle, alternative distribution families (e.g., mixture models or normalizing flows) could also be used in our framework, and we suggest exploring these extensions in future work. 
\ei

\subsection{Latent variables and priors}
\label{sec:latent_prior}
The latent variables $\bm{\beta}$ are central to the proposed framework, providing a low-dimensional and structured representation of both the original high-dimensional and irregular input functions $a$ and the PDE solution $u$. While various approaches have utilized lower-dimensional features of $a$, such as Fourier, Chebyshev, or Wavelet transforms \cite{gupta2021multiwavelet,fanaskov2023spectral,viswanath2023neural}, or other dimensionality reduction techniques, these methods yield projections of $a$ that are agnostic to the solution $u$. In contrast, the finite-dimensional vector $\bs{\beta} \in \RR^{d_{\beta}}$ acts as the hidden generator for both the PDE input field $a$ and the PDE solution $u$ through the densities $p_{\bt_a}(a|\bs{\beta})$ and $p_{\bt_u}(u|\bs{\beta})$, respectively\footnote{Since $a,u$ are functions, we interpret these densities in the context of Information Field Theory \cite{alberts_physics-informed_2023}.}, which are parametrized by $\bt_a,\bt_u$. The former has already been introduced in Equations (\ref{eq:cts}) and (\ref{eq:pwc}). For the latter, we adopt a degenerate probability density based on the Dirac-delta function:
\be
p_{\bt_u}(u|\bs{\beta})=\delta \left( u - u_{\bt_u}(\bs{\beta}) \right),
\label{eq:pub}
\ee
where $u_{\bt_u}(\bs{\beta})$ is modeled using the MultiONet architecture described in Section \ref{sec:problem}, parameterized by $\bt_u$. Notably, non-degenerate densities, such as a (conditional) Gaussian Process, could be employed in place of \eqref{eq:pub} to account for predictive uncertainty but would require inferring $u$.

In the subsequent illustrations, we assume a uniform, prior distribution for $\bs{\beta}$ over the $d_{\beta}$-dimensional hypercube $[-1,1]^{d_{\beta}}$, meaning that no learnable parameters $\bt_{\beta}$ are involved in the prior density $p_{\bt_{\beta}}(\bs{\beta})$. We note however that a learnable, parametrized prior could reveal the presence of structure in the space of latent generators (e.g. clusters) which could be very useful not only in producing accurate predictions but also in providing insight into the PDE input-output map.
\subsection{Complete probabilistic model and training}
\label{sec:train}
If we collectively denote the model parameters as $\bt=\{\bt_a,\bt_u,\bt_{\beta} \}$, the combination of the aforementioned densities and latent variables yields the following likelihood for each set of actual/virtual observables $(\hat{\bs{a}}^{(i)}, \hat{\bs{g}}^{(i)},\hat{\bs{R}}_M^{(i)})$\footnote{Given that $u^{(i)}$ is a function, we denote with $\mathcal{D}u^{(i)}$ the corresponding path integral \cite{alberts_physics-informed_2023}.}:
\be
\begin{array}{ll}
p_{\bt}(\hat{\bs{a}}^{(i)}, \hat{\bs{g}}^{(i)},\hat{\bs{R}}_M^{(i)}) & =  \int p(\hat{\bs{R}}_M^{(i)} |\hat{\bs{a}}^{(i)},  u^{(i)})~p(\hat{\bs{g}}^{(i)} |u^{(i)})~p_{\bt_u}(u^{(i)} |\bs{\beta}^{(i)})~p_{\bt_a}(\hat{\bs{a}}^{(i)} | \bs{\beta}^{(i)}) ~p_{\bt_{\beta}}(\bs{\beta}^{(i)})~d\bs{\beta}^{(i)} ~\mathcal{D}u^{(i)}\\
& = \int p(\hat{\bs{R}}_M^{(i)} |\hat{\bs{a}}^{(i)},  u_{\bt_u}(\bs{\beta}^{(i)}))~p(\hat{\bs{g}}^{(i)} | u_{\bt_u}(\bs{\beta}^{(i)}))p_{\bt_a}(\hat{\bs{a}}^{(i)} | \bs{\beta}^{(i)}) ~p_{\bt_{\beta}}(\bs{\beta}^{(i)})~d\bs{\beta}^{(i)}.
\end{array}
\label{eq:Prob_Model}
\ee
We note that the computation and maximization of the likelihood involves marginalization over the latent variables $\bs{\beta}^{(i)}$ which we overcome as in the Variational Bayesian  Expectation-Maximization  (VB-EM) scheme \cite{beal_variational_2003} by introducing an auxiliary density $q_{\bphi}(\bs{\beta}^{(i)} )$ that lower-bounds the log-likelihood as follows:
\be
\begin{array}{ll}
\log p_{\bt}(\hat{\bs{a}}^{(i)}, \hat{\bs{g}}^{(i)},\hat{\bs{R}}_M^{(i)}) &  \ge \left< \log \cfrac{p(\hat{\bs{R}}_M^{(i)} |\hat{\bs{a}}^{(i)},  \bs{u}_{\bt_u}(\bs{\beta}^{(i)}))~p(\hat{\bs{g}}^{(i)} | \bs{u}_{\bt_u}(\bs{\beta}^{(i)}))~p_{\bt_a}(\hat{\bs{a}}^{(i)} | \bs{\beta}^{(i)}) ~p_{\bt_{\beta}}(\bs{\beta}^{(i)}) }{ q_{\bphi}(\bs{\beta}^{(i)} )} \right>_{q_{\bphi}(\bs{\beta}^{(i)} )} \\
 & = \mathcal{F}^{(i)}(\bt,\bphi),
\end{array}
\ee
It can be readily shown \cite{beal_variational_2003} that the optimal $q_{\bphi}$ is the (intractable) posterior of $\bs{\beta}^{(i)}$ (given $(\hat{\bs{a}}^{(i)}, \hat{\bs{g}}^{(i)},\hat{\bs{R}}_M^{(i)})$). Furthermore, the KL-divergence between the two aforementioned densities determines the gap between the true log-likelihood and the Evidence Lower BOund (ELBO) $\mathcal{F}^{(i)}$. In subsequent illustrations, we employ anamortized variational inference \cite{kingma2013auto,ganguly2023amortized} scheme with a degenerate density for $q_{\bphi}(\bs{\beta}^{(i)})$ of the form:
\be
q_{\bphi}(\bs{\beta}^{(i)}  | \hat{\bs{a}}^{(i)})=\delta \left(\bs{\beta}^{(i)}-e_{\bphi}(\hat{\bs{a}}^{(i)}) \right).
\label{eq:q}
\ee
Here, $e_{\bphi}$ represents an encoder network parameterized by $\bphi$, the structure of which is described in \ref{sec:network}. In the next section, we demonstrate how this encoder facilitates efficient solutions to the forward problem.

Training the model, i.e. determining the optimal parameter values $\bt^*$, involves maximizing the log-likelihood of all $N$ data instances, which is lower-bounded as:
\be\label{eq:lb}
\sum_{i=1}^N \log p_{\bt}(\hat{\bs{a}}^{(i)}, \hat{\bs{g}}^{(i)},\hat{\bs{R}}_M^{(i)}) \ge \sum_{i=1}^N \mathcal{F}^{(i)}(\bt,\bphi) = \mathcal{F}(\bt,\bphi),
\ee
We employ an iterative maximization scheme of the total ELBO $\mathcal{F}$ which involves alternating  between \cite{beal_variational_2003}:
\bi
 \item \textbf{E-step:} Given $\bt$, find the optimal $\bphi$ by maximizing $\mathcal{F}$,
 \item \textbf{M-step:} Given $\bphi$, find the optimal $\bt$ by maximizing $\mathcal{F}$.
 \ei

\subsubsection{Form of the Evidence Lower Bounds $\mathcal{F}^{(i)}$}
Using the form of the associated densities from Section \ref{sec:actual_virtual_data} as well as of the variational approximation $q_{\bphi}$, we can
simplify the expressions for the lower bounds $\mathcal{F}^{(i)}(\bt,\bphi)$ in \refeqp{eq:lb}. In
particular, if we drop the data index $(i)$ which appears as a superscript in the expressions, we obtain:
\be\label{eq:loss}
\begin{array}{ll}
\mathcal{F}(\bt,\bphi) & = \left<\log p(\hat{\bs{R}}_M |\hat{\bs{a}},  \bs{u}_{\bt_u}(\bs{\beta})) \right>_{q_{\bphi}(\bs{\beta}|\hat{\bs{a}} )} + \left<\log p(\hat{\bs{g}} | \bs{u}_{g,\bt_u}(\bs{\beta})) \right>_{q_{\bphi}(\bs{\beta} |\hat{\bs{a}})} \\
& + \left<\log p_{\bt_a}(\hat{\bs{a}} | \bs{\beta})\right>_{q_{\bphi}(\bs{\beta}|\hat{\bs{a}})} + \left<\log \frac{p_{\bt_{\beta}}(\bs{\beta})}{q_{\bphi}(\bs{\beta})}\right>_{q_{\bphi}(\bs{\beta}|\hat{\bs{a}})} \\
& = \mathcal{F}_{pde}(\bt,\bphi) + \mathcal{F}_{bc}(\bt,\bphi) + \mathcal{F}_{rec,a}(\bt,\bphi) + \mathcal{F}_{kl}(\bphi).
\end{array}
\ee
The first term in \eqref{eq:loss} represents the contribution of the weighted residuals and is maximized when the latter are (on average) minimized. In particular, given the likelihood in \refeqp{eq:virtuallike} and $q_{\bphi}$ in \refeqp{eq:q}, it can be written as:
\be
\label{eq:loss_pde}
\mathcal{F}_{pde}(\bt,\bphi) = \frac{M}{2} \log \lambda_{pde} - \frac{\lambda_{pde}}{2} \sum_{j=1}^M r_{w_j}^2(\hat{\bs{a}},\bs{u}_{\bt_u}(e_{\bphi}(\hat{\bs{a}}))).
\ee 
The second term corresponds to the boundary conditions and is maximized when the discrepancy with the prescribed values is minimized. 
Given the likelihood in \refeqp{eq:likeg} and $q_{\bphi}$ in \refeqp{eq:q}, it can be written as:
\be 
\mathcal{F}_{bc}(\bt,\phi) = \frac{N_{bc}}{2}\log \lambda_{bc} - \frac{\lambda_{bc}}{2}\|\hat{\bs{g}}-\bs{u}_{g,\bt_u}(e_{\bphi}(\hat{\bs{a}}))\|^2_2,
\ee 
where $\bs{u}_{g,\bt_u}$ denotes the value of the solution according to \refeqp{eq:pub} at the boundary points considered. 

The third term pertains to the reconstruction error of the observed input coefficients $\hat{\bs{a}}$.  For the  form of the associated  in \refeqp{eq:cts} corresponding to the continuous-valued field, we have:
\be 
\mathcal{F}_{rec,a}(\bt,\bphi) = \frac{d_{a}}{2}\log\frac{\lambda_{rec}}{2\pi} - \frac{\lambda_{rec}}{2}\|\hat{\bs{a}}-\mu_{\bt_a}(e_{\bphi}(\hat{\bs{a}}))\|^2_2,
\ee 
For the piecewise-constant case of \refeqp{eq:pwc}, the term $\mathcal{F}_{rec,a}$ is given as follows:
\be 
\mathcal{F}_{rec,a}(\bt,\bphi) = \sum^{d_a}_{j=1}~z_j^{(i)} \log sig(\mu_{j,\bt_a}(e_{\bphi}(\hat{\bs{a}})) + (1-z_j^{(i)}) \log(1-sig(\mu_{j,\bt_a}(e_{\bphi}(\hat{\bs{a}}))).
\label{eq:lossbinarya}
\ee 
Finally, the last term $\mathcal{F}_{kl}$ acts as a regularizer by minimizing the KL divergence between the approximate posterior $q_{\bphi}$ and the prior $p_{\bt_\beta}(\bs{\beta})$. Since a uniform prior is used and a degenerate $q_{\bphi}$ as in \refeqp{eq:q}, this term is equal to a constant that does not affect the ELBO.

We note finally that derivatives with respect to the parameters $\bt$ (E-step) and $\bphi$ (M-step) can be efficiently computed using automatic differentiation tools. Incremental or randomized versions over the (virtual) data instances $N$ can alleviate the computational cost per iteration but necessitate the use of stochastic approximation schemes as discussed in the numerical illustrations  \cite{neal_view_1998}. The training procedure of the DGenNO framework is summarized in Algorithm~\ref{alg:foward}.
%
\begin{algorithm}[!t]
\caption{The training of the DGenNO}\label{alg:foward}
\begin{algorithmic}
\Inputs{Data $\{ \hat{\bm{a}}^{(i)}, \hat{\bm{g}}^{(i)}, \hat{\bm{R}}^{(i)}_M\}^{N}_{i=1}$, Parameters $\lambda_{pde}, \lambda_{bc}, \lambda_{rec}$}
\Initialize{Model parameters $\bmtheta=(\bmtheta_{\bm{a}}, \bmtheta_{\bm{u}}, \bmtheta_{\bmbeta})$ and $\bphi$, the learning rate $lr$.}
\While{Convergence or maximum number of iterations not reached}
\State {Compute $\bmbeta^{(i)}=e_{\bphi}(\hat{\bm{a}}^{(i)}),\ i=1,\cdots, N$} \hfill \refeqp{eq:q}
\State {
Compute $u^{(i)}=u_{\bt_u}(\bmbeta^{(i)}),\ i=1,\cdots, N$} \hfill \refeqp{eq:pub}
\State {Compute weighted residuals $$r_{j}(\hat{\bm{a}}^{(i)}, u^{(i)}),\ i=1,\cdots,N; j=1,\cdots,M.$$ 
}
\State {Calculate the ELBO $\mathcal{F}(\bmtheta,\bphi)$ according to Eq. \eqref{eq:loss} (see Eqs \ref{eq:loss_pde}-\ref{eq:lossbinarya})}
\State \textbf{M-step:} {Update parameter $\bphi$ with SGD while keeping $\bmtheta$ fixed:
$$\bphi \leftarrow \bphi + lr \odot \nabla_{\bphi} \mathcal{F}(\bmtheta,\bphi).$$}
\State \textbf{E-step:} {Update parameter $\bmtheta$ with SGD while keeping $\bphi$ fixed:
$$\bmtheta \leftarrow \bmtheta + lr \odot \nabla_{\bmtheta} \mathcal{F}(\bmtheta,\bphi).$$ }
\If{at every 2500th epoch}
\State {Update the learning rate to $lr \leftarrow lr/2$.}
\EndIf
\EndWhile
\end{algorithmic}
\end{algorithm}

\subsection{Predictions - Forward \& Inverse Problem}
\label{sec:predictions}
Once the model has been trained and the maximum likelihood estimate $\bt^*$ of the model parameters $\bt$ has been found as well as the optimal $\bphi^*$ for the variational approximation $q_{\bphi}$, the model can be used to provide {\em probabilistic} predictions of the PDE-solution $u$ for any (new) PDE-input $a$ (\textbf{forward problem}), but also of the PDE-input $a$ given some observations, say $\bs{u}_{obs}$, of the PDE-solution (\textbf{inverse problem}).

In particular, the predictive posterior density $p(u | \bs{a})$ for the \textbf{forward problem} is given by:
\be
p(u | \bs{a})  = \int p_{\bt_u^*}(u | \bs{\beta}) p_{\bt}(\bs{\beta} | \bs{a}) ~d\bs{\beta} \approx \int p_{\bt_u^*}(u | \bs{\beta}) ~q_{\bphi^*} ( \bs{\beta} | \bs{a}) ~d\bs{\beta},
\ee
where in the place of the actual posterior $p_{\bt}(\bs{\beta} | \bs{a})$ we used its variational approximation $q_{\bphi^*} ( \bs{\beta} | \bs{a})$. Hence samples of the predictive posterior can be readily obtained by:
\bi
\item sampling $\bs{\beta}$ from $q_{\bphi^*} ( \bs{\beta} | \bs{a})$ (in our case given by \refeqp{eq:q})
\item sampling $u$ from $p_{\bt_u^*}(u|\bs{\beta})$ 
 (in our case given by \refeqp{eq:pub}).
\ei

With regards to the \textbf{inverse problem} and given observations $\bs{u}_{obs}$ which relate to the solution $u$ through a likelihood $p(\bs{u}_{obs} | u)$, the predictive posterior density $p(a | \bs{u}_{obs})$ is given by\footnote{Given that $u$ is a function, we denote with $\mathcal{D}u$ the corresponding path integral \cite{alberts_physics-informed_2023}.}:
\be\label{eq:posterior_inv}
\begin{array}{ll}
p(a | \bs{u}_{obs}) & \propto \int p_{\bt} (a, u, \bs{\beta} , \bs{u}_{obs}) ~d\bs{\beta} ~\mathcal{D}u  \\
& = \int p_{\bt_a^*} (a | \bs{\beta} ) p( \bs{u}_{obs} | u) p_{\bt_u^*} (u |\bs{\beta} ) p_{\bt_{\beta}^*}(\bs{\beta}) ~d\bs{\beta} \mathcal{D}u \\
& =  \int p_{\bt_a^*} (a | \bs{\beta} ) p( \bs{u}_{obs} | u_{\bt_u^*}(\bs{\beta})) p_{\bt_{\beta}^*}(\bs{\beta}) ~d\bs{\beta} \\
& \propto \int p_{\bt_a^*} (a | \bs{\beta} ) p(\bs{\beta} | \bs{u}_{obs}) ~d\bs{\beta},
 \end{array}
  \ee
where the third equation is the result of \refeqp{eq:pub} and $p(\bs{\beta} | \bs{u}_{obs}) \propto p( \bs{u}_{obs} | u_{\bt_u^*}(\bs{\beta})) p_{\bt_{\beta}^*}(\bs{\beta})$ is the posterior of $\bs{\beta}$ given the observations $\bs{u}_{obs}$. We note that the latter can be readily sampled/approximated (e.g. using MCMC/SMC or VI) due to the fact that $\bs{\beta}$ is real-valued and $u_{\bt_u^*}(\bs{\beta})$ is a relatively inexpensive and differentiable function (see Section \ref{sec:latent_prior}).  Subsequently, the inferred $\bs{\beta}$ can be readily propagated through $p_{\bt_a^*} (a | \bs{\beta} )$ (see \refeqp{eq:cts} or \refeqp{eq:pwc}).

While the aforementioned procedure can yield very efficient predictions for the solution of the inverse problem, we have found that the accuracy can be significantly improved if, in addition to $\bs{u}_{obs}$, one conditions on weighted residuals, i.e. the virtual observables $\hat{\bs{R}}_M$ (as discussed in section \refeqp{sec:actual_virtual_data}) in order to obtain the predictive posterior $p(a | \bs{u}_{obs}, \hat{\bs{R}}_M)$ instead of just $p(a | \bs{u}_{obs})$ as in \refeqp{eq:posterior_inv}. \revise{These virtual observables encode additional physics-based constraints derived from the PDE itself. Rather than relying solely on a mismatch between $\bm{u}_{\text{obs}}$ and its prediction, in incorporating of $\hat{\bs{R}}_M$ ensures that any candidate solution $u$ (generated via a proposed $\bm{\beta}$) also satisfies the governing PDE. This joint conditioning on both $\bm{u}_{\text{obs}}$ and $\hat{\bs{R}}_M$ strengthens the inversion and regularizes the posterior, particularly in settings with limited or noisy observations.} We note that the introduction of residuals implies an increase in the computational cost which we discuss (in relation to competitors) in section \ref{sec:experiments}.

Analogously to \refeqp{eq:posterior_inv} we express the sought density as:
\be
p(a | \bs{u}_{obs}, \hat{\bs{R}}_M) \propto \int p_{\bt_a^*} (a | \bs{\beta}) p(\bs{\beta} | \bs{u}_{obs}, \hat{\bs{R}}_M) ~d\bs{\beta}.
\label{eq:postinvexp}
\ee
While the first density in the integrand is as before, we approximate the second as follows:
\be
\begin{array}{ll}
p(\bs{\beta} | \bs{u}_{obs}, \hat{\bs{R}}_M) & \propto \int p_{\bt}(\bs{a},u,\bs{\beta}, \bs{u}_{obs}, \hat{\bs{R}}_M)  ~ \mathcal{D}u ~d\bs{a} \\
 & = \int p(\bs{u}_{obs} | u) p(\hat{\bs{R}}_M | u,\bs{a}) p_{\bt^*_u}(u |\bs{\beta})  p_{\bt^*_a}(\bs{a} |\bs{\beta}) p_{\bt_{\beta}^*}(\bs{\beta})~ \mathcal{D}u~ d\bs{a} \\
& = \int  p( \bs{u}_{obs} | u_{\bt_u^*}(\bs{\beta}))  p(\hat{\bs{R}}_M | u_{\bt_u^*}(\bs{\beta}) ,\bs{a})  p_{\bt^*_a}(\bs{a} |\bs{\beta}) p_{\bt_{\beta}^*}(\bs{\beta}) ~d\bs{a} \\
& = p( \bs{u}_{obs} | u_{\bt_u^*}(\bs{\beta})) p_{\bt_{\beta}^*}(\bs{\beta}) ~\underbrace{ \int p(\hat{\bs{R}}_M | u_{\bt_u^*}(\bs{\beta}) ,\bs{a})  p_{\bt^*_a}(\bs{a} |\bs{\beta}) }_{p(\hat{\bs{R}}_M | \bs{\beta})} ~d\bs{a},
\end{array}
\label{eq:postbexp}
\ee
where, as before, the third equation is the result of \refeqp{eq:pub}.
We approximate the integral with respect to $\bs{a}$ for $p(\hat{\bs{R}}_M | \bs{\beta})$ with Monte Carlo as:
\be
p(\hat{\bs{R}}_M | \bs{\beta})=\int p(\hat{\bs{R}}_M | u_{\bt_u^*}(\bs{\beta}) ,\bs{a})  p_{\bt^*_a}(\bs{a} |\bs{\beta})  ~d\bs{a} \approx \frac{1}{K} \sum^{K}_{k=1} p(\hat{\bs{R}}_M | u_{\bt_u^*}(\bs{\beta}) ,\bs{a}^{(k)}),
\ee
where $\bs{a}^{(k)}$ are sampled from $p_{\bt^*_a}(\bs{a} |\bs{\beta})$. We have found that using a small $K$ (e.g.,  $K = 5$) is sufficient for a good approximation\footnote{\revise{We note that this selection is analogous to setting a mini-batch size in SGD: while the optimal value may vary slightly with problem characteristics, a moderate default value (like $K=5$ in our case) generally works well across a range of tasks.}}. With the help of this approximation, we obtain an  (approximate) expression for $p(\bs{\beta} | \bs{u}_{obs}, \hat{\bs{R}}_M)$ based on \refeqp{eq:postbexp} which can be readily evaluated. In combination with \refeqp{eq:postinvexp}, it suggests the following procedure for providing predictive samples for the inverse problem:
\bi
\item sample $\bs{\beta}$ from $p(\bs{\beta} | \bs{u}_{obs}, \hat{\bs{R}}_M)$ (in our experiments we achieved this efficiently using Variational Inference),
\item sample $a$ (at any location in the problem domain $\Omega$) from  $p_{\bt_a^*} (a | \bs{\beta} )$ (see \refeqp{eq:cts} or \refeqp{eq:pwc}).
\ei

\section{Numerical Experiments}
\label{sec:experiments}
In this section, we assess the efficacy and robustness of our proposed framework through a series of benchmark numerical experiments involving forward and inverse problems for parametric PDEs. To assess predictive accuracy, we compute the Relative Mean Squared Error (RMSE) on test data. Particularly, in solving forward problems, we compare the proposed DGenNO method with:
\bi 
\item PI-DeepONet  \cite{jiao2024solving}. This choice is motivated by the following factors: Firstly, PI-DeepONets is one of the most widely adopted neural operator methods, demonstrating superior performance across a broad spectrum of PDE problems. Secondly, it is highly versatile, as it can handle a wide range of PDEs and calculate derivatives using automatic differentiation (AD), which makes it well-suited for many complex scenarios such as irregular geometries and non-uniform meshes that are difficult to address with methods like PINO and PI-WNO. Additionally, the PI-DeepONet formulates its PDE loss using strong-form (i.e., collocation-type) residuals while the proposed framework employs weighted residuals. This makes PI-DeepONet an ideal baseline to highlight the benefits of the proposed novel neural operator architecture and the advantages of using weak-form residuals over strong-form ones.

\item  PI-MultiONet. This is a variation of the PI-DeepONet, which employs the novel architecture proposed in Section \ref{sec:problem} while maintaining \textbf{the same number of trainable parameters as the DeepONet} architecture. It is trained using exactly the same collocation-type residuals. The goal is to assess on the same data/residuals, the  benefits of  the proposed MultiONet architecture. 
\ei
\revise{In solving inverse problems, currently, there is no direct extension or implementation of PI-DeepONet that can recover high-dimensional or functional inputs $a$ from sparse and noisy observations of the solution $u$. However, as PI-DeepONet allows for gradient computation with respect to continuous inputs, we adapted it for solving the inverse problems with continuous coefficients following the gradient-based strategy described in Section \ref{sec:predictions} (see \eqref{eq:posterior_inv}): a neural operator $\mathcal{G}(a)$ is first trained to approximate the forward map, ideally generalizing across all admissible inputs $a$. In the inverse setting, this surrogate is then used to infer the input $a$ by minimizing the discrepancy between the predicted output with $\mathcal{G}(a)$ and the observed data. However, in solving inverse problems with discrete-valued coefficients, this adaptation is no longer applicable for PI-DeepONet because the derivatives with such $a$ are not defined.
Therefore, the PI-DeepONet is not available as a baseline method in this case. Instead, we compare the proposed DGenNO method with: the PINN method \cite{raissi2019physics} and the ParticleWNN method \cite{zang2023particlewnn}. While these two methods are not operator learning methods, they are widely used for solving forward and inverse problems in a purely physics-driven fashion—i.e., without relying on labeled data. This aligns with the core philosophy of our proposed framework.}

The specific model architectures and parameter settings for each method are provided in \ref{sec:network}. For training, we generate $N = 1000$ samples of input coefficients from predefined distributions (provided in the respective problem sections). As mentioned earlier, we do not rely on labeled data, i.e., we never solve the governing PDE to generate training data, nor do we solve it during training.
In evaluating the weighted residuals, we did not employ any specialized strategies beyond those used in existing physics-informed deep learning methods. Specifically, for the PI-DeepONet and PI-MultiONet methods, collocation points were selected using strategies similar to those adopted in the PINN framework. For the DGenNO method, particles were generated following the same procedure as in the ParticleWNN method. Moreover, for the weighting functions, we consistently used Wendland’s CSRBFs in our framework across all problems, consistent with the ParticleWNN approach.
Unless stated otherwise, we use the ADAM optimizer with an initial learning rate of $lr = 10^{-3}$. The learning rate is reduced by a factor of two every $2500$ epochs. The batch size is set to $50$, and training continues for $10,000$ epochs to ensure convergence. To provide a fair comparison, all experiments are conducted on the same hardware, i.e. on a 64-core AMD Ryzen CPU equipped with an RTX 4090 GPU.

\subsection{Darcy's flow with piecewise constant coefficients}
\label{sec:darcy_flow}
In the first example, we consider  Darcy’s flow in a 2D domain. The governing equation can be written as:
\be\label{eq:darcy_flow}
\begin{split}
-\nabla \cdot (a(x_1,x_2) \nabla u(x_1,x_2)) &= f(x_1,x_2), \quad (x_1,x_2) \in \Omega = [0,1]^2, \\
u(x_1,x_2) &= 0, \quad (x_1,x_2)\in \partial\Omega,
\end{split}
\ee
where $a$ is the permeability field, $u$ is the pressure field, and $f$ is a source term that is set as a fixed constant, i.e., $f = 10$. For this forward problem, we are interested in learning the mapping from the permeability field $a(x_1,x_2)$ to the pressure field $u(x_1,x_2)$, i.e., $\Gcal: a(x_1,x_2) \rightarrow u(x_1,x_2)$.
For the permeability field $a(x_1,x_2)$, we consider piecewise constant functions generated using a cutoff Gaussian Process $\mathcal{GP} (0,(-\Delta + 9I)^{-2})$ (\cite{li2020fourier,goswami2022physics}). In particular, we set where $a(\bm{x}) = 10$ if the underlying GP-value is greater than $0$ and $a(\bm{x}) = 5$  otherwise. This problem poses significant challenges for physics-informed methods that employ collocation-type residuals due to the discontinuities in $a(\bm{x})$ which can give rise to big errors when approximating derivatives. In contrast, the proposed framework employs weighted residuals (see \ref{sec:weak_residual}) in which $a(\bx)$ and not its derivatives appear. 
For the proposed DGenNO method, we set the number of weighted residuals $M = 300$ and the number of integration points $N_{int} = 25$ for numerical computation of integrals appearing in the residuals as in \cite{zang2023particlewnn}. For the PI-DeepONet, the derivatives of the input coefficient function are approximated numerically with finite differences.
The sensors $\Xi$ correspond to a regular $29\times 29$ grid on the domain $\Omega$. Therefore, the input of the branch network in the PI-DeepONet is a (binary) vector of dimension $29^2=841$.

The values of the input coefficient field at the same locations constitute the training dataset $ \{ \hat{\bs{a}}^{(i)} \}_{i=1}^{N}$ (see \refeqp{eq:pwc}) employed by DGenNO.
Furthermore,  we employed a $128-$dimensional vector for the latent generators, i.e.  $\bm{\beta} \in \mathbb{R}^{128}$. We enforce the Dirichlet boundary condition a priori as discussed in Section \ref{sec:actual_virtual_data}\footnote{Specifically, we employ  $\psi(x_1,x_2)=\sin(\pi x_1)\sin(\pi x_2)$ and $g(x_1,x_2)=0$ which allows us to omit the BC from the observables.}. We set the hyperparameter corresponding to the PDE loss $\lambda_{pde} = 1$ for all methods. In DGenNO, we apply a recovery loss weight of $\lambda_{rec}=0.25$ corresponding to the input $a$.

\begin{figure}[!htbp]
    \centering  
    \subfigure[In-distribution sampled coefficient filed $a$ and PDE solution $u$]{\label{fig:darcy_in}
        \includegraphics[width=0.8\textwidth]{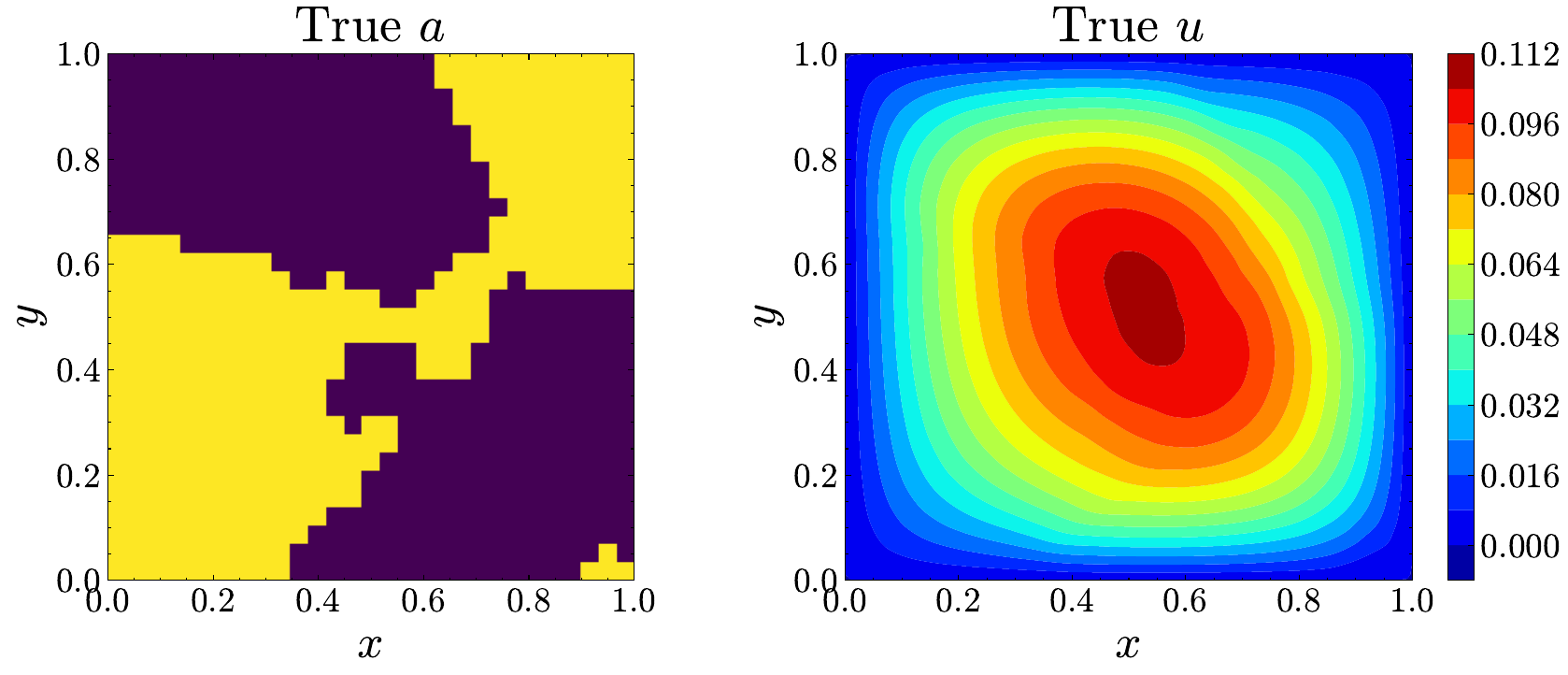}}
    \subfigure[Predicted solution $u$]{\label{fig:darcy_u_in}
        \includegraphics[width=0.99\textwidth]{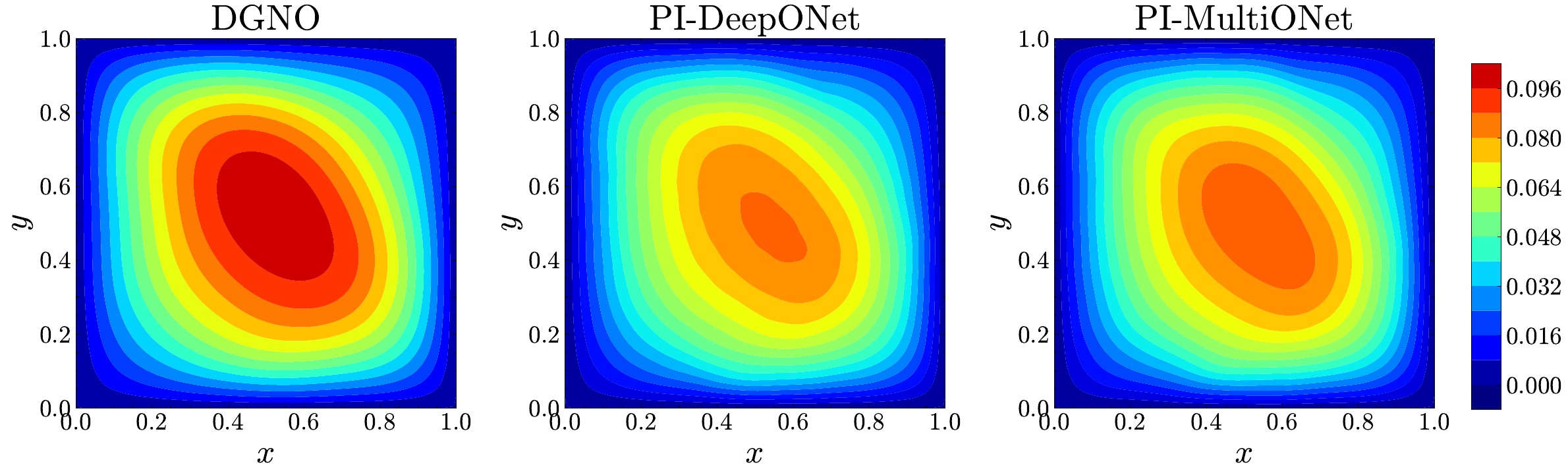}}
    \subfigure[Pointwise absolute errors]{\label{fig:darcy_uabs_in}
        \includegraphics[width=0.99\textwidth]{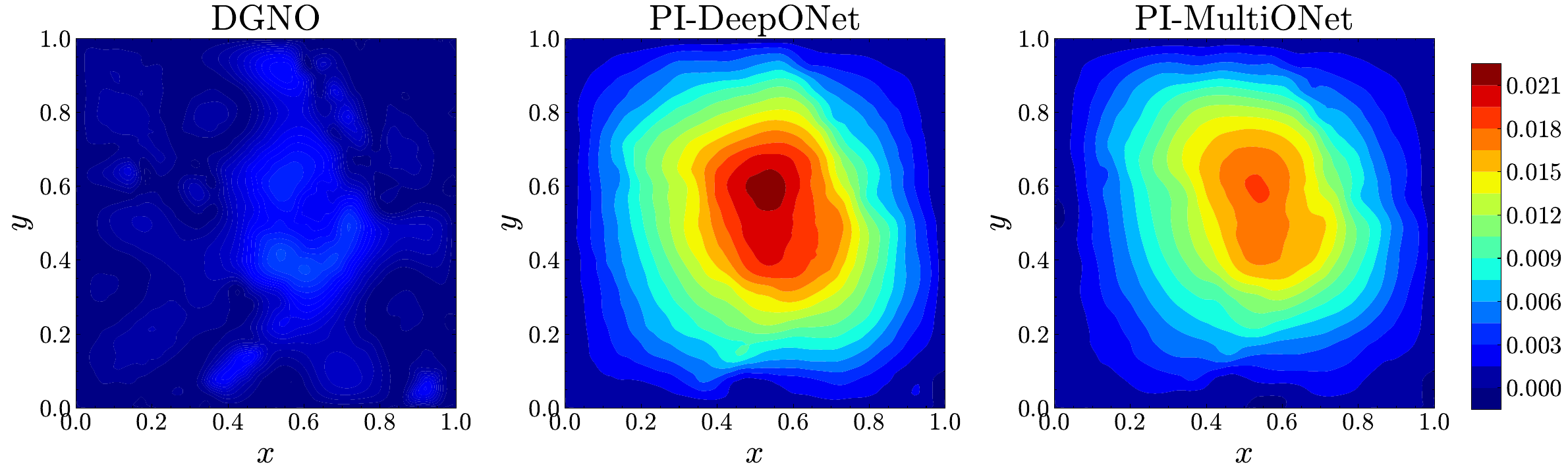}}
    \vspace{-0.25cm}
    \caption{Indicative in-distribution test-case for the Darcy-flow equation.}
    \label{fig:darcy_pwc_in}
\end{figure}

\begin{figure}[!htbp]
    \centering  
    \subfigure[Out-of-distribution sampled coefficient filed $a$ and PDE solution $u$]{\label{fig:darcy_out}
        \includegraphics[width=0.8\textwidth]{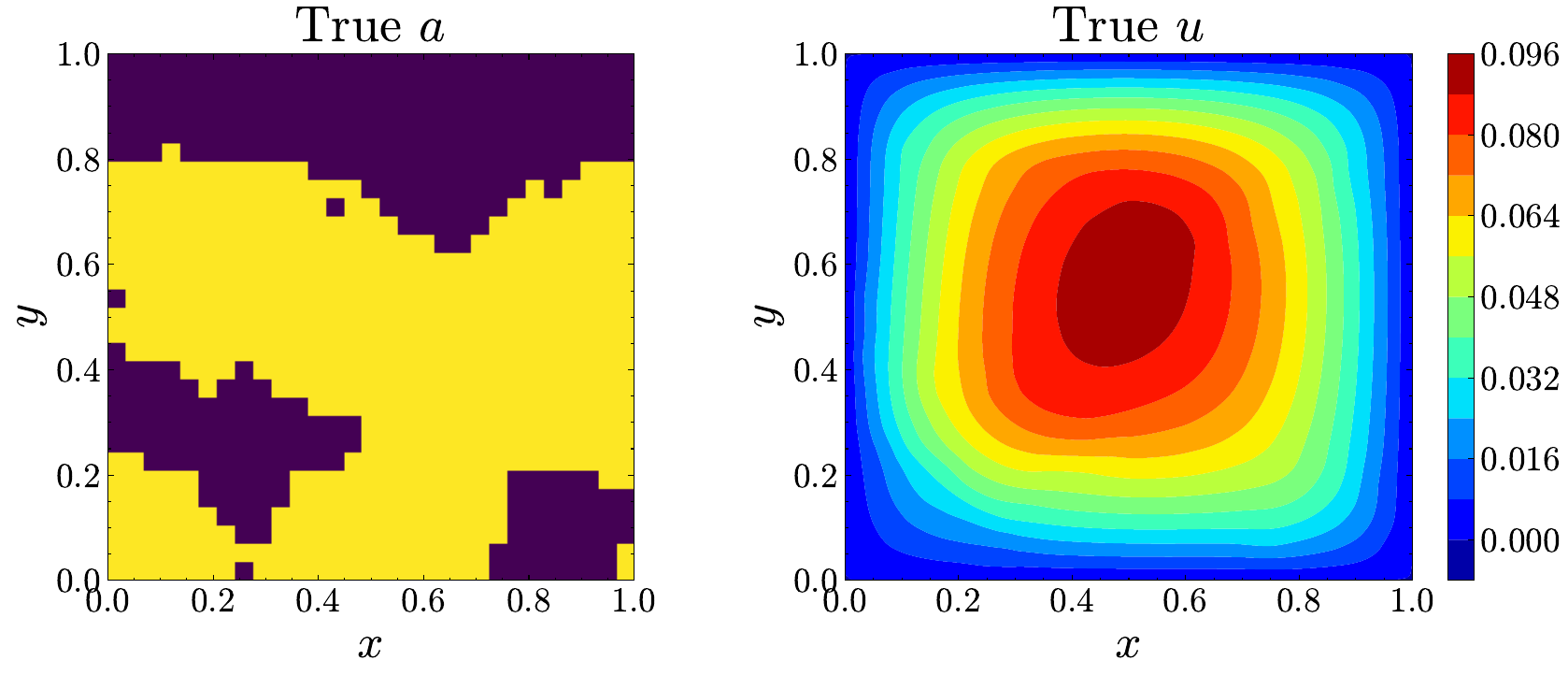}}
    \subfigure[Predicted solution $u$]{\label{fig:darcy_u_out}
        \includegraphics[width=0.99\textwidth]{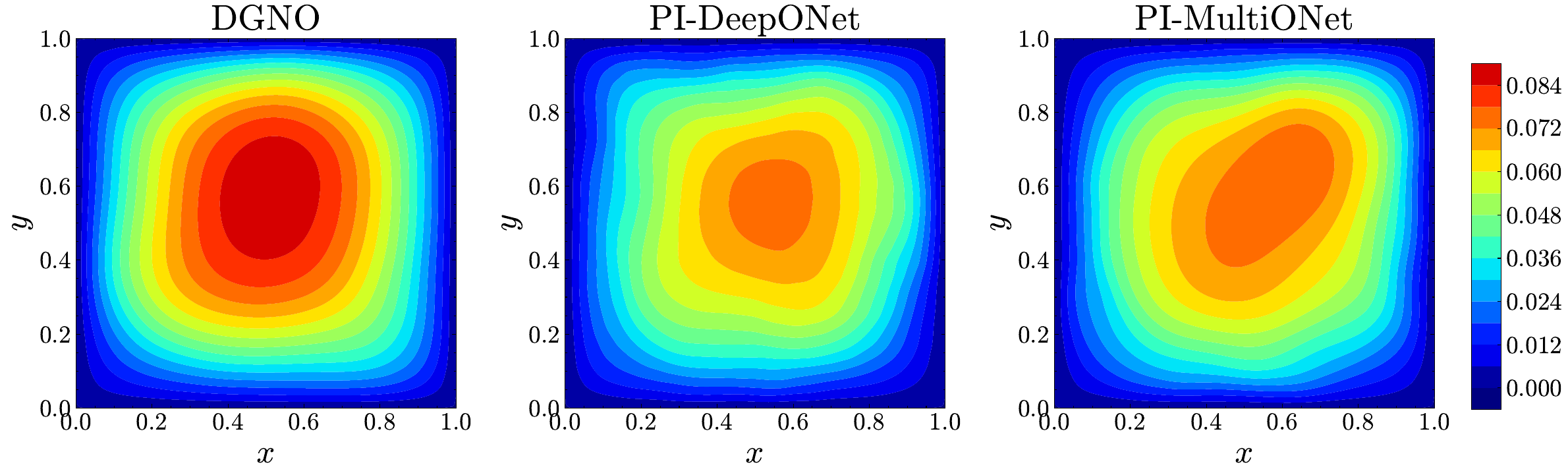}}
    \subfigure[Pointwise absolute errors]{\label{fig:darcy_uabs_out}
        \includegraphics[width=0.99\textwidth]{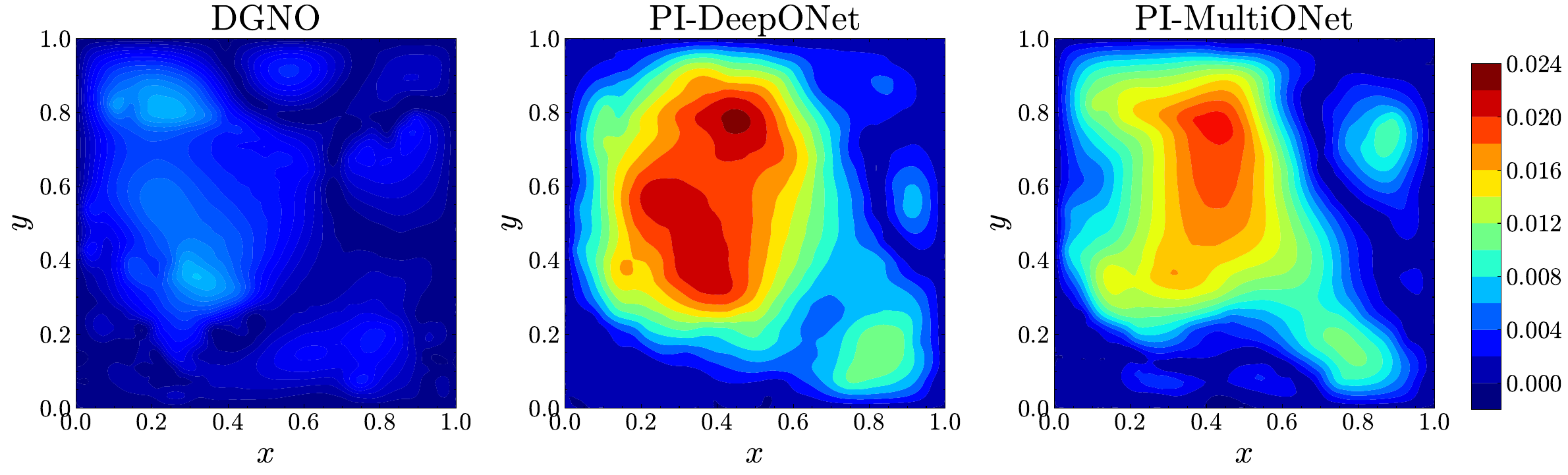}}
    \vspace{-0.25cm}
    \caption{Indicative out-of-distribution test-case for the Darcy-flow equation.}
    \label{fig:darcy_pwc_out}
\end{figure}

To assess the performance of the aforementioned methods, we generated two test datasets:
\bi 
\item an \textbf{in-distribution} test dataset by sampling $200$ coefficient fields $a$ from the same distribution used in the training data as described above, and,
\item an \textbf{out-of-distribution} test dataset by sampling coefficient fields $a$ from a zero-cutoff $GP(0, (-\Delta + 16I)^{-2})$. While the values of the coefficient field are the same, the second and higher-order correlation functions of $a(\bs{x})$ are different.
\ei
We generated 200 samples for each dataset and computed the corresponding ground truth PDE solutions $u$ using the FEM on a uniform $29\times29$ mesh. Table \ref{tab:darcy_pwc} presents the RMSE values and prediction times for each dataset. As observed, the proposed DGenNO framework achieves significantly lower RMSE compared to both PI-DeepONet and PI-MultiONet. Additionally, Figure \ref{fig:darcy_uabs_in} and \ref{fig:darcy_uabs_out} illustrate that the solutions predicted by PI-DeepONet and PI-MultiONet exhibit larger point-wise absolute errors than the proposed DGenNO method. This highlights the challenges strong-form residual methods face when handling discontinuous coefficients, further demonstrating the superiority of our proposed framework. Moreover, the RMSE results indicate that PI-MultiONet outperforms PI-DeepONet while being trained with the same data and loss, highlighting the effectiveness of the MultiONet architecture. 

In Figures \ref{fig:darcy_pwc_in} and \ref{fig:darcy_pwc_out}, we also present indicative samples from each dataset and plot the predicted solutions $u$ obtained by each of the three methods discussed.
Although the performance of the proposed framework degrades on the out-of-distribution test dataset, it still significantly outperforms PI-DeepONet and PI-MultiONet. This is particularly evident in Figure \ref{fig:darcy_uabs_out}, which illustrates the absolute errors in the PDE solution that, in part at least, can be attributed to the challenges that collocation-residual-based methods face when handling piecewise constant input coefficient fields.

We also conducted numerical experiments for a nonlinear version of the governing equation. The results contained in \ref{sec:nonlinear} reinforce the observations made for the linear PDE above and indicate the versatility of the proposed framework, since the only change required relates to the module computing weighted residuals.
\begin{table}[tb]\small
\centering
\begin{tabular}{c|c|c|cc} \bottomrule
                    {} & DGenNO & PI-DeepONet  & PI-MultiONet \\ \hline
    {RMSE (in)} & \textbf{$2.88e^{-2}\pm1.25e^{-2}$} & $0.175\pm 0.054$  & $0.159\pm 0.053$ \\
    {RMSE (out)} & \textbf{$5.35e^{-2}\pm2.19e^{-2}$} & $0.191\pm 0.069$ & $0.184\pm 0.055$ \\
    {Time(s)} & $4.49e^{-3}\pm9.08e^{-4}$ & \textbf{$1.30e^{-3}\pm9.61e^{-4}$} & $3.16e^{-3}\pm9.35e^{-4}$ \\\toprule
\end{tabular}
\caption{Performance of each method in solving the Darcy's problem \eqref{eq:darcy_flow}: RMSE(in): the RMSE of each method in the in-distribution dataset; RMSE(out): the RMSE of each method in the out-of-distribution dataset; Time(s): time consumption of each method to predict numerical solution}
\label{tab:darcy_pwc}
\end{table}

\subsubsection{The role of the latent generators $\beta$}
\begin{figure}[!htbp]
    \centering  
    \subfigure{\label{fig:darcy_betaInterp}
        \includegraphics[width=1.\textwidth]{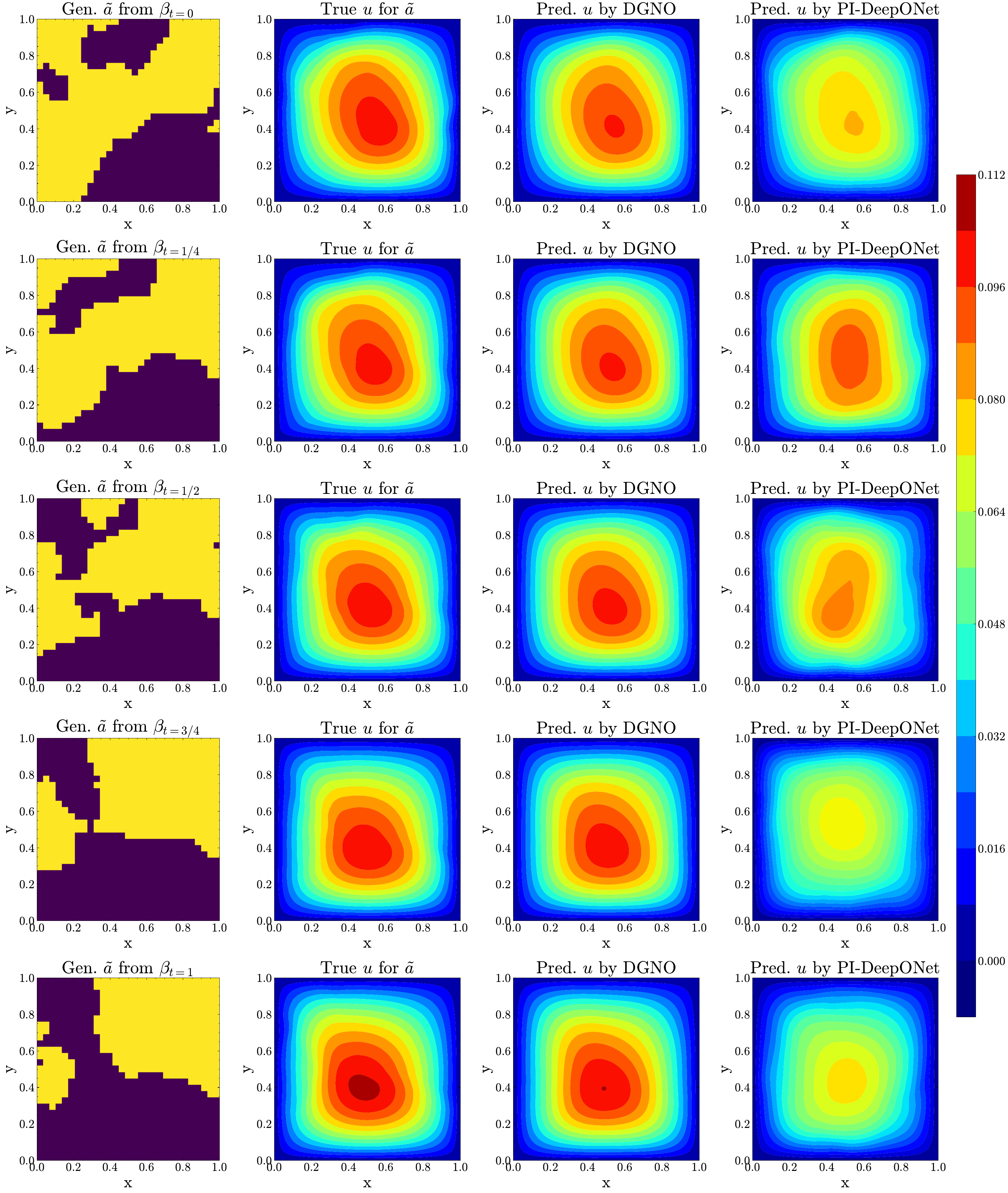}}
    \vspace{-0.25cm}
    \caption{For $\bm{\beta}_t = t\bm{\beta}_0 + (1-t)\bm{\beta}_1$, we predict the corresponding $a$ (denoted as Gen. $\tilde{a}$ in the first column) and solution $u$ (third column) using DGenNO. The second column depicts the true $u$ obtained by FEM and the fourth the $u$ predicted by PI-DeepONet. The RMSEs (from top to bottom) of DGenNO are $0.039$, $0.033$, $0.028$, $0.027$, and $0.026$, and of PI-DeepONet are $0.187$, $0.084$, $0.171$, $0.266$, and $0.270$, respectively.} 
    \label{fig:darcy_pwc_interp}
\end{figure}

To elucidate the role of the generative, latent variables $\bs{\beta}$ in learning representations for both PDE-inputs and -outputs, we conducted the following experiment:
\bi
\item we randomly sampled two coefficients $a_0,a_1$ (from the in-distribution test dataset) and computed their corresponding latent representations $\bs{\beta}_0$ and $\bs{\beta}_1$  using the trained encoder $q_{\bphi^*}(\bs{\beta}|\bs{a})$.
\item We linearly interpolated  between $\bs{\beta}_0$ and $\bs{\beta}_1$,  i.e.  we considered the line  $\bs{\beta}_t = t \cdot \bs{\beta}_0 + (1-t) \cdot \bs{\beta}_1$ and specifically the $\bm{\beta}$'s corresponding to    $t = 0, \frac{1}{4}, \frac{1}{2}, \frac{3}{4}, 1$, in order to obtain five points in $\bs{\beta}$-space. 
\item For these five points, we employ the trained decoders $p_{\bt^*_a}(a|\bs{\beta})$, $p_{\bt^*_u}(u|\bs{\beta})$ in the proposed DGenNO framework to generate the five corresponding coefficients $a$ (denoted as $\tilde{a}$) and PDE-solutions $u$  shown in the first and third columns of Figure \ref{fig:darcy_pwc_interp} respectively. Reference PDE-solutions obtained using FEM are displayed in the second column and those predicted by PI-DeepONet in the fourth.
\ei
As seen in Figure \ref{fig:darcy_pwc_interp}, DGenNO’s predicted solutions closely match the reference, achieving RMSEs of $0.039, 0.033, 0.028, 0.027, 0.026$, compared to significantly higher errors of $0.187, 0.084, 0.171, 0.266, 0.270$ for PI-DeepONet. More importantly, the latent variables $\bs{\beta}$ are able to simultaneously encode the PDE-input and -output, providing an alternative link in the relationship between $a$ and $u$ that would prove extremely useful in solving inverse, in addition to forward, problems (see section \ref{sec:inverse}). 

\subsubsection{Performance with labeled data}
Although our primary focus is on evaluating the proposed DGenNO method in a purely physics-driven setting for solving parametric PDEs and inverse problems, we emphasize that the DGenNO framework is flexible and can seamlessly incorporate any number of labeled training pairs to enhance performance and training efficiency. To demonstrate this capability, we revisit the problem in Eq.~\eqref{eq:darcy_flow} using the same experimental setup as before, with one key modification: a subset of the training data is now assumed to be labeled. Specifically, among the 1,000 input samples $a$ in the training dataset, we randomly select 200 instances for which the corresponding solutions $u$ are computed using the finite element method (FEM). As a result, paired input–output data $(a, u)$ are available for these 200 cases, while the remaining 800 training samples include only the input $a$ without the corresponding solution $u$.

After training the DGenNO framework using this hybrid dataset, which includes both labeled and unlabeled samples, we evaluate its performance on both in-distribution and out-of-distribution testing sets. The RMSEs achieved are $1.61e^{-2} \pm 1.04e^{-2}$ and $3.82e^{-2} \pm 1.71e^{-2}$, respectively. Compared to those obtained without any labeled training pairs (see Table \ref{tab:darcy_pwc}), these results demonstrate a clear improvement in predictive accuracy across both settings, indicating that incorporating even a limited number of labeled instances can enhance the performance of DGenNO in a physics-driven learning context.

\subsubsection{Performance on extended out-of-distribution testing dataset}
\label{sec:extented_out}
We further investigate the generalization capabilities of the proposed DGenNO method by testing it on a more challenging out-of-distribution dataset, where the Gaussian Process used to generate the coefficient fields $a$ includes non-zero means. Specifically, we create an out-of-distribution test set by sampling 200 coefficient fields from a zero-cutoff Gaussian Process $GP(\mu, (\Delta + 16I)^{-2})$, where the mean $\mu$ is drawn randomly from a range $[\mu_{lb},\mu_{ub}]$ for each instance. It is important to note that the coefficient field is generated by applying a cutoff to the values of the GP image (i.e., $a(\bm{x}) = 10$ if the GP value is greater than $0$, and $a(\bm{x}) = 5$ otherwise). A large absolute value of $\mu$ would result in more homogeneous fields, where $a(\bm{x})$ is either $10$ or $5$ throughout most of the domain. To mitigate this and ensure more diverse coefficient fields, we sample $\mu$ from the interval $[-0.2, 0.2]$.

The RMSE achieved by the DGenNO method on this out-of-distribution dataset is $6.90e^{-2}\pm 3.19e^{-2}$ which is comparable to the value reported in Table  \ref{tab:darcy_pwc}. For comparison, the RMSEs obtained by PI-DeepONet and PI-MultiONet are $0.204 \pm 0.078$ and $0.190 \pm 0.057$, respectively.

\subsection{Burgers' equation}
The second experiment involves the time-dependent, viscous Burgers' equation in dimension one. This problem, which has been frequently used as a standard benchmark to evaluate the effectiveness of neural operators, involves the governing equation:
\be\label{eq:burgers}
\begin{split}
u_t + uu_x &= \nu u_{xx}, \quad (x,t) \in \Omega_T = [-1,1] \times (0,1], \\
u(x,0) & = a(x), \\
\end{split}
\ee
In our experiments, the viscosity term is set to $\nu = 0.1/\pi$, and zero boundary conditions are applied. The PDE-input, in this case, corresponds to the initial condition $a(x)$ and our goal is to learn its dependence on the time-dependent PDE-solution $u(x,t)$.

We generate {\em unlabeled} training data $\hat{\bs{a}}^{(i)}$ (see \refeqp{eq:cts}) by sampling a $GP(0, 49^2(-\Delta + 49I)^{-2})$ over $128$ locations on a uniform grid in the interval $[-1,1]$. The latter serve as the sensors $\Xi = \{\xi_i\}_{i=1}^{128}$ employed in the PI-DeepONet. The values of the field $a$ at these points serve as the observables $\hat{\bs{a}}$ and as the inputs in the encoder of \refeqp{eq:q}. For DGenNO, we employ a $64$-dimensional vector of latent variables, i.e. $\bm{\beta}\in\mathbb{R}^{d_\beta=64}$ (for further details, see \ref{sec:burger}).
Furthermore, we evaluate the PDE loss using $M=100$ weighted residuals (see Equation \ref{eq:weak_residual_burgers}) which are computed using $N_{int} = 10$ integration points. The latter also serves as the collocation points,  totaling $1000$, for the strong-form residuals in PI-DeepONet. We enforce the boundary conditions a priori as discussed in Section \ref{sec:actual_virtual_data}\footnote{Specifically, we employ  $\psi(x)=\sin(\frac{\pi}{2}(x  +1))$ and $g(x)=0$ which allows us to omit the BC from the observables.}.
We set the hyperparameter $\lambda_{ic}=10$ for initial condition enforcement across all methods. Notably, in this problem, the recovery loss in DGenNO coincides with the initial condition loss. Finally, we set the hyperparameter corresponding to the PDE loss $\lambda_{pde} = 1$ for all methods.

As in the previous example, we assess the predictive performance on two test datasets consisting of $200$ samples of $a$ each. In particular:
\bi
\item an \textbf{in-distribution} dataset obtained by sampling $a$ from the same Gaussian Process $GP (0, 49^2(-\Delta + 49I)^{-2})$ as was done for the training data, and,
\item an \textbf{out-of-distribution} dataset obtained by sampling $a$ from the  $GP (0, 36^2(-\Delta + 36I)^{-2})$.
\ei
Reference solutions $u(x,t)$ for each of these $a$ functions are obtained with the Chebfun Package \cite{driscoll2014chebfun} on a rectangular spatiotemporal grid of size $636\times101$. The RMSEs and computational times for predicting the solution using the three methods, i.e., DGenNO, PI-DeepONet, and PI-MultiONet, are contained in Table \ref{tab:burgers}. 
We observe that the proposed DGenNO method, as well as PI-MultiONet, achieves RMSE values that are smaller than those obtained with PI-DeepONet, whereas the first two are roughly equivalent. Hence, the MultiONet architecture proposed exhibits superior performance as compared to DeepONet, but the generative model proposed in DGenNO does not appear to offer an advantage. These conclusions hold for both the in- and out-of-distribution test datasets, despite the degradation in accuracy for the latter one. We note, nevertheless, that the computational time is shorter for PI-DeepONet in comparison to the other two. This is mainly because the MultiONet architecture needs to compute the inner product of the outputs from multiple hidden layers in the trunk and branch networks, whereas the DeepONet architecture only computes the inner product for the last layers. Therefore, the time used by the proposed DGenNO and the PI-MultiONet is similar since both methods share the same neural-operator architecture.

\begin{figure}[!htbp]
    \centering 
    \subfigure[In-distribution sampled coefficient filed $a$ and PDE solution $u$]{
        \includegraphics[width=0.8\textwidth]{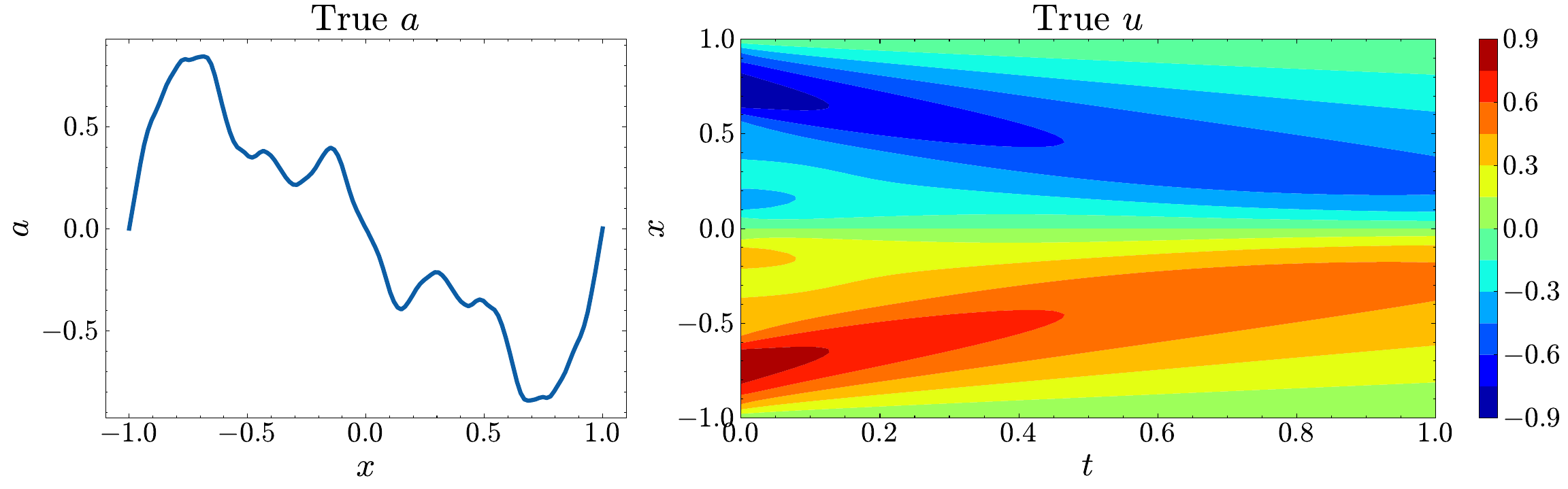}}
    \subfigure[DGenNO]{\label{fig:DGenNO_in}
        \includegraphics[width=1.\textwidth]{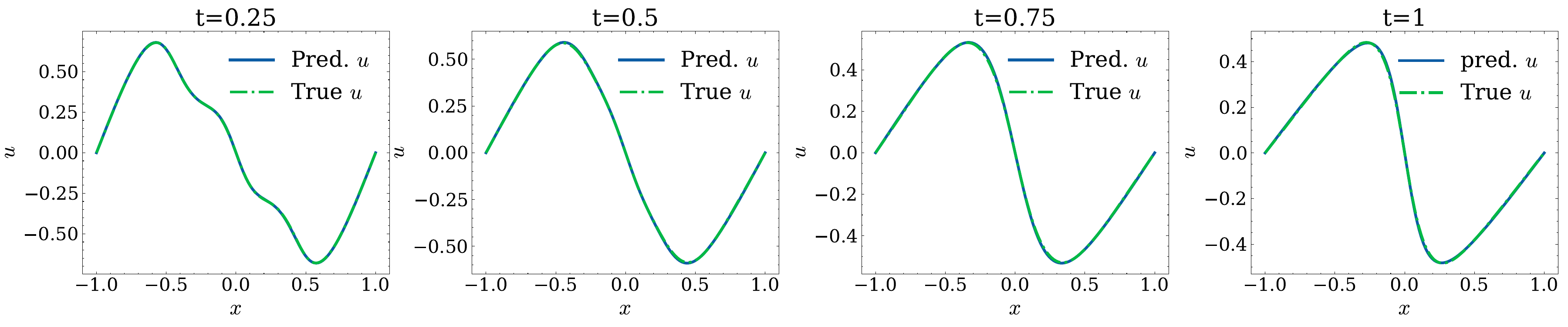}}
    \subfigure[PI-DeepONet]{\label{fig:pideeponet_in}
        \includegraphics[width=1.\textwidth]{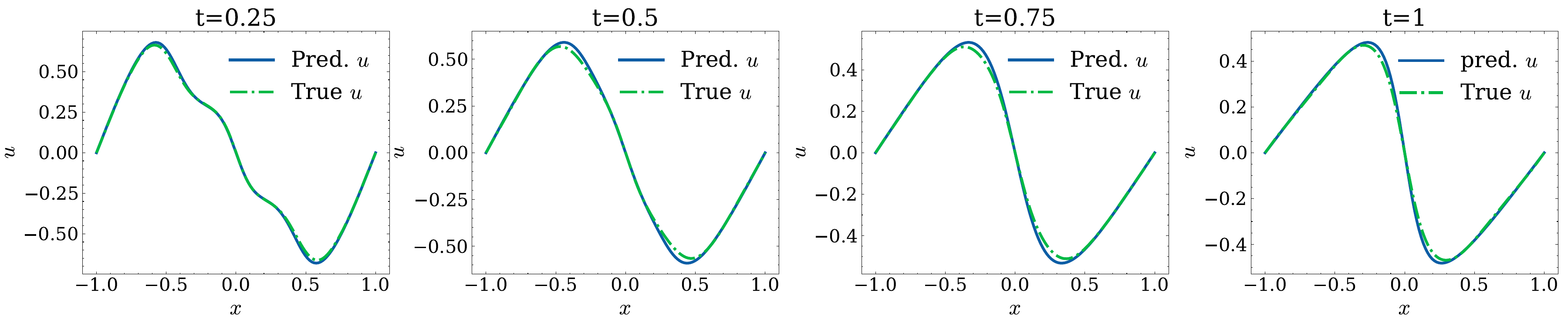}}
    \subfigure[PI-MultiONet]{\label{fig:pimultionet_in}
        \includegraphics[width=1.\textwidth]{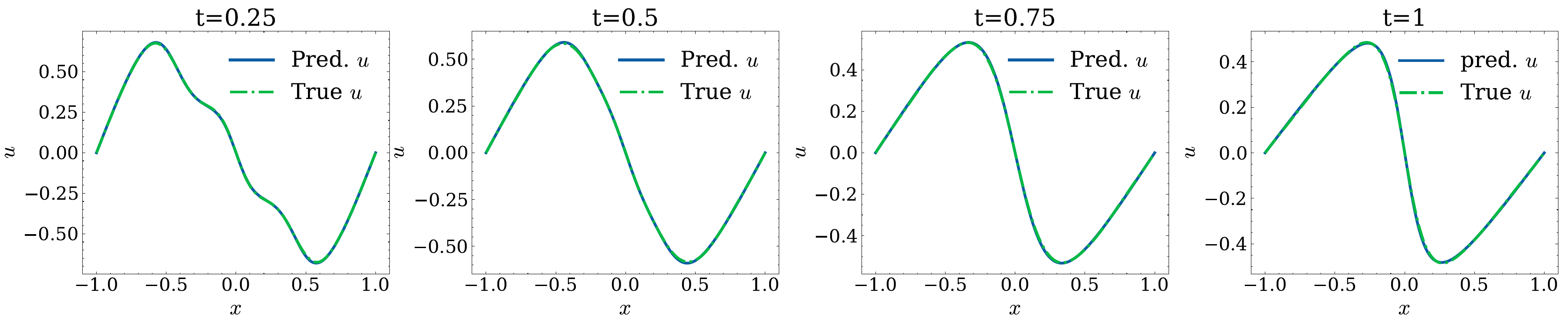}}
    \vspace{-0.25cm}
    \caption{Predicted solution profiles of the Burgers' equation at $t=0.25s, 0.5s, 0.75s, 1s$ obtained with the three competitive methods for an indicative initial condition from the \textbf{in-distribution} dataset.
    } 
    \label{fig:burgers_in}
\end{figure}

\begin{figure}[!htbp]
    \centering      
    \subfigure[Out-of-distribution sampled coefficient filed $a$ and PDE solution $u$]{
        \includegraphics[width=0.8\textwidth]{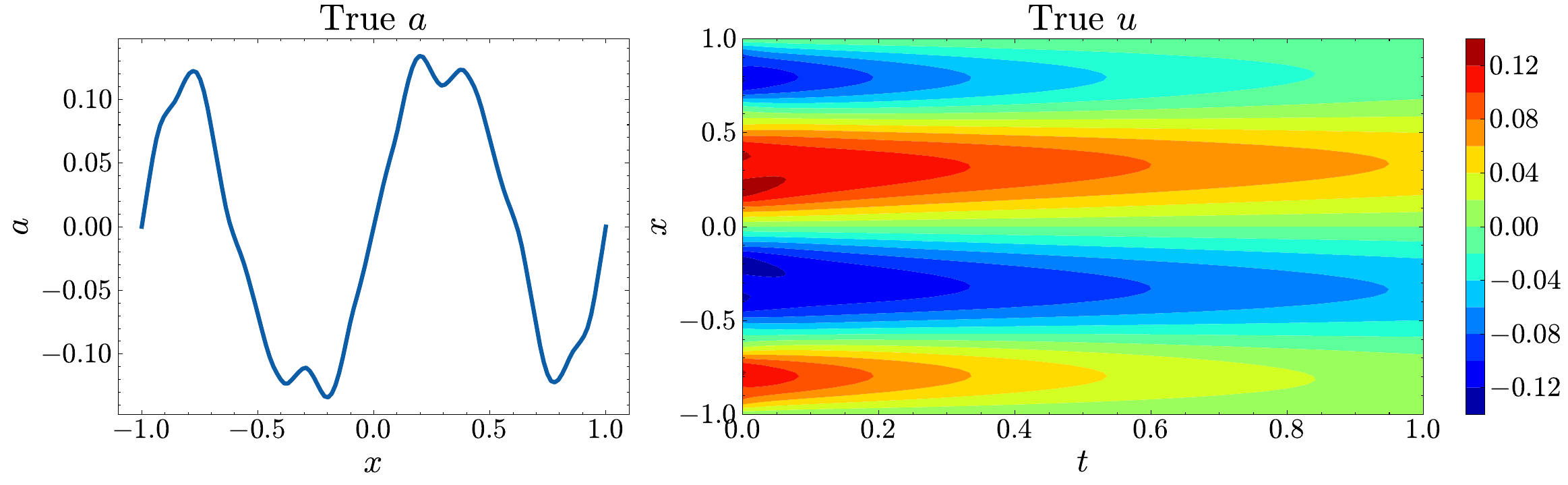}}
    \subfigure[DGenNO]{\label{fig:DGenNO_out}
        \includegraphics[width=1.\textwidth]{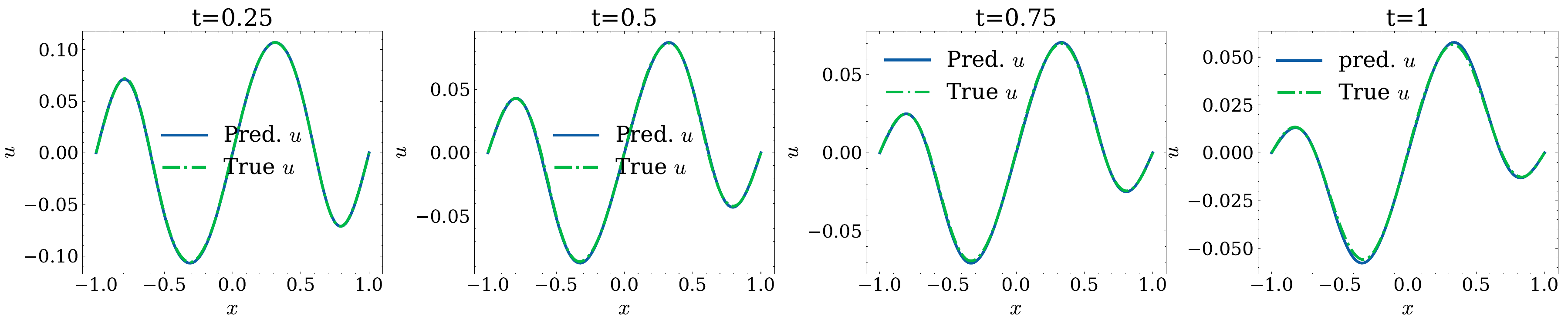}}
    \subfigure[PI-DeepONet]{\label{fig:pideeponet_out}
        \includegraphics[width=1.\textwidth]{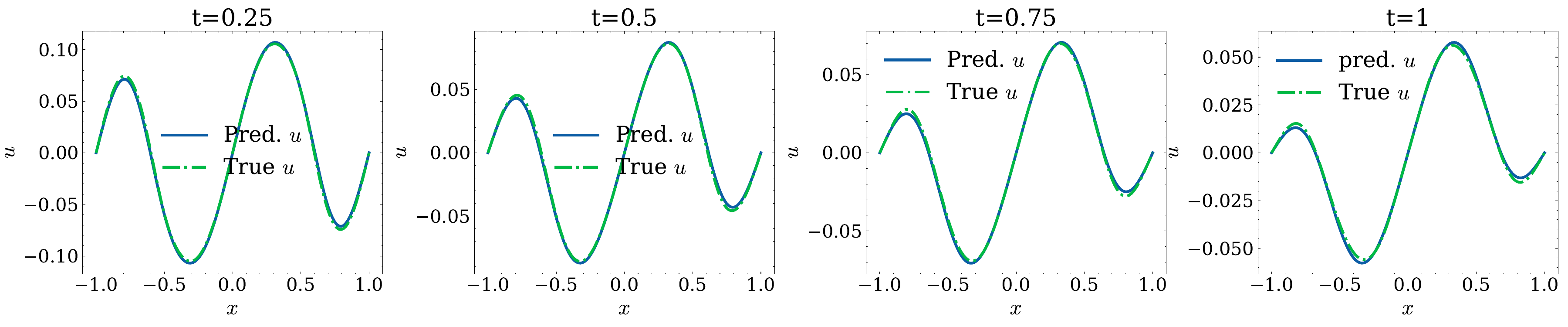}}
    \subfigure[PI-MultiONet]{\label{fig:pimultionet_out}
        \includegraphics[width=1.\textwidth]{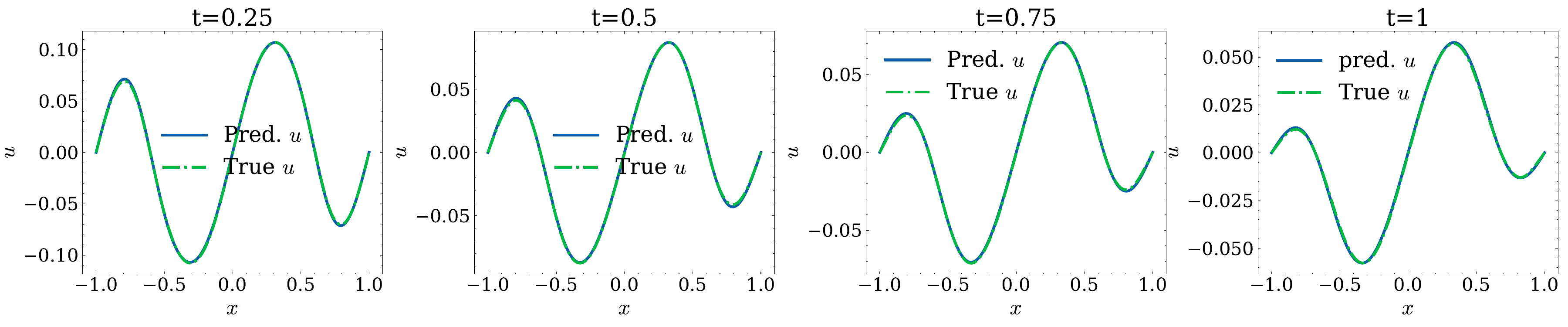}}
    \vspace{-0.25cm}
    \caption{Predicted solution profiles of the Burgers' equation at $t=0.25s, 0.5s, 0.75s, 1s$ obtained with the three competitive methods for an indicative initial condition from the \textbf{out-of-distribution} dataset.
    } 
    \label{fig:burgers_out}
\end{figure}

\begin{figure}[!htbp]
    \centering  
    \subfigure[Predicted solution $u$ (in-distribution test-case)]{\label{fig:burgers_u_in}
        \includegraphics[width=0.9\textwidth]{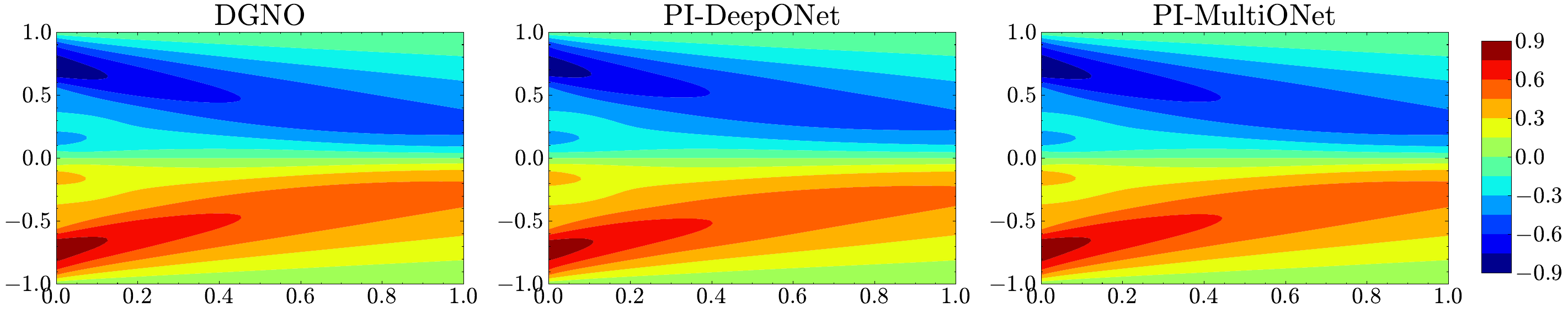}}
    \subfigure[Pointwise absolute errors (in-distribution test-case)]{\label{fig:burgers_uabs_in}
        \includegraphics[width=0.9\textwidth]{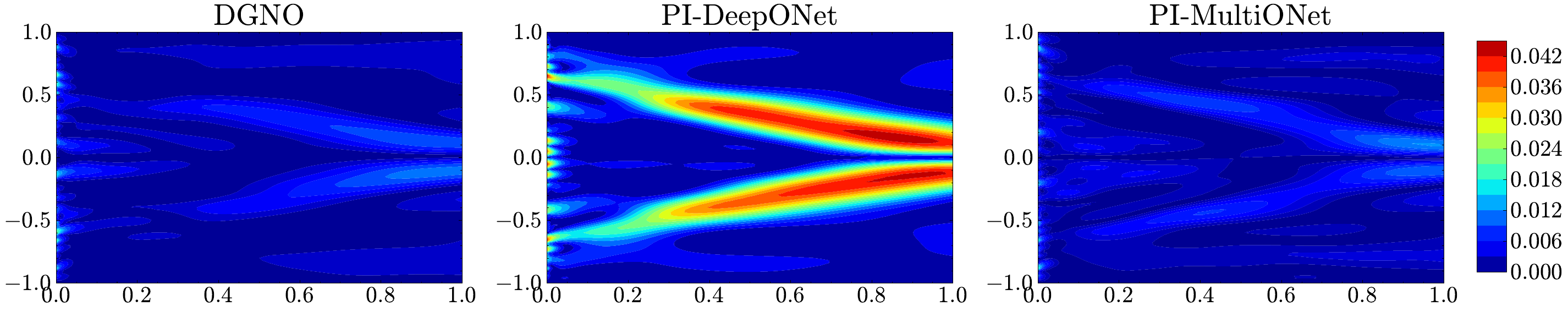}}
    \subfigure[Predicted solution $u$ (out-of-distribution test-case)]{\label{fig:burgers_u_out}
        \includegraphics[width=0.9\textwidth]{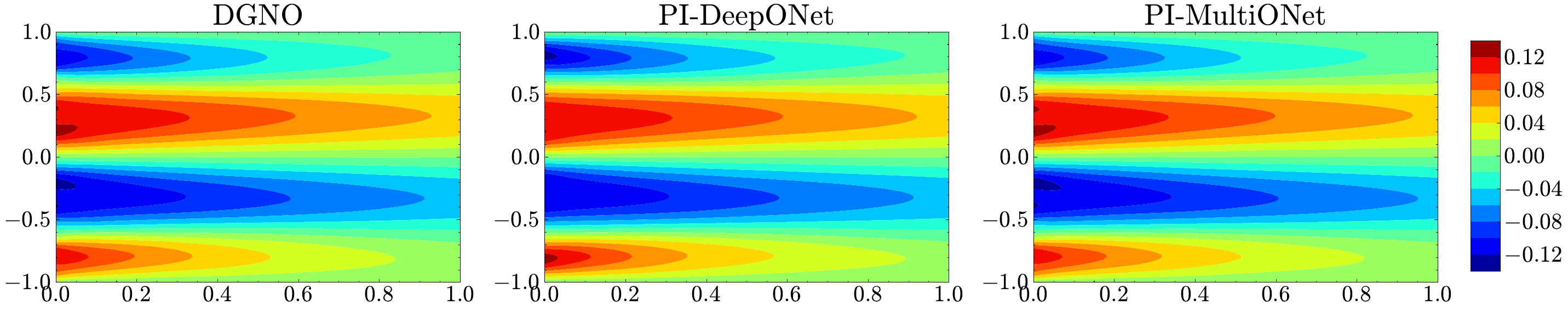}}
    \subfigure[Pointwise absolute errors (out-of-distribution test-case)]{\label{fig:burgers_uabs_out}
        \includegraphics[width=0.9\textwidth]{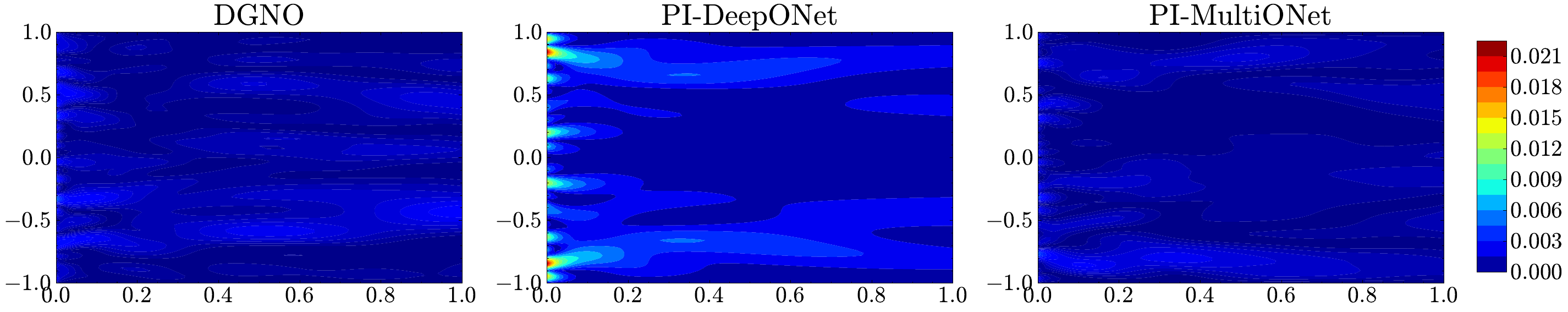}}
    \vspace{-0.25cm}
    \caption{Performance of each method on a representative test case for Burgers’ equation: (a) Predicted solution $u$ in the in-distribution case; (b) Pointwise absolute errors in the in-distribution case; (c) Predicted solution $u$ in the out-of-distribution case; (d) Pointwise absolute errors in the out-of-distribution case.} 
    \label{fig:burgers_in_out}
\end{figure}

In Figures \ref{fig:burgers_in} and \ref{fig:burgers_out}, we plot predictions of the solution profile at time instants $t=0.25s, 0.5s, 0.75s, 1s$ obtained by each method for an indicative $a$ extracted from the in-distribution and out-of-distribution test datasets, respectively. In Figure \ref{fig:burgers_in_out} we depict the whole solution as a function of space and time predicted by each of the three methods, as well as point-wise errors in the $(x,t)$ space. The illustrations support the summary results shown in Table \ref{tab:burgers}.
\begin{table}[tb]\small
\centering
\begin{tabular}{c|c|c|cc} \bottomrule
                    {} & DGenNO & PI-DeepONet  & PI-MultiONet \\ \hline
    {RMSE (in)} & $1.24e^{-2}\pm7.43e^{-3}$ & $3.01^{-2}\pm 1.75e^{-2}$  & $1.23e^{-2}\pm 7.88e^{-3}$ \\
    {RMSE (out)} & $2.01e^{-2}\pm1.59e^{-2}$ & $3.09e^{-2}\pm 2.26e^{-2}$ & $1.79e^{-2}\pm 1.69e^{-2}$ \\
    {Time(s)} & $0.0423\pm0.022$ & $0.0167\pm0.0062$ & $0.0403\pm0.0077$ \\\toprule
\end{tabular}
\caption{Performance of each method in solving the Burgers' problem \eqref{eq:burgers}: RMSE(in): the RMSE of each method in the in-distribution dataset; RMSE(out): the RMSE of each method in the out-of-distribution dataset; Time(s): time consumption of each method to predict numerical solutions.}
\label{tab:burgers}
\end{table}
\subsection{Stokes flow with a cylindrical obstacle}
\begin{figure}[tb]
\centering
\includegraphics[width=0.75\textwidth]{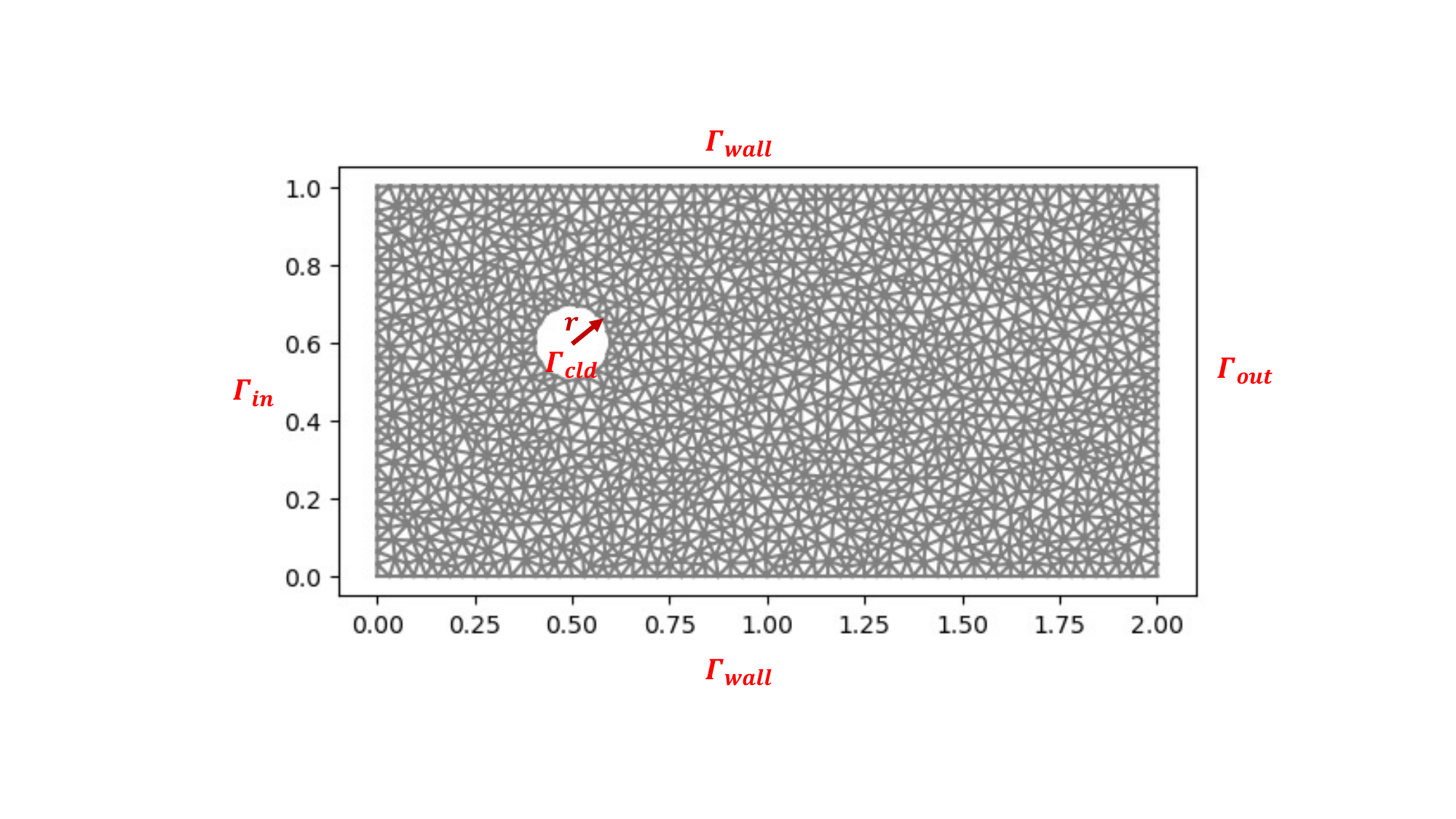}
\caption{The problem domain for the Stokes equation \eqref{eq:stokes} and the FEM mesh  used  for obtaining the reference PDE-solutions.}
\label{fig:stokes_mesh}
\end{figure}

In this section, we consider the Stokes equations in a domain containing a cylindrical obstacle, a benchmark problem frequently studied in related works \cite{rao2020physics,hu2023applying}. The governing PDEs and associated boundary conditions are:
\be\label{eq:stokes}
\begin{split}
-\mu\nabla^2{\bf u} + \nabla p &={\bf 0}, \quad \text{in}\ \Omega/\Omega_{cld}, \\
\nabla\cdot {\bf u} &= 0, \quad \text{in}\ \Omega/\Omega_{cld}, \\
\bm{u}(0,x_2) &= (a(x_2), 0), \quad \text{on}\ \Gamma_{in}, \\
p &= 0, \quad \text{on}\ \Gamma_{out}, \\
\bm{u} &= (0,0), \quad \text{on}\ \Gamma_{wall}\cup\Gamma_{cld},
\end{split}
\ee
where ${\bf u} = (u_1,u_2)$ is the velocity vector, $p$ is the pressure, and $\mu$ is the dynamic viscosity, which is set to $\mu = 0.01$ here. The problem domain is a rectangular region $\Omega = [0,2]\times[0,1]$ containing a cylinder $\Omega_{cld}$ centered at $(0.5,0.6)$ with a radius $r=0.1$, which is shown in Figure \ref{fig:stokes_mesh}.
The PDE-input $a$ pertains to the horizontal inflow velocity profile and is assumed to have the form $a(x_2) = 2\sin(\pi x_2)*(1. + \sin(k_1 x_2) + \cos(k_2 x_2))$, where the wavenumber parameters $k_1$ and $k_2$ are independently sampled from the uniform distribution $\text{U}(0,2\pi)$. 
For PI-DeepONet we consider  sensors $\Xi = \{\xi_i\}_{i=1}^m$ on a regular grid of $m=256$ on the interval $[0,1]$. Hence, the input  for the branch network in the PI-DeepONet is the vector $a(\Xi) \in \mathbb{R}^{256}$ which also serves as the training data $\hat{\bs{a}}^{(i)}$ in the proposed DGenNO method. For this, we use a latent vector $\bm{\beta}\in\mathbb{R}^{d_{\beta}=64}$ (details on the encoder network's structure are contained in the \ref{sec:stokes}). This problem presents several challenges: Firstly, the presence of a cylinder obstacle complicates the use of neural operators that rely on regular grid-based differentiation, such as PINO. Secondly, the Stokes system couples two equations with multiple boundary conditions, making it difficult for strong-form-based methods (e.g., PINNs, PI-DeepONet) to achieve high accuracy.

In DGenNO, we employ $M = 256$ weighted residuals in total (see Equation \ref{eq:weak_residual_stokes} in \ref{sec:weak_residual}), of which $56$ correspond to weighting functions with support in  a refinement region $\Omega_{refine} = [0.3,0.7]\times[0.4,0.8]/\Omega_{cld}$. The number of integration points is set to $N_{int} = 32$ for computing the associated integrals. For a fair comparison, we use the same integration points as collocation points in the strong-form residuals for the PI-DeepONet method, totaling $8192$, out of which $1792$ are located in the refinement region $\Omega_{refine}$.
To enforce the boundary conditions, we consider $256$ equally spaced points along each of the boundaries $\Gamma_{in}$, $\Gamma_{out}$, $\Gamma_{wall}$ and $\Gamma_{cld}$, resulting in a total of $dim(\hat{\bs{g}}) = 1280$ boundary observables (\refeqp{eq:likeg}). The associated hyperparameter was set to $\lambda_{bc} = 2$. Finally, we set the hyperparameter corresponding to the PDE loss $\lambda_{pde} = 10$ for the DGenNO method and $\lambda_{pde} =2$ \cite{jiao2024solving}  for PI-DeepONet and PI-MultiONet. The models are trained using the ADAM optimizer with an initial learning rate of $lr = 5 \times 10^{-4}$, which is reduced by a factor of two every $2500$ epoch.

\begin{figure}[!htbp]
    \centering  
    \subfigure[Reference velocity fields $u_x$ and $u_y$ (in-distribution case)]{\label{fig:stokesHole_problem_in}
        \includegraphics[width=0.75\textwidth]{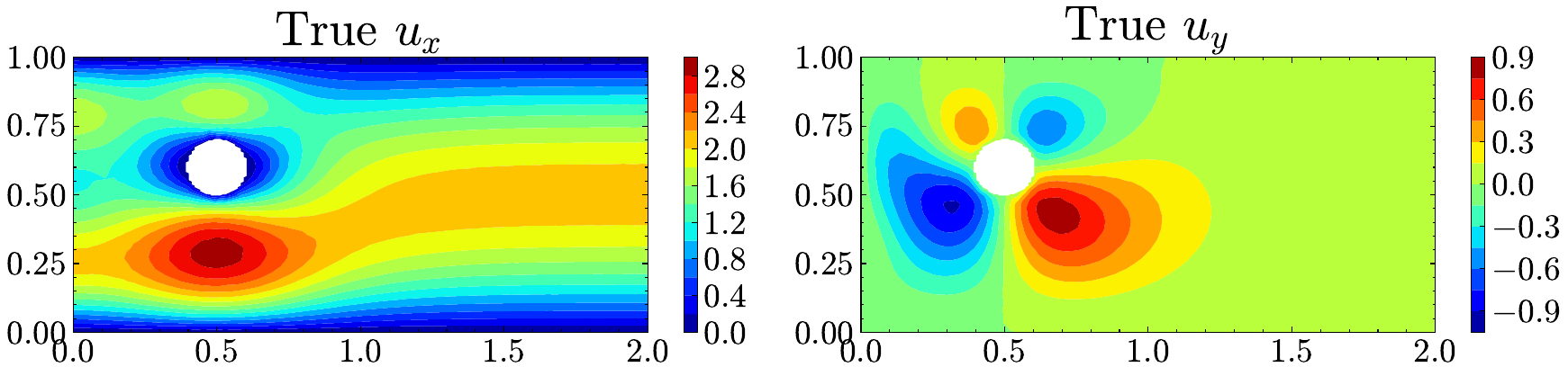}}
    \subfigure[The prediction of $u_x$]{\label{fig:stokesHole_u_in}
        \includegraphics[width=0.9\textwidth]{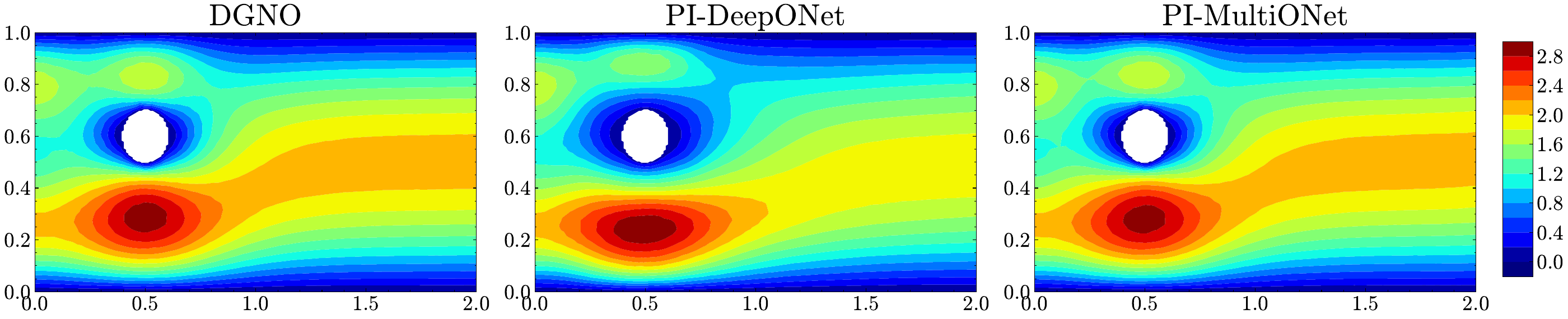}}
    \subfigure[The pointwise absolute error for $u_x$]{\label{fig:stokesHole_uabs_in}
        \includegraphics[width=0.9\textwidth]{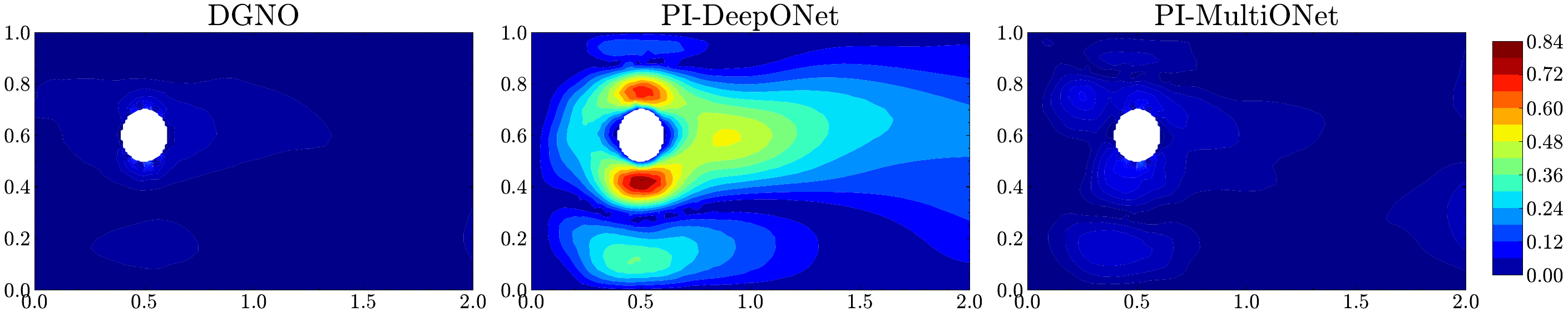}}
    \subfigure[The prediction of $u_y$]{\label{fig:stokesHole_v_in}
        \includegraphics[width=0.9\textwidth]{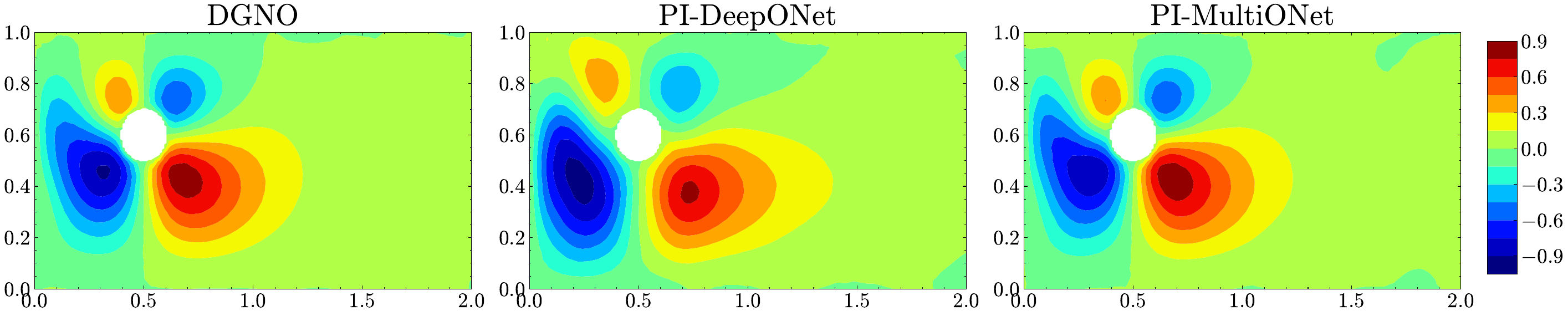}}
    \subfigure[The pointwise absolute error for $u_y$]{\label{fig:stokesHole_vabs_in}
        \includegraphics[width=0.9\textwidth]{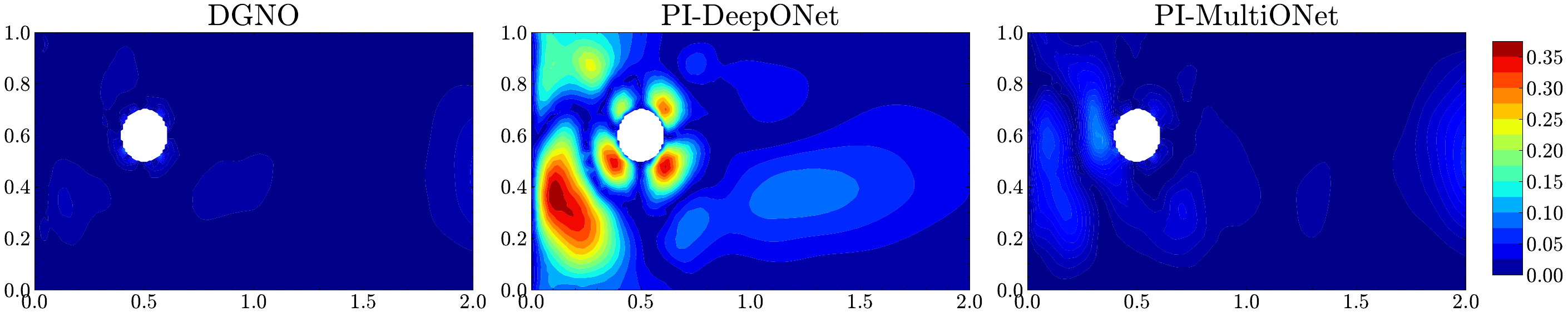}}
    \vspace{-0.25cm}
    \caption{Indicative in-distribution test-case for the Stokes equation.}
    \label{fig:stokesHole_in}
\end{figure}

We assess the predictive performance on two test datasets consisting of $200$ samples of $a$ each. In particular:
\bi
\item an \textbf{in-distribution} dataset obtained by by sampling $k_1,k_2$ from $\text{U}(0,2\pi)$, and
\item an \textbf{out-of-distribution} dataset obtained by sampling  $k_1,k_2$ from $\text{U}(2\pi,2\pi+\pi/4)$.
\ei
Reference solutions of the velocity $\bs{u}$ were obtained with the FEM  mesh shown in Figure \ref{fig:stokes_mesh}. The RMSE and computational time  for predicting the solution are shown in Table \ref{tab:stokes} for each method. In Figure \ref{fig:stokesHole_in}, we plot the predictions of the velocity field $\bm{u}$ for an instance sampled from the in-distribution testing dataset and compare with the  ground truth which is seen in Figure \ref{fig:stokesHole_problem_in}.
The RMSE achieved by the proposed DGenNO framework is much smaller than the ones for PI-DeepONet and PI-MultiONet. Furthermore, as we can see in Figure \ref{fig:stokesHole_uabs_in} and \ref{fig:stokesHole_vabs_in}, the PI-DeepONet method performs poorly near the cylinder's boundary $\Gamma_{cld}$. However, the MultiONet architecture exhibits improved accuracy, especially near the cylinder. Given that it was trained on exactly the same data as PI-DeepONet, these results suggest that the MultiONet architecture has better approximation capabilities than the DeepONet one. 
In addition, the superior performance of the DGenNO method compared to the PI-MultiONet method indicates the advantage of the weak-form residuals and the generative framework.

\begin{figure}[!htbp]
    \centering  
    \subfigure[Reference velocity fields $u_x$ and $u_y$ (out-of-distribution case)]{\label{fig:stokesHole_problem_out}
        \includegraphics[width=0.75\textwidth]{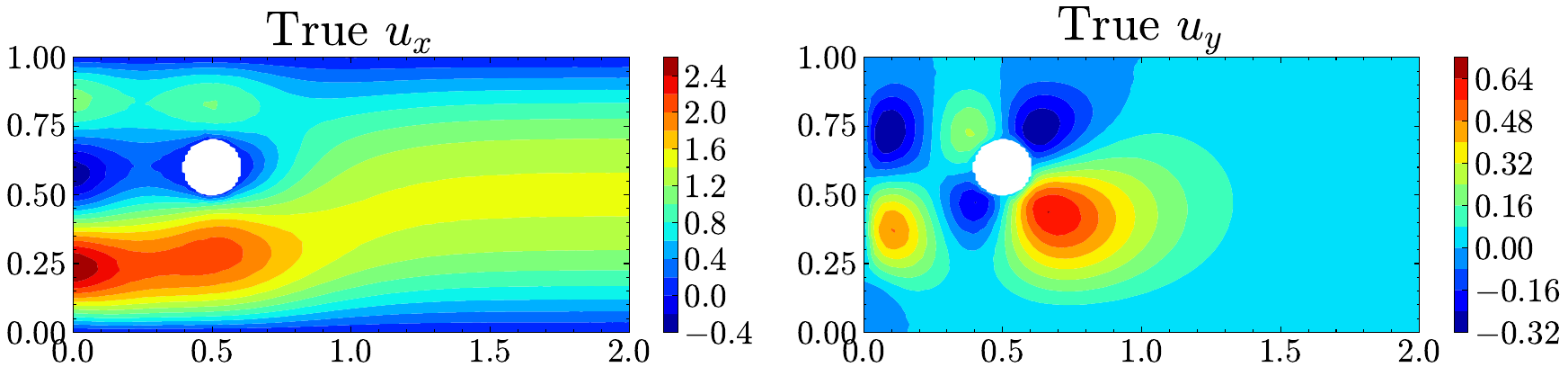}}
    \subfigure[The prediction of $u_x$]{\label{fig:stokesHole_u_out}
        \includegraphics[width=0.9\textwidth]{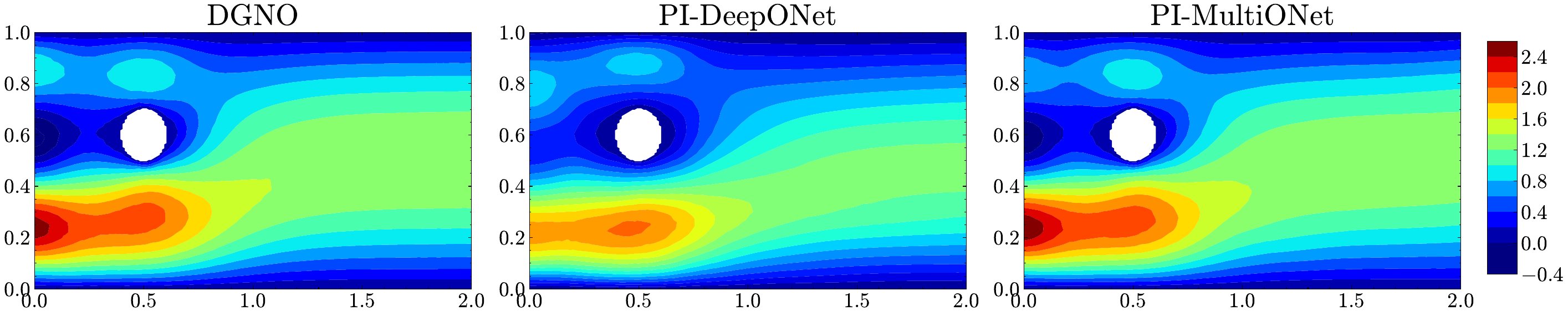}}
    \subfigure[The pointwise absolute error for $u_x$]{\label{fig:stokesHole_uabs_out}
        \includegraphics[width=0.9\textwidth]{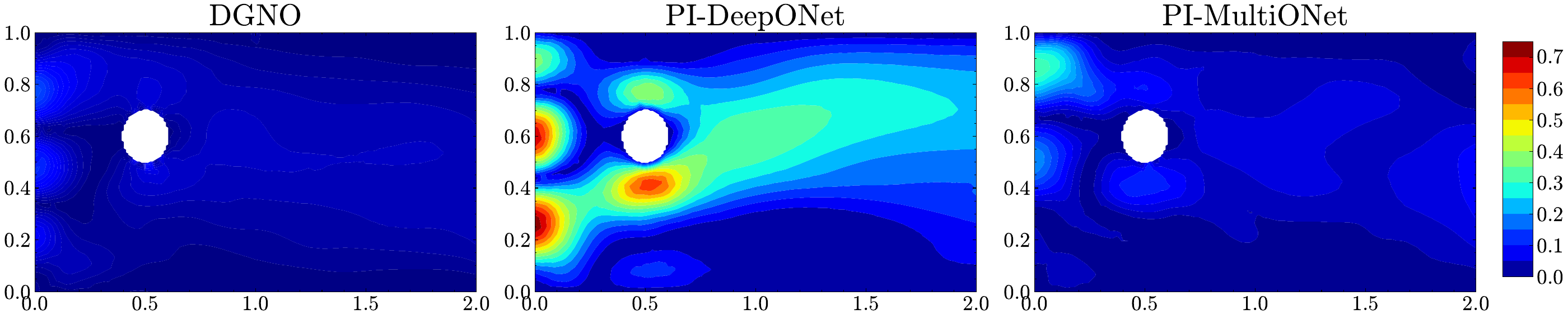}}
    \subfigure[The prediction of $u_y$]{\label{fig:stokesHole_v_out}
        \includegraphics[width=0.9\textwidth]{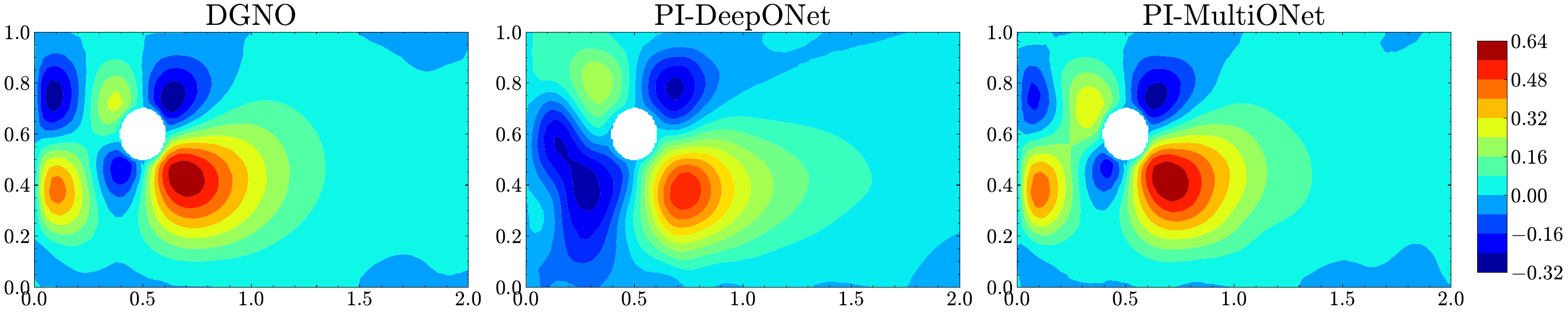}}
    \subfigure[The pointwise absolute error for $u_y$]{\label{fig:stokesHole_vabs_out}
        \includegraphics[width=0.9\textwidth]{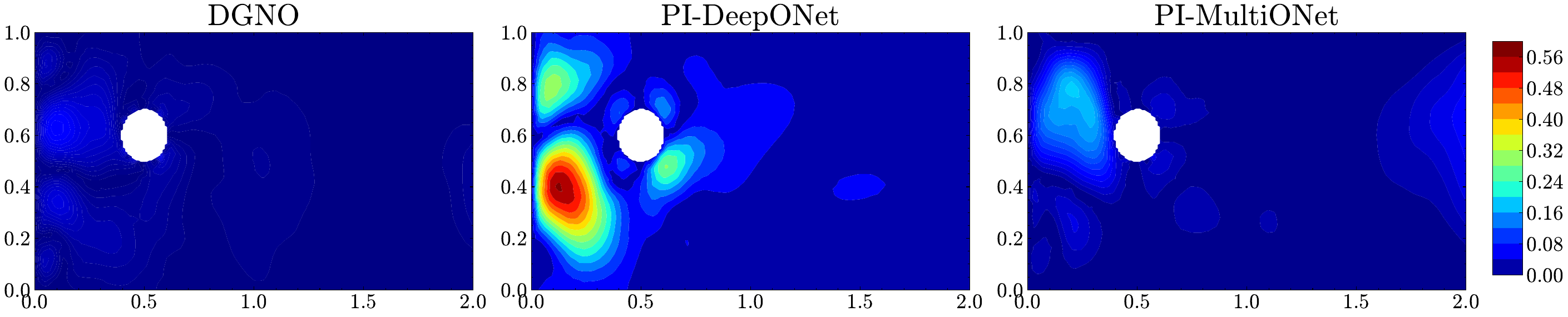}}
    \vspace{-0.25cm}
    \caption{Indicative out-of-distribution test-case for the Stokes equation.}
    \label{fig:stokesHole_out}
\end{figure}

In the out-of-distribution test dataset (see Table \ref{tab:stokes} and Figure \ref{fig:stokesHole_out}) the  performance of all three methods degrades, but  the DGenNO  still significantly outperforms the PI-DeepONet and the PI-MultiONet. This suggests a superior  generalization capability for this problem. The performance of PI-MultiONet is significantly better than that of PI-DeepONet, which reinforces the previously stated conclusions. 
\begin{table}[tb]\small
\centering
\begin{tabular}{c|c|c|c|cc} \bottomrule
                    \multicolumn{2}{c|}{}& DGenNO & PI-DeepONet  & PI-MultiONet \\ \hline
    \multirow{2}{*}{RMSE(in)} & 
    {$u_x$} & $3.16e^{-2}\pm2.02e^{-2}$ & $0.142\pm 0.010$  & $3.46e^{-2}\pm1.84e^{-2}$ \\ 
    {} & {$u_y$} & $3.07e^{-2}\pm4.31e^{-3}$ & $0.328\pm0.034$ & $6.70e^{-2}\pm1.76e^{-2}$ \\ \hline
    \multirow{2}{*}{RMSE(out)} & 
    {$u_x$} & $3.71e^{-2}\pm1.51e^{-2}$ & $0.198\pm 0.045$ & $6.64e^{-2}\pm2.04e^{-2}$ \\
    {} & {$u_y$} & $1.16e^{-1}\pm4.89e^{-2}$ & $0.581\pm0.125$ & $2.62e^{-1}\pm8.41e^{-2}$\\ \hline
    \multicolumn{2}{c|}{Time(s)} & $0.0213\pm0.0185$ & $0.0081\pm0.0229$ & $0.020\pm0.021$ \\\toprule
\end{tabular}
\caption{Performance of each method in solving the Stokes equation (\refeqp{eq:stokes}): RMSE(in): the RMSE of each method in the in-distribution dataset; RMSE(out): the RMSE of each method in the out-of-distribution dataset; Time(s): time consumption of each method to predict numerical solution}
\label{tab:stokes}
\end{table}

\subsection{Inverse problems - Darcy flow}
\label{sec:inverse}
In this section, we consider the solution of model-based inverse problems to demonstrate the crucial role of the latent variables $\bm{\beta}$ and show the superiority of the proposed method over state-of-the-art alternatives. Specifically, we consider the recovery of the permeability field $a$ of a two-phase medium from noisy measurements of the pressure field $u$ on sparsely distributed sensors across the problem domain in \refeqp{eq:darcy_flow}. 

This problem presents a significant challenge as the target coefficient is discontinuous due to the fact that $a(\bx)$ can take two values. We note that (irrespective of the discretization of $a(\bx)$) even the most accurate surrogate (in the form of a neural operator or not) would not be able to yield efficient solutions due to the absence of derivatives with respect to $a(\bx)$ (or the discretization/representation thereof).
To comparatively assess the performance of the proposed DGenNO framework, we consider the following baseline methods: the popular PINN method \cite{raissi2019physics} and the ParticleWNN method \cite{zang2023particlewnn}. The latter has been shown to outperform  PINNs in a similar problem where the unknown permeability field is a continuous function.

\begin{figure}[!htbp]
    \centering 
    \subfigure[Reference $a$ and corresponding solution $u$ (black dots indicated location of observations)]{\label{fig:inverse_darcy_pwc_true}
        \includegraphics[width=0.6\textwidth]{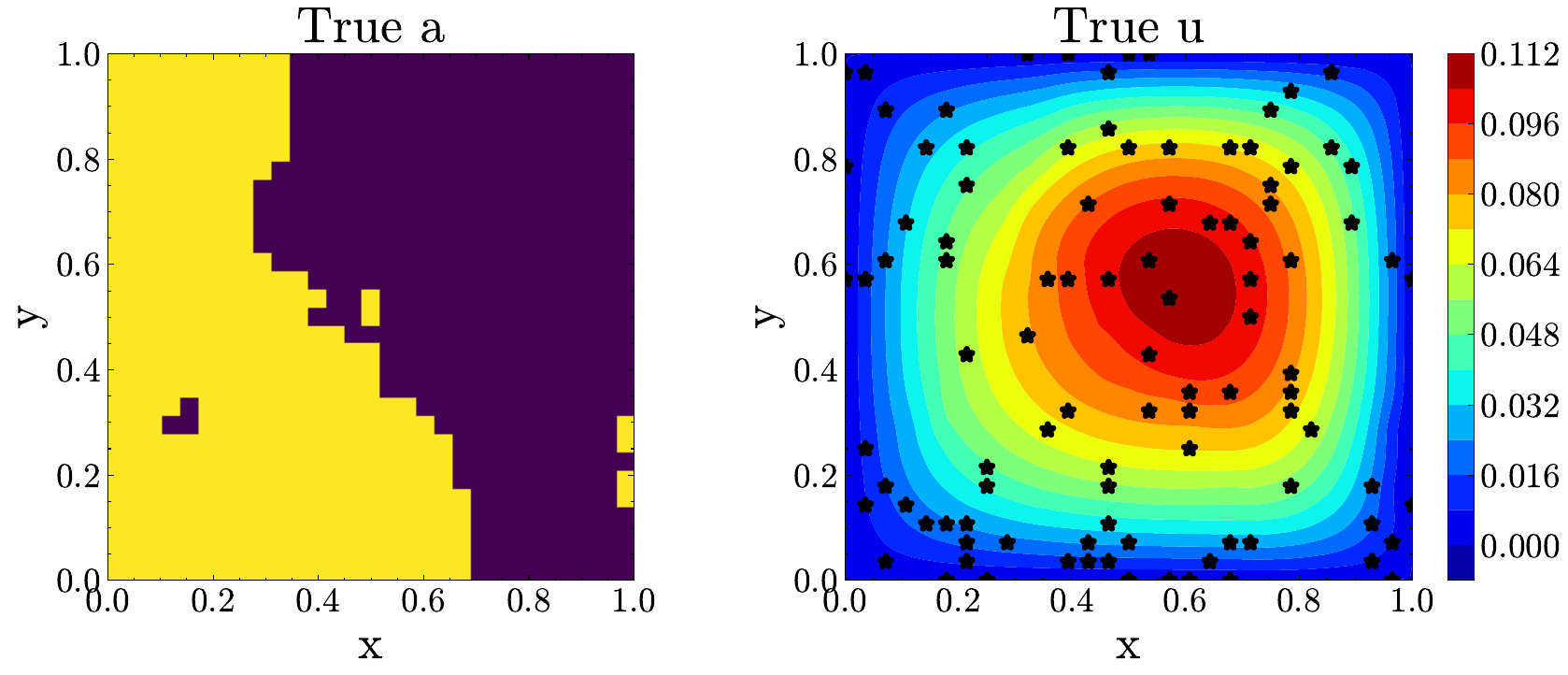}}
    \subfigure[SNR=100]{\label{fig:inverse_darcy_pwc_snr100}
        \includegraphics[width=1.\textwidth]{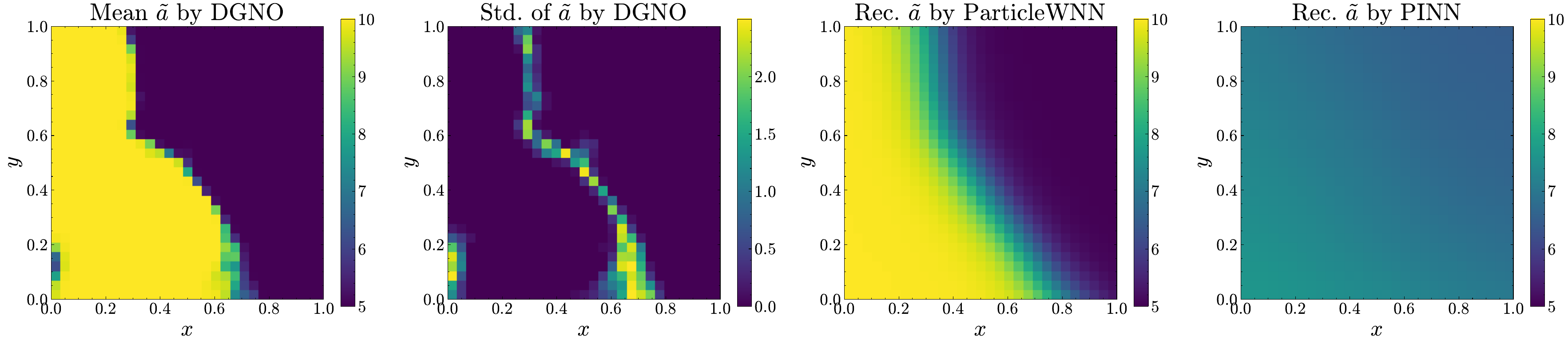}}
    \subfigure[SNR=50]{\label{fig:inverse_darcy_pwc_snr50}
        \includegraphics[width=1.\textwidth]{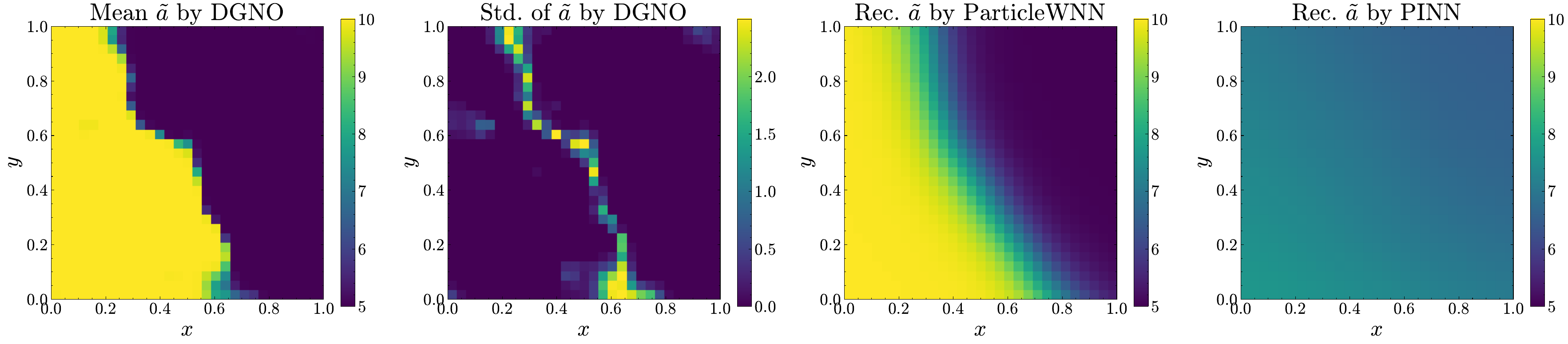}}
    \subfigure[SNR=20]{\label{fig:inverse_darcy_pwc_snr20}
        \includegraphics[width=1.\textwidth]{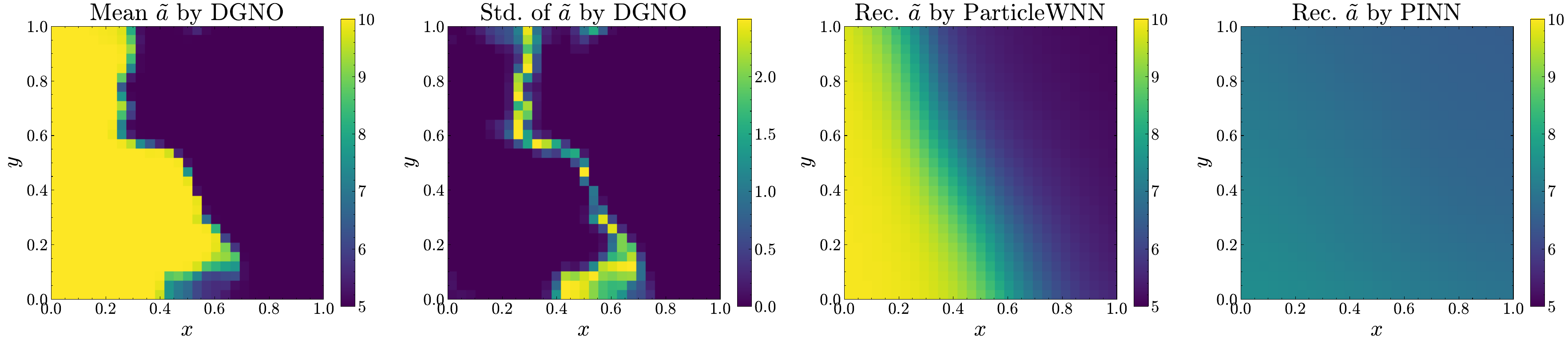}}
    \vspace{-0.25cm}
    \caption{The performance of the DGenNO method, the PINN method, and the ParticleWNN method in the problem of recovering a piecewise constant coefficient field in Darcy's equation \eqref{eq:darcy_flow}: (a) the ground-truth coefficient $a$ (left) and solution $u$ (right) obtained by the FEM, where black dots represent observation sensors; The recovered coefficient obtained by the DGenNO ($\tilde{a}$) and other methods (Pred. $a$) under noise levels: (b) SNR=100, (c) SNR=50, and (d) SNR=20.} 
    \label{fig:inverse_darcy_pwc}
\end{figure}

We generated synthetic data by sampling a permeability field from a zero-cutoff $GP(0,(-\Delta + 9I)^{-2})$ (as in section \ref{sec:darcy_flow}) and by obtaining the reference solution $u$ using the same FEM solver discussed in section \ref{sec:darcy_flow}. We collected the solution values on a randomly sampled\footnote{using the uniform distribution.} set of $100$ grid points $X_{obs}=\{\bm{x}_{k}\}^{K=100}_{k=1}$  which are shown in Figure \ref{fig:inverse_darcy_pwc_true}. We contaminated these values with additive Gaussian noise to obtain the $100$ observations as: 
\be 
u_{obs}(\bm{x}_k) = u(\bm{x}_k) + \epsilon_k, \quad \epsilon_k\sim\Ncal(0,\sigma^2)
\label{eq:darcyuobs}
\ee 
where $\sigma$ is determined by the Signal-to-noise ratio (SNR) through the following relation:
\be 
SNR = 10\log_{10}\frac{\frac{1}{K}\sum^{K}_{k=1}u^2(\bx_k)}{\sigma^2}.
\ee 
We consider three different noise levels, i.e. low (SNR=100), medium (SNR=50), and high (SNR=20).
Since RMSE is no longer suitable for measuring the inversion performance of piecewise constant coefficient fields, we use the cross-correlation indicator \cite{bourke1996cross} $I_{corr}$ to measure the discrepancy between recovered $\tilde{a}$ and the ground-truth coefficient $a$.  For a piecewise-constant, binary field, $I_{corr}$ is defined as:
\be 
I_{corr} = \frac{\sum_i a^2(\bm{\xi}_i)\tilde{a}^2(\bm{\xi}_i)}{\sqrt{\sum_i a^2(\bm{\xi}_i)}\sqrt{\sum_i \tilde{a}^2(\bm{\xi}_i)}}.
\ee 
The value of $I_{corr}$  ranges from 0 to 1, where a higher $I_{corr}$ indicates a greater similarity between the two images, with $I_{corr} = 1$ representing a perfect match.
We solve this problem by using the DGenNO method, the PINN method, and the ParticleWNN method, with their model setups described in \ref{sec:inverse_pwc}. We note that in the case of the competitive methods in order to ensure the requisite differentiability, the output of the neural network with which the field $a$ is represented can take any value between the values of the two phases, i.e. it is not guaranteed to be binary as one can see in the ensuing illustrations.  
The recovered coefficient fields obtained by the DGenNO (denoted as $\tilde{a}$) and other methods (denoted as Pred. $a$) under noise levels $SNR=100, 50, 20$ are displayed in Figures \ref{fig:inverse_darcy_pwc_snr100}, \ref{fig:inverse_darcy_pwc_snr50}, and \ref{fig:inverse_darcy_pwc_snr20}, respectively. The cross-correlation $I_{corr}$ for each method is reported in Table \ref{tab:inverse_darcy_pwc}. We observe that the proposed DGenNO method outperforms the competitors for all noise-level scenarios. Although the ParticleWNN method also obtained acceptable $I_{corr}$ values, it can only capture general features of the underlying field, but not finer details, especially along the edges of the two different phases (see  Figure \ref{fig:inverse_darcy_pwc}). Therein we also observe that the PINN method fails in the task for all three noise levels. This is because PINNs rely on strong-form-residuals to define the PDE loss which causes significant difficulties in dealing with discontinuous fields as those appearing in two- or multi-phase media.
Moreover, Table \ref{tab:inverse_darcy_pwc} suggests that the DGenNO method is more robust to noise than the other methods, as the $I_{corr}$ metric degrades by a small amount as the noise level increases (i.e. the SNR decreases). The latter is also reflected in the increased posterior standard deviation in the inferred field as can be seen in the second column of Figure \ref{fig:inverse_darcy_pwc}. For the ParticlWNN method, we observe that the $I_{corr}$ metric drops more drastically as the SNR decreases whereas PINN retains the same, albeit rather low, $I_{corr}$ values.
\begin{table}[tb]\small
\centering
\begin{tabular}{c|c|c|cc} \bottomrule
                    {} & DGenNO & ParticleWNN  & PINN \\ \hline
    {SNR=100} & $0.948$ & $0.891$  & $0.595$ \\
    {SNR=50} & $0.927$ & $0.886$ & $0.596$ \\
    {SNR=20} & $0.916$ & $0.805$ & $0.563$ \\\toprule
\end{tabular}
\caption{Cross-correlations $I_{corr}$ obtained by different methods in the problem of recovering piecewise constant coefficient in Darcy's equation \eqref{eq:darcy_flow} under different noise levels.}
\label{tab:inverse_darcy_pwc}
\end{table}

\begin{figure}[!htbp]
    \centering  
    \subfigure[Reference $a$ and corresponding solution $u$ (black dots indicated location of observations)]{\label{fig:inverse_darcyCTS}
        \includegraphics[width=0.6\textwidth]{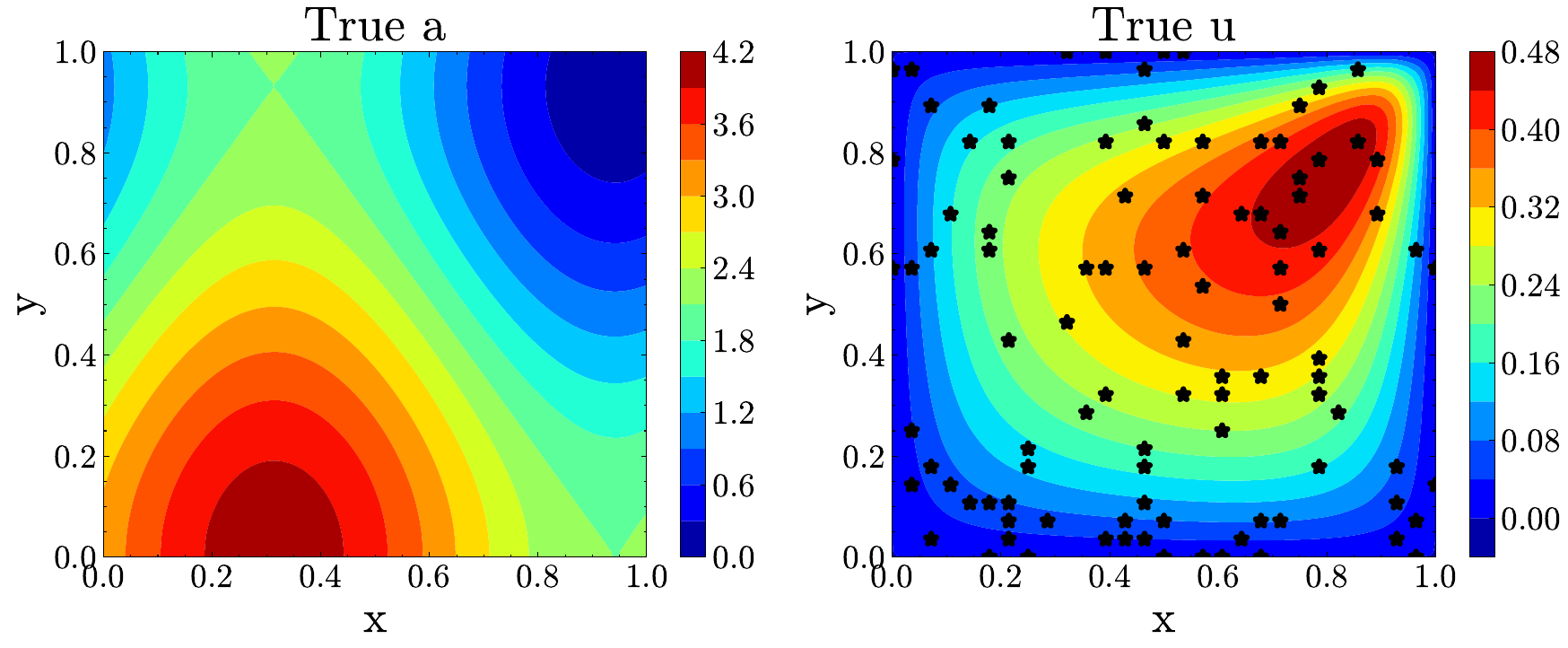}}
    \subfigure[SNR=100]{\label{fig:inverse_darcy_cts_snr100}
        \includegraphics[width=0.98\textwidth]{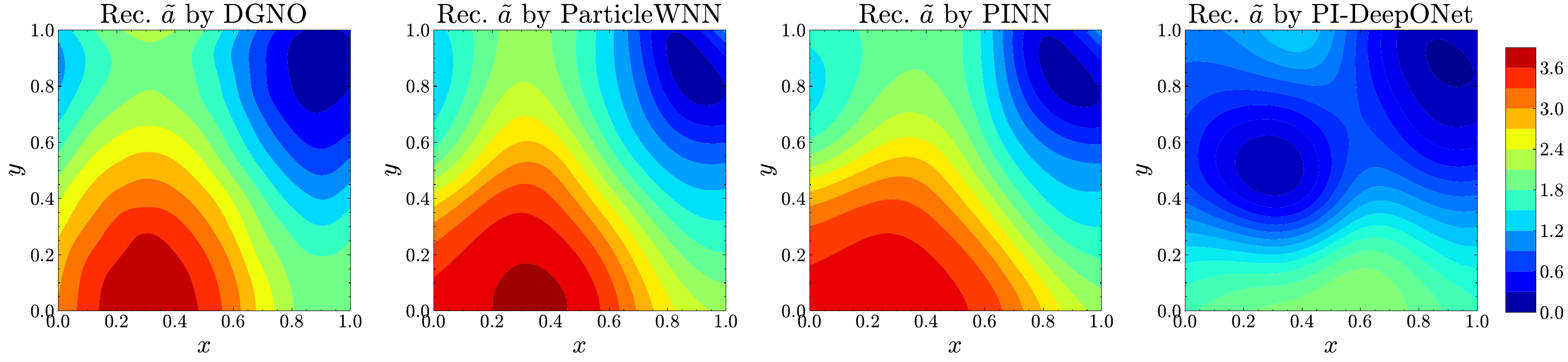}}
    \subfigure[SNR=50]{\label{fig:inverse_darcy_cts_snr50}
        \includegraphics[width=0.98\textwidth]{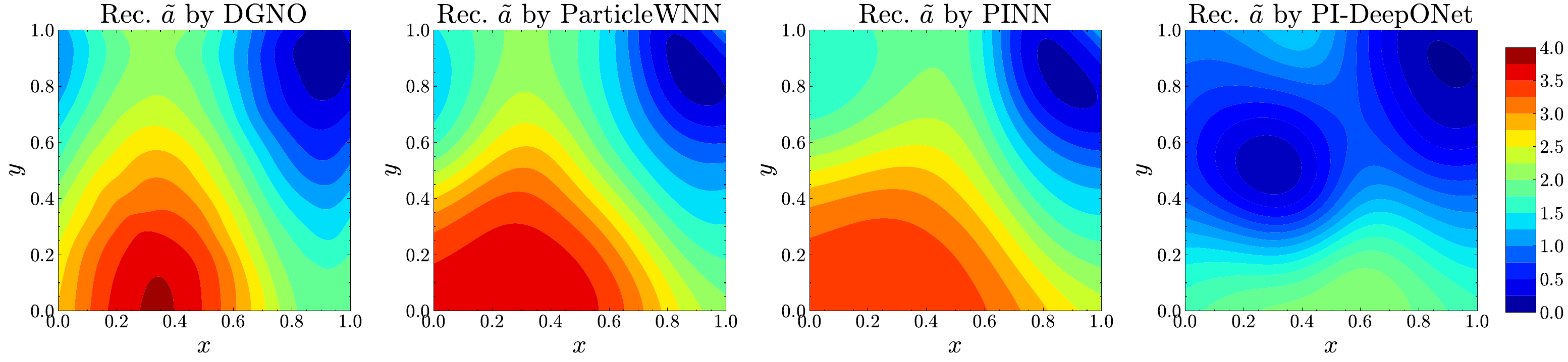}}
    \subfigure[SNR=20]{\label{fig:inverse_darcy_cts_snr20}
        \includegraphics[width=0.98\textwidth]{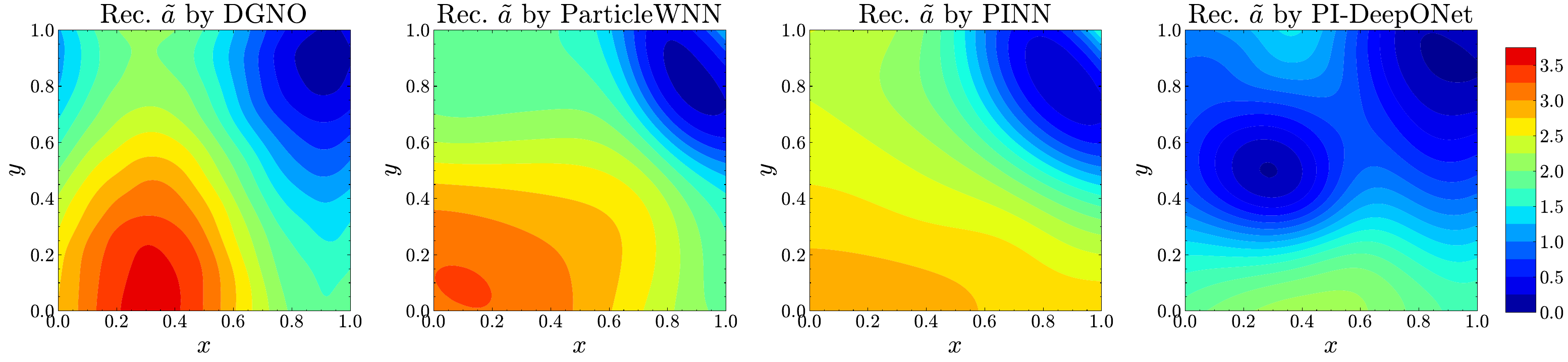}}
    \vspace{-0.25cm}
    \caption{The performance of the DGenNO method, the PINN method, the ParticleWNN method, and the PI-DeepONet method (adapted from \eqref{eq:posterior_inv}) in the problem of recovering continuous coefficient field in Darcy's equation \eqref{eq:darcy_flow}: (a) the ground-truth $a$ (left) and solution $u$ (right) obtained by the FEM, where black dots represent observation sensors; The recovered coefficient obtained by the DGenNO (Rec. $\tilde{a}$) and by other methods (Pred. $a$) under noise levels: (b) SNR=100, (c) SNR=50, and (d) SNR=20.} 
    \label{fig:inverse_darcy_cts}
\end{figure}

We also consider solving the same model-based inverse problem when the unknown coefficient field is a continuous function. In this case, the PI-DeepONet is also employed as a comparison. We note that the PI-DeepONet method differs from the other methods as the PDE (e.g., residuals) is not to be used for solving the inverse problem. Therefore, in presenting its performance, we explicitly distinguish PI-DeepONet results from other methods in both Table \ref{tab:inverse_darcy_cts} and Figure \ref{fig:inverse_darcy_cts}.
To train both the DGenNO and the PI-DeepONet models, we keep all problem settings identical to those in section \ref{sec:darcy_flow} and  \ref{sec:inverse_cts}. We considered  permeability fields $a$ of the form $a(x_1,x_2) = 2.1 + \sin(k_1 x_1) + \cos(k_2 x_2)$, where $k_1$ and $k_2$ are sampled from the uniform distribution $U(0, 2\pi)^2$ independently.
The RMSE of the forward prediction of $u$ obtained by the DGenNO and the PI-DeepONet in an in-distribution dataset, which is generated by sampling $200$ pairs of $(k_1,k_2)$ from $U(0, 2\pi)^2$, are $4.31e^{-3}\pm5.76e^{-3}$ and $7.23e^{-3}\pm 4.33e^{-3}$, respectively.
For the inverse problem, we consider a specific target coefficient $a$ obtained from a single $(k_1,k_2)$ sampled from $U(0, 2\pi)^2$. We obtain the reference solution $u$ through FEM and contaminate it with additive Gaussian noise as in \refeqp{eq:darcyuobs}.

We then solve this problem with the proposed DGenNO, the PINN, the ParticleWNN, and the PI-DeepONet methods with their setting described in \ref{sec:inverse_cts}.
The recovered coefficient fields obtained by the DGenNO (denoted as Rec. $\tilde{a}$) and other methods (denoted as Pred. $a$) under noise levels SNR=$100, 50, 20$ are shown in Figure \ref{fig:inverse_darcy_cts_snr100}, \ref{fig:inverse_darcy_cts_snr50}, and \ref{fig:inverse_darcy_cts_snr20}, respectively. The RMSE between the recovered coefficient and the ground-truth coefficient obtained by each method is recorded in Table \ref{tab:inverse_darcy_cts}. It is evident that the PI-DeepONet method almost failed in solving the inverse problem, whereas the other three methods were successful at least for low to moderate noise levels. \revise{The primary reason for the failure of PI-DeepONet lies in the difficulty of directly optimizing the input function $a$ in a high-dimensional and potentially irregular space. This becomes especially challenging when the available observations are sparse and noisy. In contrast, the proposed DGenNO method addresses this challenge by introducing a latent space and exploring optimal solutions in a smooth, low-dimensional space, making the optimization problem significantly more tractable.} In particular, the proposed DGenNO method and the ParticleWNN method obtain comparable results, and both are significantly better than the PINN method. This also demonstrates the advantage of the weak-form-residual-based framework over the strong-form-residual-based framework in dealing with this inverse problem.
For high noise levels (i.e. low SNR), the accuracy of the DGenNO method drops, but by a smaller amount as compared to the ParticleWNN and PINN methods. This suggests that the lower-dimensional embedding and the generative framework enhance the proposed DGenNO method's robustness against noise.
\begin{table}[tb]\small
\centering
\begin{tabular}{c|c|c|c||cc} \bottomrule
                    {} & DGenNO & ParticleWNN  & PINN & PI-DeepONet (adapted from \eqref{eq:posterior_inv})\\ \hline
    {SNR=100} & $0.051$ & $0.053$  & $0.078$ & $0.605$ \\
    {SNR=50} & $0.069$ & $0.065$ & $0.099$ & $0.606$ \\
    {SNR=20} & $0.092$ & $0.163$ & $0.218$ & $0.590$\\  
    \toprule
\end{tabular}
\caption{RMSE obtained by different methods in the problem of recovering continuous coefficient in Darcy's equation \eqref{eq:darcy_flow} under different noise levels.}
\label{tab:inverse_darcy_cts}
\end{table}
\section{Concluding remarks}
\label{sec:conclusion}
We introduced the Deep Generative Neural Operator (DGenNO), a novel framework that combines the power of deep, generative probabilistic models and neural operators to solve forward and inverse problems governed by PDEs. A key feature of DGenNO is the utilization of unlabeled-data as well as weighted-residuals,  which eliminates the need for labeled data and constitutes making DGenNO an efficient tool for real-world applications where labeled data is sparse or unavailable.
The use of weak-form residuals enhances its ability to handle problems with abrupt changes or irregularities. This makes DGenNO particularly suitable for handling discontinuities and localized phenomena where strong-form-based methods struggle. Additionally, the use of CSRBFs as weighting functions improves DGenNO's computational efficiency and adaptability to complex geometries. In combination with the proposed MultiONet architecture, it can yield more accurate approximations as compared to popular alternatives with the same number of trainable parameters.

At the core of DGenNO’s effectiveness is a set of latent variables, which enable learning structured, lower-dimensional generative representations of the potentially discontinuous PDE-input functions as well as of the PDE-solutions. This formulation provides a robust mechanism for solving both forward and inverse problems. Numerical experiments demonstrate that DGenNO outperforms state-of-the-art methods such as PI-DeepONet, achieving superior accuracy both in solving complex parametric PDEs and recovering discontinuous PDE-input fields from noisy and sparsely distributed measurements. Its robust performance across varying noise levels and out-of-distribution cases underscores its generalization capabilities. 

The framework’s adaptability to diverse problem settings and its ability to handle sparse, noisy data make it a promising tool for applications in engineering, physics, and applied mathematics. On the methodological front, the automatic identification of the latent space's dimension would significantly enhance the usability of the proposed framework. As discussed previously, an unexplored avenue involves using a learnable, parameterized prior for the latent variables, which could reveal underlying structure in the space of latent generators (e.g., clusters). This could be highly beneficial not only for producing accurate predictions but also for providing valuable insight into the PDE input-output map. One final aspect that we have not explored is the use of inexpensive,  unlabeled data (i.e. just PDE-inputs). Generative models, in contrast to discriminative ones, can make use of such data which can improve the predictive accuracy of the model by enhancing  the identification
of the lower-dimensional encodings of the PDE-input \cite{rixner_probabilistic_2021}.
In terms of applications, the proposed DGenNO model is highly- suited for problems in inverse materials' design and especially in the context of multi-phase media and metamaterials \cite{mcdowell_integrated_2009}. Existing strategies are largely data-based \cite{bastek_inverse_2023}, require fine-tuning microstructural features  \cite{generale_inverse_2024}, and as such they do not exhibit good generalization performance required in design problems, which inherently involve extrapolative tasks.

\section*{Acknowledgements}
Partially funded by the Deutsche Forschungsgemeinschaft (DFG) through Project Number 499746055  and the Excellence Strategy of the Federal Government and the Länder in the context of the ARTEMIS Innovation Network.

\newpage
\bibliographystyle{elsarticle-num}
\bibliography{ref.bib}

\appendix
\section{Weighted residuals}
\label{sec:weak_residual}
Strong-form, collocation-type residuals are most commonly used in physics-informed methods such as PINNs and PI-DeepONets. They could be considered as special cases of the weighted residuals advocated in this work. For example, in the second-order elliptic equation with $\Ncal_a[u]=-\nabla\cdot(a\nabla u)$, the strong-form residual is given as:
\be 
r_{w_j}(a,u) = \int_{\Omega} (-\nabla\cdot(a\nabla u)- f)w_j ~d\bm{x} = -\nabla\cdot(a(\bm{x}_j)\nabla u(\bm{x}_j))- f(\bm{x}_j) 
\ee
where the weighting function $w_j=\delta(\bm{x}-\bm{x}_j),\ \bm{x}_j\in\Omega$ is a Dirac-delta function. This type of residual is straightforward to implement and avoids the need for integral approximations. However, it requires the computation of higher-order derivatives, leading to increased computational cost and higher regularity demands on the solution and coefficients. More importantly, when the derivative information of the coefficient $a$ is unavailable  - a common scenario in many practical problems - the strong residual cannot be computed. Consequently, the strong-form residual is less suitable for problems involving singularities in coefficients or solutions, as their gradients do not exist at discontinuities. As a result, numerical approximations of those gradients will introduce significant errors near discontinuities.

To address these limitations, we employ weak-form residuals, which for the same PDE as above, can be expressed as
\be\label{eq:residual_weak}
r_{w_j}(a,u) = \int_{\Omega} (a\nabla u\nabla w_j - fw_j) ~d\bmx
\ee 
where $w_j\in H^1_0(\Omega)$. While the weak-form residual circumvents the need for higher-order derivatives, its application in deep neural operators is hindered by the computational expense of accurately approximating the required integrals, especially when many integration points are needed. This challenge is manageable when solving a single PDE problem, as demonstrated in approaches such as DRM \cite{yu2018deep}, WAN \cite{zang2020weak}, and VPINN \cite{kharazmi2019variational}. However, for parametric PDEs requiring solutions over a family of problems, this computational demand becomes significant.
Recent advancements, particularly the ParticleWNN method in \cite{zang2023particlewnn}, offer a promising solution. This method reduces the need for extensive integration points in computing the weak residuals by leveraging compactly supported radial basis functions (CSRBFs) as weight functions. These CSRBFs are designed to be nonzero only within small, localized regions (“particles”) and vanish outside these regions, allowing efficient integral approximation with much fewer integral points even for complex or high-dimensional solutions. 
Specifically, Wendland’s CSRBFs \cite{wendland1995piecewise} are used as weight functions in \refeqp{eq:residual_weak} each of which has support over the ball $B(\bm{x}_j, R)$ centered at $\bm{x}_j$ and with radius $R=10^{-4}$. This value was selected to ensure accurate integral approximation with minimal integration points while reducing the likelihood of overlapping balls.
Given that a common hyperparameter $\lambda_{pde}$ multiplies all residuals in the virtual likelihood of \refeq{eq:virtuallike}, we rescale all weighted residuals as follows:
\be\label{eq:residual_rescale}
r_{w_j}(a,u) = \frac{1}{|B(\bm{x}_j,R)|}\int_{B(\bm{x}_j,R)} (a\nabla u\nabla w_j - fw_j) ~d\bmx
\ee 
where $|B(\bm{x}_j, R)|$ is the volume of the ball $B(\bm{x}_j,R)$. 
The final challenge is numerically approximating the integral in \refeqp{eq:residual_rescale}, which requires evaluating the integrand (including the coefficient $a$) at certain  integration points. We specify the number of integration points $N_{int}$ for each of the  problems in the respective Section \ref{sec:experiments}. 

\revise{In our DGenNO framework, the residuals are computed based on predicted quantities rather than true (reference) solutions. Specifically, for each input $a$ sampled or inferred during training, we obtain a corresponding latent variable $\bm{\beta}$ via an encoder $e_{\bphi}$, and then use the decoder $u_{\bm{\theta}_u}$ to produce a predicted solution $u$. The residual $r_j$ corresponding to the weighting function $w_j$ is then evaluated with the predicted $u$ and $a$ through Eq. \eqref{eq:residual_rescale}.} In the sequel, we provide the explicit expressions of the weighted residuals employed using the CSRBF-type weighting functions $w_j$ for each of the PDEs considered in Section \ref{sec:experiments}:
\paragraph{Darcy's flow problem} The  weighted residuals  for the Darcy's flow problem \eqref{eq:darcy_flow} were   given in \refeqp{eq:residual_rescale}.
\paragraph{Burger's equation} For the time-dependent Burger’s equation \eqref{eq:burgers}, one approach to handle the time variable is to treat it as an additional spatial dimension, thus transforming it into a time-independent PDE. However, this increases the dimensionality of the problem and requires more integration points to approximate the integrals in the residuals. We adopt a more efficient approach  whereby for each (space-dependent)  weight function $w_j$ we uniformly  sample a time stamp $t_j \in [0,T]$. 
The corresponding  residual $r_{w_j}$ 
is then given as:
\be\label{eq:weak_residual_burgers}
r_{w_j}(u) = \frac{1}{B(\bm{x}_j, R)}\int_{B(\bm{x}_j, R)} u_t w_j + u u_x w_j+ \nu u_x(w_j)_x ~d\bm{x}, \quad \text{where}\ t=t_j.
\ee
\paragraph{Stokes equation with cylindrical obstacle}
For the weighted residuals of  the Stokes flow PDE in  \refeqp{eq:stokes}, we follow the approach outlined in \cite{rao2020physics} and  re-express it as follows:
\be\label{eq:stokes_new}
\begin{split}
\nabla \cdot \bm{\sigma} &={\bf 0}, \quad \text{in}\ \Omega/\Omega_{cld}, \\
\nabla\cdot {\bf u} &= 0, \quad \text{in}\ \Omega/\Omega_{cld},
\end{split}
\ee
where $\bm{\sigma}$ is the Cauchy stress tensor which is defined as follows :
\be
\bm{\sigma} = -\nabla p\bm{I}+\mu(\nabla\bm{u} +\nabla\bm{u}^T),
\ee
and $p$ satisfies $p = -\text{tr}(\bm{\sigma})/2$.
In order to automatically enforce the incompressibility constraint $\nabla\cdot {\bf u} =0$, we employ a stream function $\psi(x_1,x_2)$,  i.e.  define the velocity $\bm{u}$ as $\bm{u} = (u_1,u_2) = (\nabla_{x_2} \psi, -\nabla_{x_1} \psi)$ and use the neural operator $\psi_{\bt_\psi}$ to approximate $\psi$.
%
For the stress tensor $\bm{\sigma}$, we model it using a neural operator with three outputs, denoted as $\bs{\sigma}_{\bt_\sigma}$. Additionally, we model the pressure term $p$ with a third neural operator model $p_{\bt_p}$. Their architectures are given in \ref{sec:stokes}.
As a result, the weak-form weighted residual $r_{w_j}$ for the Stokes equation \eqref{eq:stokes_new} is expressed as:
\be\label{eq:weak_residual_stokes}
r_{w_j}(\bm{\sigma}) = \frac{1}{B(\bm{x}_j, R)}\int_{B_j(\bm{x}_j,R)} \bm{\sigma} : \nabla \bm{w}_j ~ d\bm{x}.
\ee
where $\bm{w}_j$ is a vector-valued function of dimension two expressed with  CSRBFs with support in $\Omega/\Omega_{cld}$, where $\Omega = [0,2]\times[0,1]$ and $\Omega_{cld}$ is a cylinder centered at $(0.5,0.6)$ with radius $0.1$.

\section{Network structures}
\label{sec:network}
In this section, we provide details on the neural-network   architectures as well as the associated hyperparameters for each of the  problems considered in Section \ref{sec:experiments}. 
\subsection{Darcy's problem}
\label{sec:darcy}
\paragraph{The Encoder $e_{\bphi}$} For the encoder model $e_{\bphi}$, which extracts latent representations from the input coefficient, a Convolutional Neural Network (CNN) followed by a Feed-Forward Fully Connected Network (FFCN) was utilized. The CNN  consists of three hidden layers, each with an output channel size of 64. The kernel size is set to (3,3), and the stride is 2. The subsequent FFCN contains two hidden layers, each comprising 128 neurons. The SiLU activation function is applied to all hidden layers in the encoder model, while the Tanh activation function is used in the output layer to ensure that the output resides within a standard cubic region.

\paragraph{The neural operator models $u_{\bt_u}$ and $\mu_{\bt_a}$} Both the MultiONet-based neural operator models $u_{\theta_u}$ and $\mu_{\theta_a}$ in the DGenNO framework share the same architecture. Specifically, the branch network is an FFCN with six hidden layers, each consisting of 80 neurons. The trunk network has an identical structure to the branch network. In both networks, we use a custom activation function, originally proposed in \cite{zang2023particlewnn}, for all hidden layers. This activation, referred to as Tanh\_Sin, combines the Tanh and Sinc functions in the following form:
\be\label{eq:tanh_sin}
\text{Tanh\_Sin}(x) = \text{Tanh}(\sin(\pi x + \pi)) + x.
\ee
For the neural operator $\mu_{\bt_a}$, a Sigmoid activation is also applied in the output layer to map its output values to the range $[0,1]$.  

For the neural operator for the PDE-solution $u$ in the PI-DeepONet framework, we adopt the same trunk network structure as in the proposed DGenNO framework. For the branch network, we connect the above encoder network and the branch network from the DGenNO framework in series, serving as the branch network in the PI-DeepONet framework. As a result, the number of training parameters for both models remains nearly identical, ensuring a fair comparison.

\subsection{Burgers' equation}
\label{sec:burger}
\paragraph{The Encoder $e_\phi$} To extract the latent representation $\bs{\beta}$ from the input coefficient, we employ an FFCN with four hidden layers. The number of neurons in each hidden layer is set to 128, 128, 64, and 64, respectively. The Exponential Linear Unit (ELU) activation function is applied to all hidden layers, while the output layer employs the Tanh activation to ensure that the latent space resides within a standardized cubic region.

\paragraph{The MultiONet-based neural operator models $u_{\bt_u}$}  For the Burgers’ problem \eqref{eq:burgers}, the input function $a$ corresponds to the initial condition, i.e., the value of the solution $u$ at $t=0$. As a result, only the neural operator model $u_{\bt_u}$ is required for the solution, and the recovered input coefficient can be obtained by evaluating $u_{\bt_u}(\bs{\beta})$ at the initial moment. For the model $u_{\bt_u}$ in the DGenNO framework, the branch network is an FFCN with six hidden layers, each consisting of 80 neurons. The trunk network has an identical structure to the branch network. The Tanh\_Sin activation function \eqref{eq:tanh_sin} is applied to all hidden layers. 

For the neural operator  for the PDE-solution $u$ in the PI-DeepONet framework, we adopt the same trunk network structure as in the proposed framework. To match the branch network, we concatenate the encoder network and the branch network from the DGenNO framework in series, serving as the branch network in the PI-DeepONet framework. As a result, the number of trainable parameters in both models remains nearly identical, ensuring an unbiased comparison.

\subsection{Stokes equation with a  cylindrical obstacle}
\label{sec:stokes}
\paragraph{The Encoder $e_{\bphi}$} To extract the latent representation $\bs{\beta}$ from the input coefficient, an FFCN with three hidden layers is used. The hidden layers consist of 256, 128, and 64 neurons, respectively. The Tanh activation function is applied to all hidden layers, including the output layer, to ensure that the latent space resides within a standardized cubic region.

\paragraph{The neural operator models $\psi_{\bt_\psi}$, $p_{\bt_p}$, and $\bs{\sigma}_{\bt_\sigma}$} As discussed in Section \ref{sec:weak_residual}, we employ three, MultiONet-based, neural operator models for the Stokes' problem \eqref{eq:stokes} (or \eqref{eq:stokes_new}): $\psi_{\bt_\psi}$ for the stream function $\psi$,  $p_{\bt_p}$ for the pressure field $p$, and $\bs{\sigma}_{\bt_\sigma}$ for the stress tensor $\bs{\sigma}$. For these three models, we use the same neural operator architecture as in the Burgers’ problem, with necessary modifications to accommodate the output dimension.

For a fair comparison, we adopt the same trunk network structure used in the proposed DGenNO framework for the neural operator representing the solution $u$ in the PI-DeepONet framework. For the branch network, we connect the Encoder network and the branch network from the DGenNO framework in series, using it as the branch network for the neural operators in the PI-DeepONet framework. As a result, the number of trainable parameters in both models remains nearly identical.
\subsection{Inverse problems: the piecewise-constant case}
\label{sec:inverse_pwc}
The network architectures of the proposed DGenNO framework for this problem are detailed in Section \ref{sec:darcy}. Once the forward model is trained, the target coefficient can be directly obtained using the trained model $\mu_{\bt_a}(\bs{\beta}^*)$, provided that the optimal $\bs{\beta}^*$ is identified. To achieve this, an FFCN is used to parameterize the latent variable $\bs{\beta}$, consisting of three hidden layers with 64, 64, and 128 neurons, respectively. The ReLU activation function is applied to all hidden layers, while the Tanh activation is used in the output layer to ensure the output remains within a standard cubic region. The number of weighted residuals (i.e., virtual observations) is set to  $M=100$, with $N_{int}=25$  integral points used for computing the integral. The hyperparameters are set as $\lambda_{pde} = 1$ for virtual observation and  $\lambda_{data} = 50$ for noisy solution observation. The model is trained using the ADAM optimizer with an initial learning rate of  $lr=10^{-2}$, which decays to $1/3$ of its value every 250 iterations over a total of 1000 iterations.

As discussed in Section \ref{sec:inverse}, the ParticleWNN and PINN methods serve as state-of-the-art benchmarks. In the ParticleWNN method, the network model used to parameterize the solution $u$ is an FFCN with six hidden layers, each containing 80 neurons. The Tanh\_Sin activation is applied to all hidden layers. For approximating the target coefficient $a$, an FFCN with three hidden layers (64, 64, and 128 neurons) is employed, with ReLU activation used for all hidden layers. The number of weighted residuals is set to $M=100$, with  $N_{int}=25$ integral points used for computing the integral. The hyperparameters are set as  $\lambda_{pde} = 1$ for virtual observation and  $\lambda_{data} = 50$ for noisy solution observation. The ADAM optimizer is used for training with an initial learning rate of $lr=10^{-3}$, which decays to $1/3$ of its value every 2500 iterations over a total of 10000 iterations to ensure convergence.

For a fair comparison, the models used to parameterize the solution $u$ and target coefficient $a$ in the PINN method are identical to those in the ParticleWNN method. The number of collocation points is set to $N=2500$, and the hyperparameters are  $\lambda_{pde} = 1$ for virtual observation derived from the strong-form residuals and $\lambda_{data} = 50$ for noisy solution observation. The PINN model is trained using the ADAM optimizer with an initial learning rate of $lr=10^{-3}$, which decays to $1/3$ of its value every 2500 iterations over a total of 10000 iterations to ensure convergence.
\subsection{Inverse problems: the continuous case}
\label{sec:inverse_cts}
For the inverse problem with continuous targets, the network architectures in the proposed DGenNO framework remain the same as in Section \ref{sec:darcy}, except that the Sigmoid activation in the output layer of the neural operator $\mu_{\bt_a}$ is no longer necessary. The model used to parametrize the latent variable $\bs{\beta}$ follows the same structure as in the piecewise-constant case in Section \ref{sec:inverse_pwc}. We maintain the number of weighted residuals at $M=100$ and use $N_{int}=25$  integral points for computing the integral. The hyperparameters are set to $\lambda_{pde} = 1$ for virtual observation and $\lambda_{data} = 25$ for noisy solution observation. The ADAM optimizer is employed for training, with an initial learning rate of $lr=10^{-2}$, which decays to $1/3$ of its value every 250 iterations over a total of 1000 iterations.

For the ParticleWNN method, the network structures for parametrizing both the solution $u$ and the coefficient $a$ remain the same as in the piecewise-constant case. However, the activation functions in the hidden layers of the coefficient model are replaced with Tanh\_Sin activations. For loss computation, we set $M=100$ weighted residuals and  $N_{int}=25$ integral points. The hyperparameter  $\lambda_{pde}$ is set to $1$ for virtual observation, while $\lambda_{data}$ for noisy solution observation is adjusted based on different noise levels: 20, 50, 100  for SNR values of 20, 50, 100, respectively. The ADAM optimizer is used for training, with an initial learning rate of $lr=10^{-3}$, which decays to $1/3$ of its value every 250 iterations over a total of 1000 iterations.

For the PINN method, the models for parametrizing the solution $u$ and the target coefficient $a$ are identical to those in the ParticleWNN method. The number of collocation points is set to $N=2500$, and $\lambda_{pde}$ is set to $1$ for virtual observation. The hyperparameter $\lambda_{data}$ for noisy solution observation is adjusted to 200, 500, and 1000 for different noise levels with SNR values of  20, 50, and 100, respectively. The PINN model is trained using the ADAM optimizer, with an initial learning rate of  $lr=10^{-3}$, decaying to $1/3$ of its value every 2500 iterations over 10000 iterations to ensure convergence.

As mentioned in Section \ref{sec:inverse}, we also solve the inverse problem with continuous targets using the PI-DeepONet method for comparison. The neural operator structure for the solution $u$ in the PI-DeepONet framework remains the same as in the piecewise-constant case in Section \ref{sec:darcy}. To recover the coefficient $a$ from the noisy solution observation $\bs{u}_{obs}$, we follow Equation \eqref{eq:posterior_inv} in Section \ref{sec:predictions} to infer the posterior $p(a|\bs{u}_{obs})$, as the latent variable $\bs{\beta}$ is not available in the PI-DeepONet method. The model used to parametrize the coefficient $a$ follows the same network structure as in the ParticleWNN and PINN methods. The ADAM optimizer is used for optimization, with an initial learning rate of $lr=10^{-3}$, decaying to $1/3$ of its value every 2500 iterations over 10000 iterations to ensure convergence.

\section{Nonlinear Poisson equation}
\label{sec:nonlinear}
To demonstrate the capability of the proposed DGenNO framework in solving nonlinear PDEs, we consider the following nonlinear Poisson equation, as introduced in \cite{chatzopoulos2024physics}:
\be\label{eq:nonlinear}
\begin{split}
-\nabla \cdot (a(\bm{x})e^{\alpha(u(\bm{x})-\bar{u})}\nabla u(\bm{x})) &= f(\bm{x}), \quad \bm{x}\in \Omega = [0,1]^2, \\
u(\bm{x}) &= 0, \quad \bm{x}\in \partial\Omega,
\end{split}
\ee
where the meanings of $u$, $a$, and $f$ are consistent with those in Section~\ref{sec:darcy_flow}. The parameters $\alpha$ and $\bar{u}$\footnote{Following \cite{chatzopoulos2024physics}, we use $\alpha = 0.05$ and $\bar{u} = 5$ in our experiments.} are two additional scalars (assumed constant across $\bm{x}$). The parameter $\alpha$ (and to a lesser extent $\bar{u}$) controls the degree of nonlinearity: when $\alpha = 0$, Eq. \eqref{eq:nonlinear} reduces to the linear Darcy flow equation in Eq. \eqref{eq:darcy_flow}; as $\alpha$ increases, the PDE becomes increasingly nonlinear.
Notably, the DGenNO framework requires no architectural changes to handle this nonlinear case. The only modification lies in the formulation of the weighted residuals. Specifically, Eq.~\eqref{eq:residual_rescale} is adapted as:
\be\label{eq:residual_nonlinear}
r_{w_j}(a,u) = \frac{1}{|B(\bm{x}_j,R)|}\int_{B(\bm{x}_j,R)} \left(ae^{\alpha(u - \bar{u})}\nabla u \cdot \nabla w_j - f w_j\right) ~d\bm{x},
\ee
where $|B(\bm{x}_j, R)|$ denotes the volume of the ball $B(\bm{x}_j, R)$.

\begin{figure}[!htbp]
    \centering  
    \subfigure[True $a$, the truth $u$ (non-linear case), and truth $u$ (linear case)]{\label{fig:problem_nlin}
        \includegraphics[width=1.\textwidth]{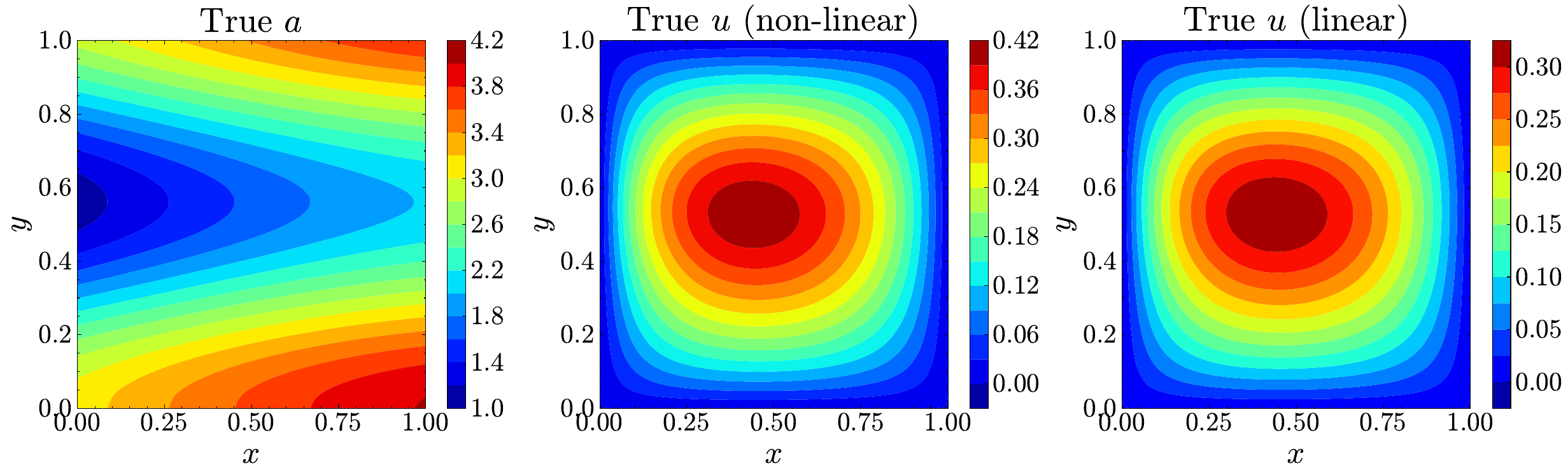}}
    \subfigure[Non-linear case]{\label{fig:nlin}
        \includegraphics[width=0.8\textwidth]{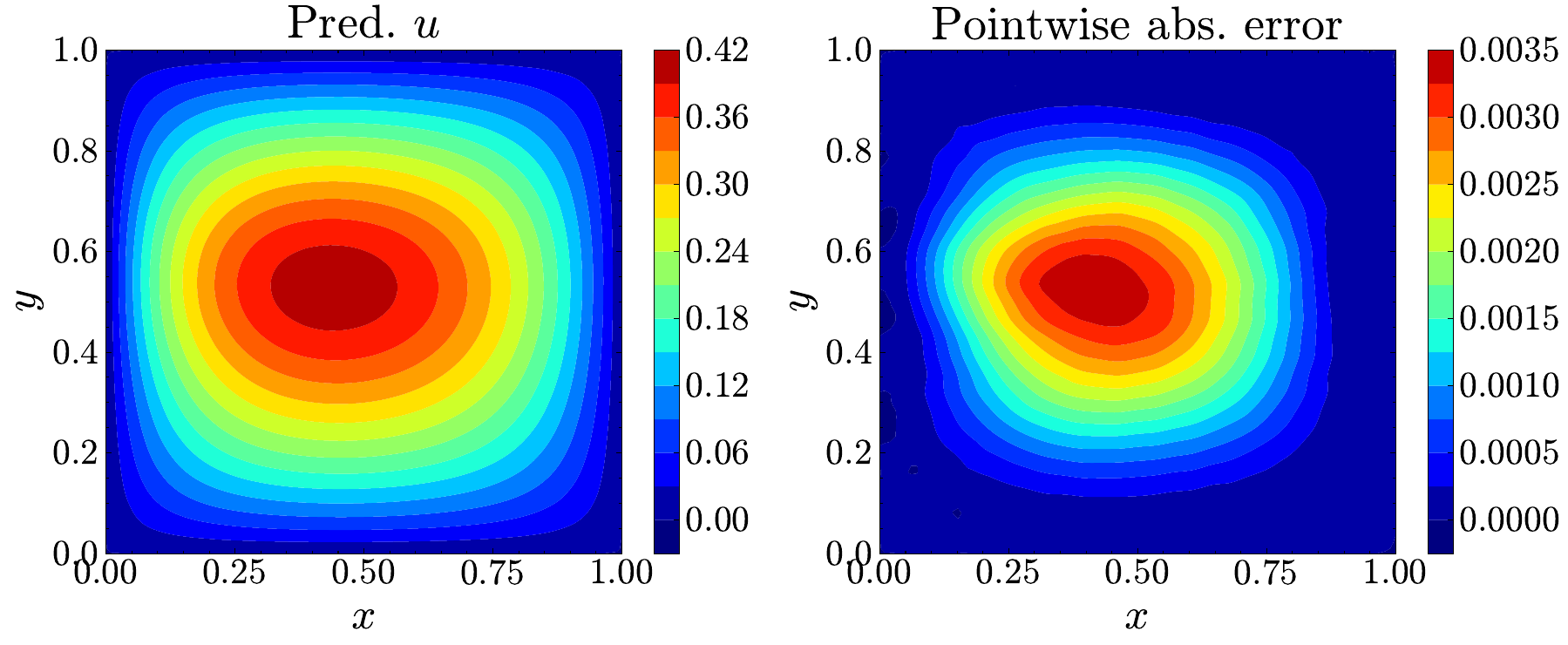}}
    \subfigure[Linear case]{\label{fig:lin}
        \includegraphics[width=0.8\textwidth]{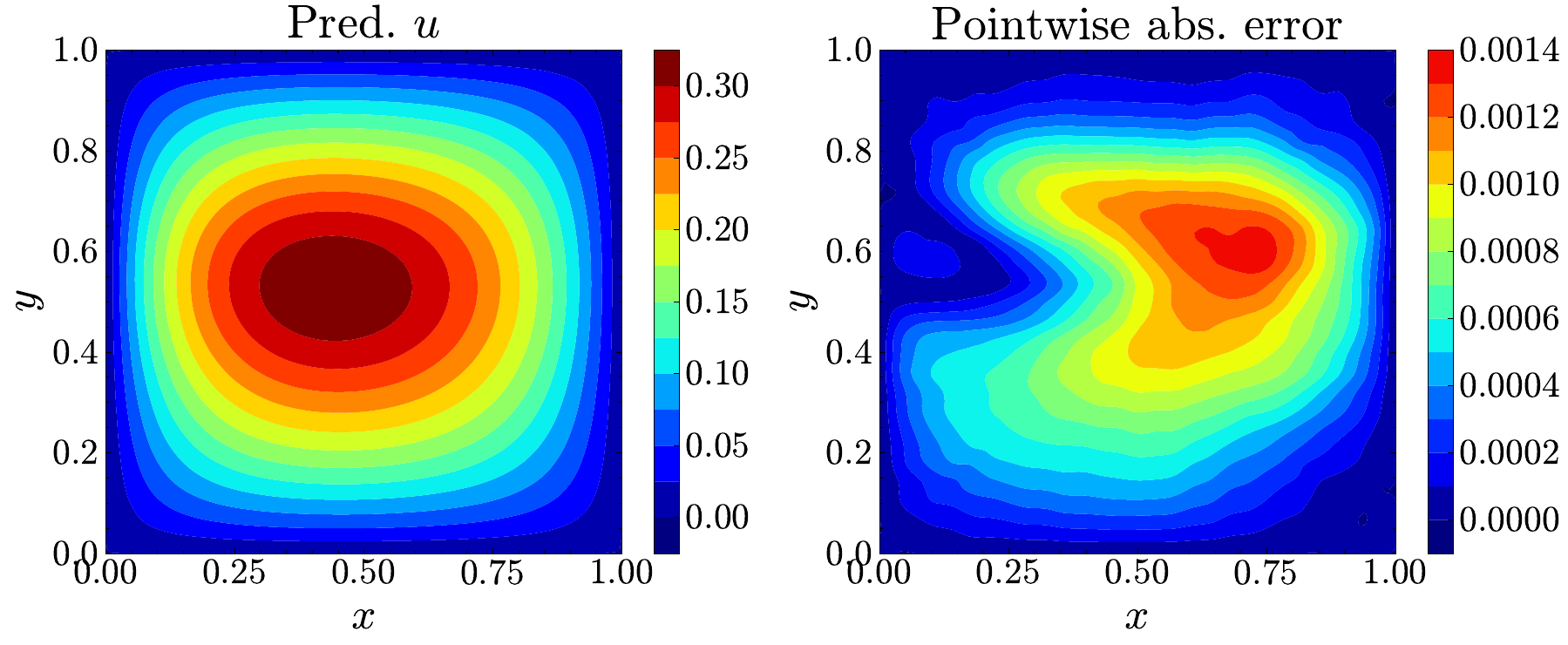}}
    \vspace{-0.25cm}
    \caption{The performance of DGenNO in solving a nonlinear PDE and corresponding linear counterpart, both with the same coefficient field $a$: (a) true $a$ (left), true solution $u$ in non-linear case (middle), and true solution $u$ in linear case (right); (b) Pred. $u$ and pointwise absolute error obtained by DGenNO in non-linear case; (c) Pred. $u$ and pointwise error obtained by DGenNO in linear case.} 
    \label{fig:nlin_vs_lin}
\end{figure}

We solve the parametric nonlinear PDE~\eqref{eq:nonlinear} to learn the operator map $\mathcal{G}: a \mapsto u$, using the same class of continuous permeability fields as described in Section~\ref{sec:inverse}. Specifically, the input field takes the form $a(x_1, x_2) = 2.1 + \sin(k_1 x_1) + \cos(k_2 x_2)$, where $k_1$ and $k_2$ are independently sampled from the uniform distribution $U(0, 2\pi)^2$. All experimental settings for training the DGenNO model are kept identical to those used in Section~\ref{sec:inverse}.
On the same testing dataset in Section~\ref{sec:inverse}., the DGenNO achieves a root mean square error (RMSE) of $6.91 \times 10^{-3} \pm 3.72 \times 10^{-3}$ in predicting $u$.
The performance shows a slight degradation compared to the linear case (where the RMSE is $4.31 \times 10^{-3} \pm 5.76 \times 10^{-3}$), which reflects the increased complexity introduced by the nonlinear term in the PDE. In Figure~\ref{fig:nlin_vs_lin}, we compare the performance of DGenNO in solving a nonlinear PDE and its corresponding linear counterpart, both with the same coefficient field. As shown in Figures~\ref{fig:nlin} and~\ref{fig:lin}, the pointwise absolute errors are significantly smaller in the linear case, highlighting the increased complexity and difficulty introduced by the nonlinearity.
Nevertheless, the DGenNO framework remains capable of producing accurate predictions, demonstrating its robustness and adaptability to nonlinear PDEs. 
\end{document}